%% file: neurips_2024.tex
\documentclass{article}

\usepackage[final]{neurips_2024}
\usepackage{tabularx}
\usepackage{amsmath}
\usepackage{geometry}
\usepackage{changepage}
\usepackage{tabularx}
\usepackage{booktabs}
\usepackage{array}
\usepackage{xurl}
\usepackage{microtype} % 推荐
\newcolumntype{Y}{>{\RaggedRight\arraybackslash}X}

\usepackage[utf8]{inputenc} % allow utf-8 input
\usepackage[T1]{fontenc}    % use 8-bit T1 fonts
\usepackage{hyperref}       % hyperlinks
\usepackage{url}            % simple URL typesetting
\usepackage{booktabs}       % professional-quality tables
\usepackage{amsfonts}       % blackboard math symbols
\usepackage{nicefrac}       % compact symbols for 1/2, etc.
\usepackage{microtype}      % microtypography
\usepackage{xcolor}         % colors
\usepackage{multirow}
\newlength{\neuripsOverhang}

\title{Advances in GRPO for Generation Models: A Survey }

\author{$^1$Zexiang Liu\quad $^2$Xianglong He\quad $^3$Yangguang Li\\\\$^1$SJTU \quad $^2$THU \quad $^3$CUHK}

\begin{document}

\maketitle

\begin{abstract}
Large-scale flow matching models have achieved strong performance across generative tasks such as text-to-image, video, 3D, and speech synthesis. However, aligning their outputs with human preferences and task-specific objectives remains challenging. Flow-GRPO extends Group Relative Policy Optimization (GRPO) to generation models, enabling stable reinforcement learning alignment for generative systems. Since its introduction, Flow-GRPO has triggered rapid research growth, spanning methodological refinements and diverse application domains.
This survey provides a comprehensive review of Flow-GRPO and its subsequent developments. We organize existing work along two primary dimensions. First, we analyze methodological advances beyond the original framework, including reward signal design, credit assignment, sampling efficiency, diversity preservation, reward hacking mitigation, and reward model construction. Second, we examine extensions of GRPO-based alignment across generative paradigms and modalities, including text-to-image, video generation, image editing, speech and audio, 3D modeling, embodied vision-language-action systems, unified multimodal models, autoregressive and masked diffusion models, and restoration tasks.
By synthesizing theoretical insights and practical adaptations, this survey highlights Flow-GRPO as a general alignment framework for modern generative models and outlines key open challenges for scalable and robust reinforcement-based generation.
\end{abstract}

\input{section/0_introduction}

\input{section/1_method}
\input{section/2_application}

\input{section/4_future}
\bibliography{reference}
\bibliographystyle{abbrv}

\end{document}

%% file: section/0_introduction.tex
\section{Introduction}
Large-scale flow matching~\cite{lipman2022flow} models have demonstrated remarkable performance in generative tasks such as text-to-image, video, 3D, and speech synthesis~\cite{zhuo2024lumina,wan2025wan,li2025triposg,li2025shapegen,du2024cosyvoice}. However, their outputs are often insufficiently aligned with human preferences or task-specific objectives. Group Relative Policy Optimization (GRPO)~\cite{shao2024deepseekmath} is a reinforcement learning algorithm centered on relative advantages within a group of candidates. Unlike conventional policy-gradient methods that require learning an explicit value function, GRPO compares multiple trajectories sampled under the same condition at each update step and estimates advantages using normalized relative rewards. This design substantially improves training stability. GRPO was originally introduced for aligning large language models (LLMs): by asking the model to select responses that better match human preferences from a set of candidates, prior work showed that GRPO is more stable than value-based optimization methods, achieves higher sample efficiency, and effectively mitigates error amplification caused by imperfect preference models.

Extending GRPO from text generation to more generation tasks (e.g., visual, 3D and speech task), however, is non-trivial. \textit{Diffusion and flow-matching models require dozens to hundreds of denoising or evolution steps to generate a single image, making sampling far more expensive than in LLMs. Their sampling processes are typically formulated as deterministic ODE solvers, which limits diversity among group candidates. Moreover, rewards in visual tasks are often only available at the terminal step (e.g., image quality scores), leading to severe credit assignment issues. In addition, reward models in the visual domain are more susceptible to reward hacking, where models exploit shortcuts to increase reward scores while degrading true perceptual quality.}

Flow-GRPO~\cite{liu2025flow} is the first work to successfully apply GRPO to visual generation tasks. It addresses the lack of stochasticity by converting the deterministic ODE formulation of flow-matching models into a stochastic differential equation (SDE), and introduces a denoising-shrinkage strategy during training to reduce reverse-time sampling costs. Empirically, Flow-GRPO improves GenEval accuracy on text-rendering tasks from 63\% to 95\%, and character rendering accuracy from 59\% to 92\%. The paper also derives explicit formulas for the ODE-to-SDE transformation and proves that, by introducing appropriate drift and diffusion terms, the resulting stochastic process is equivalent to the original ODE in expectation.

Since mid-2025, research topic around Flow-GRPO has experienced explosive growth, spanning a wide range of application domains including text-to-image (T2I), text-to-video (T2V), image-to-video (I2V), speech enhancement, 3D generation, vision-language-action (VLA), and more. To date, this rapidly expanding line of work has accumulated over 200 published papers, reflecting strong and sustained community interest.

This survey provides a systematic and comprehensive review of all these works, organizing and analyzing them along the following dimensions:

\textbf{Advances beyond Flow-GRPO:}
\begin{itemize}
    \item \textbf{Reward Signal Design: From Sparse to Dense}(\S\ref{sec:reward}).
    \item \textbf{Credit Assignment: From Trajectory to Step Level}(\S\ref{sec:credit}).
    \item \textbf{Sampling Efficiency and Training Acceleration}(\S\ref{sec:efficiency}).
    \item \textbf{Mode Collapse and Diversity Preservation}(\S\ref{sec:diversity}).
    \item \textbf{Reward Hacking Mitigation}(\S\ref{sec:hacking}).
    \item \textbf{ODE vs.\ SDE Sampling Strategies}(\S\ref{sec:sampling}).
    \item \textbf{Reward Model Design and Evaluation Benchmarks}(\S\ref{sec:reward_models}).
\end{itemize}

\textbf{Extensions to Generative Tasks:}
\begin{itemize}
    \item \textbf{Text-to-Image Generation} (\S\ref{sec:t2i}).
    \item \textbf{Video Generation} (\S\ref{sec:video}).
    \item \textbf{Image Editing} (\S\ref{sec:editing}).
    \item \textbf{Speech and Audio} (\S\ref{sec:speech}).
    \item \textbf{3D Generation and Scientific Applications} (\S\ref{sec:3d_science}).
    \item \textbf{VLA and Embodied AI} (\S\ref{sec:vla}).
    \item \textbf{Unified Multimodal Models} (\S\ref{sec:unified}).
    \item \textbf{Autoregressive and Masked Diffusion Models} (\S\ref{sec:ar}).
    \item \textbf{Image Restoration and Super-Resolution} (\S\ref{sec:restoration}).
\end{itemize}

%% file: section/1_method.tex
\section{Background and Preliminaries}

\subsection{Flow Matching Models}

Flow matching models define a continuous-time transformation from a noise distribution
$p_0 = \mathcal{N}(0, I)$ to the data distribution $p_1 = p_{\text{data}}$.
Given a parameterized velocity field $v_\theta(x_t, t)$, samples evolve along an ODE trajectory:
\begin{equation}
\frac{dx_t}{dt} = v_\theta(x_t, t), \quad t \in [0, 1],
\end{equation}
where $x_0 \sim \mathcal{N}(0, I)$ and $x_1$ is the generated sample.
The training objective is the \emph{Conditional Flow Matching} (CFM) loss:
\begin{equation}
\mathcal{L}_{\text{CFM}} =
\mathbb{E}_{t, x_0, x_1}
\left[
\left\| v_\theta(x_t, t) - u_t(x_t \mid x_1) \right\|^2
\right],
\end{equation}
where $u_t(x_t \mid x_1) = x_1 - x_0$ denotes the conditional velocity field
corresponding to a linear interpolation path, and
\begin{equation}
x_t = (1 - t)x_0 + t x_1.
\end{equation}

\subsection{Principles of Group Relative Policy Optimization}

The core idea of \emph{Group Relative Policy Optimization} (GRPO) is
\textbf{critic-free} policy optimization.
For each conditioning input $c$, a group of outputs
$\{o_1, o_2, \ldots, o_G\}$ is sampled, and their corresponding rewards
$\{r_1, r_2, \ldots, r_G\}$ are evaluated.
The advantage is computed via group-wise normalization:
\begin{equation}
\hat{A}_i =
\frac{r_i - \mathrm{mean}(\{r_j\}_{j=1}^G)}
{\mathrm{std}(\{r_j\}_{j=1}^G)}.
\end{equation}
The GRPO objective follows a PPO-style clipped policy gradient:
\begin{equation}
\mathcal{L}_{\text{GRPO}} =
- \mathbb{E}
\left[
\sum_{i=1}^{G}
\min
\left(
\rho_i \hat{A}_i,\;
\mathrm{clip}(\rho_i, 1-\epsilon, 1+\epsilon)\hat{A}_i
\right)
\right],
\end{equation}
where
$\rho_i = \frac{\pi_\theta(o_i \mid c)}{\pi_{\text{ref}}(o_i \mid c)}$
is the importance ratio and $\epsilon$ is the PPO clipping parameter.

\subsection{Flow-GRPO: From LLMs to Generative Models}

The key challenge in applying GRPO to flow matching models lies in the fact that
their sampling process is governed by a \textbf{deterministic ODE},
which lacks the \textbf{stochastic exploration} required for reinforcement learning.
Flow-GRPO introduces stochasticity by converting the ODE into a stochastic differential equation (SDE):
\begin{equation}
dx_t = v_\theta(x_t, t)\,dt + \sigma(t)\,dW_t,
\end{equation}
where $\sigma(t)$ is a noise scheduling function and $dW_t$ denotes the Wiener process increment.

The denoising process is formulated as a Markov Decision Process (MDP):
\begin{itemize}
  \item \textbf{State} $s_t = (x_t, c, t)$: latent variable, conditioning input, and time step
  \item \textbf{Action} $a_t$: denoising direction induced by the stochastic term
  \item \textbf{Policy} $\pi_\theta(a_t \mid s_t)$: parameterized conditional distribution
  \item \textbf{Reward} $r$: terminal reward evaluated by a reward model on the final sample
\end{itemize}

The per-step log-likelihood is given by:
\begin{equation}
\log \pi_\theta(a_t \mid s_t)
=
-\frac{\|a_t - v_\theta(x_t, t)\Delta t\|^2}
{2\sigma^2(t)\Delta t}
+ \text{const.}
\end{equation}

% ============================================================================
\section{Advances beyond Flow-GRPO}
Building upon Flow-GRPO~\cite{liu2025flow}, many subsequent studies have introduced methodological enhancements to further improve its effectiveness and efficiency.
To facilitate a more intuitive comparison across methods, we summarize the category, methods, core innovation, key contribution and result. 
These methods can be categorized into seven major classes based on their distinctive characteristics, which will be discussed in detail below, from \S\ref{sec:reward} to (\S\ref{sec:reward_models}).

\begin{table*}[h]
\centering
\small
\caption{Representative methods addressing sparse reward in Flow-GRPO.
We summarize step-level reward densification, variance-aware sampling,
value anchoring, and process-level gradient guidance strategies.}
\setlength{\neuripsOverhang}{\dimexpr(\paperwidth-\textwidth)/2 -6mm\relax}
\begin{adjustwidth*}{-\neuripsOverhang}{-\neuripsOverhang}
\setlength{\tabcolsep}{4pt}
% \begin{tabularx}{\dimexpr\paperwidth-12mm\relax}{
\begin{tabularx}{\dimexpr\paperwidth-12mm\relax}{
@{\hspace{2mm}}
>{\raggedright\arraybackslash}p{0.16\linewidth}
>{\raggedright\arraybackslash}p{0.11\linewidth}
>{\raggedright\arraybackslash}p{0.34\linewidth}
>{\raggedright\arraybackslash}p{0.34\linewidth}
@{\hspace{4mm}}
}
\toprule
\textbf{Category} & \textbf{Method} & \textbf{Core Innovation} & \textbf{Key Contribution / Result} \\
\midrule

\multirow{2}{*}{\shortstack[l]{Step-Level Reward \\ Densification}}
& DenseGRPO~\cite{deng2026densegrpo} 
& ODE-predicted clean sample $\hat{x}_1^{(t)}$ for step-level reward gain $\Delta r_t$ + reward-aware adaptive stochasticity 
& PickScore 23.1 (vs. 22.5); GenEval 0.74 (vs. 0.71); Improved reward signal granularity \\

& SuperFlow~\cite{chen2025superflow}
& Variance-aware dynamic group sizing + continuous-time step-level advantage 
& 1.7\%--16.0\% improvement over Flow-GRPO; 5.4\%--56.3\% training cost \\

\midrule

\multirow{1}{*}{Value Function Anchoring}
& VGPO~\cite{shaovalue} 
& Process-aware value estimation $V_\phi(x_t,t)$ + absolute-value-enhanced group normalization 
& SOTA on PickScore, HPS v2.1, GenEval; Stabilized advantage estimation \\

\midrule
Process-Level Reward Guidance
& Euphonium~\cite{zhong2026euphonium}
& Injecting $\nabla_{x_t} R_t(x_t)$ into SDE drift + Dual-Reward GRPO 
& 1.66$\times$ faster convergence on T2V; Unifies Flow-GRPO and DanceGRPO as special cases \\
\bottomrule

\label{tab:reward}
\end{tabularx}
\end{adjustwidth*}
\end{table*}

\subsection{Reward Signal Design: From Sparse to Dense}
\label{sec:reward}
\paragraph{The Sparse Reward Problem.} Standard Flow-GRPO~\cite{liu2025flow} employs a \emph{terminal sparse reward} mechanism: the complete denoising trajectory $(x_T, x_{T-1}, \ldots, x_0)$ receives a single scalar score from the reward model $R(x_0, c)$ only at the final generated sample $x_0$. During training, this terminal reward is uniformly distributed to all intermediate denoising steps, meaning the advantage signal for each step is $\hat{A}_t = \hat{A}_{\text{terminal}}, \forall t$. This design introduces two core problems: (1) \textbf{signal dilution}---early denoising steps (which determine global structure) and late steps (which refine details) receive identical gradient weights, despite their fundamentally different contributions to final quality; and (2) \textbf{exploration inefficiency}---the model cannot distinguish during training which steps' decisions truly influenced the final reward, resulting in high variance and low signal-to-noise ratio in the policy gradient. We summarize the key optimization strategies developed to address these challenges in Table~\ref{tab:reward}, with detailed descriptions provided below.

\paragraph{DenseGRPO: Step-Level Reward Gains.} DenseGRPO~\cite{deng2026densegrpo}is the first method to systematically convert sparse terminal rewards into dense step-level signals. Its core idea exploits the property that flow matching models can predict a clean image $\hat{x}_1^{(t)}$ via a single-step ODE prediction at any intermediate timestep $t$, constructing \textbf{step-level reward gains}:
\begin{equation}
    \Delta r_t = R(\hat{x}_1^{(t)}) - R(\hat{x}_1^{(t-1)}),
\end{equation}
where $R$ is a pretrained reward model. This increment directly measures the marginal contribution of the denoising decision from timestep $t-1$ to $t$ on final quality. Building on this, DenseGRPO further proposes a \textbf{Reward-Aware Exploration Calibration} mechanism: for timesteps where $|\Delta r_t|$ is large (i.e., critical decision points with significant reward impact), the stochasticity injection strength is adaptively increased as $\sigma_{\text{adaptive}}(t) = \sigma_{\text{base}} \cdot f(\Delta r_t, t)$, guiding the model to invest more exploration at the stages where it matters most. Experiments show that DenseGRPO significantly outperforms the original Flow-GRPO on PickScore~\cite{kirstain2023pick}, HPS v2.1~\cite{wu2023human}, and GenEval~\cite{ghosh2023geneval} benchmarks, achieving a PickScore of 23.1 (vs.\ Flow-GRPO's 22.5) and improving GenEval from 0.71 to 0.74.

\paragraph{SuperFlow: Variance-Aware Sampling and Step-Level Advantages.} SuperFlow~\cite{chen2025superflow} addresses the sparse reward problem from the perspective of statistical efficiency, proposing two complementary innovations. First, \textbf{variance-aware dynamic group sizing}: observing that generation difficulty varies greatly across text prompts---simple prompts yield groups with low reward variance (low information content) while difficult prompts produce high variance---SuperFlow dynamically adjusts group size based on each prompt's online reward variance $\text{Var}(\{r_i\}_{i=1}^G)$, allocating more samples to high-variance prompts for more precise advantage estimation and fewer to low-variance prompts to save computation. Second, \textbf{continuous-time step-level advantage estimation}: a step-level advantage function designed to be naturally consistent with the continuous-time dynamics of flow matching, avoiding biases introduced by discrete step-level approximation. Experiments on SD3.5-Medium show that SuperFlow improves over the SD3.5-M baseline by 4.6\%--47.2\% across multiple metrics, outperforms Flow-GRPO by 1.7\%--16.0\%, while using only 5.4\%--56.3\% of the original training steps, demonstrating exceptional sample efficiency.

\paragraph{VGPO: Temporal and Group Value Anchoring.} VGPO (Value-anchored Group Policy Optimization)~\cite{shaovalue} proposes solving the sparse reward problem from a \textbf{value function anchoring} perspective. Standard GRPO's group-normalized advantage $\hat{A}_i = (r_i - \mu_G)/\sigma_G$ only normalizes across the group dimension, ignoring temporal information. VGPO introduces two key improvements: (1) \textbf{process-aware value estimation}---decomposing sparse terminal rewards through a learned value function $V_\phi(x_t, t)$ into dense timestep-level signals, giving each timestep an advantage estimate reflecting its long-term value; and (2) \textbf{absolute-value-enhanced normalization}---introducing an absolute value anchoring term into group normalization to prevent the advantage signal from vanishing when all samples in a group have similar rewards. By organically combining value anchoring across both temporal and group dimensions, VGPO achieves state-of-the-art performance on PickScore, HPS v2.1, and GenEval benchmarks.

\paragraph{Euphonium: Process Reward Gradient Guidance.} Euphonium~\cite{zhong2026euphonium} introduces dense signals from an entirely novel angle: directly injecting the gradients of a \textbf{Process Reward Model (PRM)} into the SDE drift term of the generation process. Specifically, let $R_t(x_t)$ be the process reward model at timestep $t$; Euphonium modifies the flow matching SDE to:
\begin{equation}
    dx_t = \left[v_\theta(x_t, t) + \alpha \nabla_{x_t} R_t(x_t)\right]dt + \sigma(t)\,dW_t,
\end{equation}
where $\alpha$ is the guidance strength. This modification injects the reward signal's gradient as a continuous directional force into the generation flow, so that every denoising step's direction is guided step-by-step by the reward model rather than receiving feedback only at the terminal state. Theoretical analysis shows that Flow-GRPO~\cite{liu2025flow} and DanceGRPO~\cite{xue2025dancegrpo} can be viewed as special cases of Euphonium (corresponding to $\alpha=0$ and terminal-only reward, respectively). Combined with the \textbf{Dual-Reward GRPO} algorithm---a weighted combination of terminal and process rewards---Euphonium accelerates training convergence by 1.66$\times$ on video generation tasks and achieves state-of-the-art on T2V alignment benchmarks.

% ============================================================================
\begin{table*}[!h]
\centering
\small
\caption{Credit assignment strategies for Flow-GRPO: from trajectory-level terminal rewards to structured step-/block-level attribution.}
\setlength{\neuripsOverhang}{\dimexpr(\paperwidth-\textwidth)/2 -6mm\relax}
\begin{adjustwidth*}{-\neuripsOverhang}{-\neuripsOverhang}
\setlength{\tabcolsep}{4pt}
% \begin{tabularx}{\dimexpr\paperwidth-12mm\relax}{
\begin{tabularx}{\dimexpr\paperwidth-12mm\relax}{
@{\hspace{2mm}}
>{\raggedright\arraybackslash}p{0.16\linewidth}
>{\raggedright\arraybackslash}p{0.11\linewidth}
>{\raggedright\arraybackslash}p{0.34\linewidth}
>{\raggedright\arraybackslash}p{0.34\linewidth}
@{\hspace{4mm}}
}
\toprule
\textbf{Strategy Type} & \textbf{Method} & \textbf{Core Mechanism} & \textbf{Key Benefit / Result} \\
\midrule

\multirow{3}{*}{Tree / Branch Search}
& TreeGRPO~\cite{ding2025treegrpo} 
& Reconstructs denoising as a search tree; Compares sibling rewards to assign step credit:
$\hat{A}_t^{\text{tree}}=\frac{\bar{r}_{\text{children}}(t)-\bar{r}_{\text{siblings}}(t)}{\mathrm{std}(r_{\text{siblings}}(t))}$ 
& 2.4$\times$ training speedup with maintained alignment quality \\
& BranchGRPO~\cite{li2025branchgrpo} 
& Flexible on-demand branching at arbitrary timesteps; Depth-wise advantage estimation; Branches adapt to reward variance 
& +16\% alignment over DanceGRPO; 55\% training time reduction; Hybrid variant 4.7$\times$ speedup \\
& Dynamic-TreeRPO~\cite{fu2025dynamic} 
& Dynamic noise intensity across tree levels (larger early, smaller late) + LayerTuning-RL (update only selected layers) 
& +4.9\% (HPS v2.1), +5.91\% (PickScore), +8.66\% (ImageReward); $\sim$50\% efficiency gain \\

\midrule

\multirow{2}{*}{\shortstack[l]{Grouping / \\Decomposition}}
& Multi-GRPO~\cite{lyu2025multi} 
& Temporal grouping (MCTS-like branching across segments) + reward grouping for multi-objective:
$\hat{A}_i=\sum_{k=1}^{K} w_k \hat{A}_i^{(k)}$ 
& Reduces interference from reward scale/variance; Improves multi-objective stability \\
& Chunk-GRPO~\cite{luo2025sample} 
& Groups consecutive steps into coherent chunks of length $k$; Chunk-level optimization with shared advantage 
& Lower-variance credit estimation while preserving temporal hierarchy \\

\midrule

\multirow{2}{*}{\shortstack[l]{Noise Causality \& \\Temporal Awareness}}
& G$^2$RPO~\cite{zhou2025fine} 
& Singular stochastic sampling: inject noise only at one random timestep $t^*$ (others ODE);
multi-granularity advantage integration across scales:
$\hat{A}_i^{\text{multi}}=\sum_s \alpha_s \hat{A}_i^{(s)}$ 
& Strengthens causal link between a specific step's stochastic decision and terminal reward \\
& TempFlow-GRPO~\cite{he2025tempflow} 
& Temporal branching to estimate step ``process reward'' via terminal reward differences;
Noise-aware weighting (explore early, exploit late) + seed-grouping by semantic similarity 
& More temporally fair comparisons and timestep-aware attribution in flow matching \\

\midrule

Theory-Guided Reweighting
& PCPO~\cite{lee2025pcpo} 
& Proportionate timestep reweighting: Credit proportional to per-step policy gradient norm 
& Strong gains over uniform-credit baselines (e.g., DanceGRPO), especially for long trajectories \\

\bottomrule
\label{tab:credit}
\end{tabularx}
\end{adjustwidth*}
\end{table*}

\subsection{Credit Assignment: From Trajectory to Step Level}
\label{sec:credit}

Credit assignment is a core challenge in reinforcement learning: given a terminal reward for a complete denoising trajectory, how can one accurately determine each intermediate step's contribution to the final outcome? Standard Flow-GRPO distributes the terminal reward uniformly across all steps, an ``egalitarian'' approach that ignores the differentiated roles of different timesteps. The methods reviewed in this section achieve more precise step-level credit assignment through various structured strategies. The summary is shown in Table~\ref{tab:credit}, and the details are as follows.

\paragraph{TreeGRPO: Tree-Structured Search.} \textbf{TreeGRPO}~\cite{ding2025treegrpo} reconstructs the denoising trajectory as a \textbf{search tree structure}, representing a pioneering contribution to credit assignment. The core idea is: at critical branching points in the denoising process, multiple child trajectories are spawned from the same parent node $x_t$, each reaching its own terminal state with an independent terminal reward. By comparing the reward differences among sibling trajectories branching from the same node, credit can be precisely attributed to the decision at the branching point. Specifically, the step-level advantage is computed as:
\begin{equation}
    \hat{A}_t^{\text{tree}} = \frac{\bar{r}_{\text{children}}(t) - \bar{r}_{\text{siblings}}(t)}{\text{std}(r_{\text{siblings}}(t))},
\end{equation}
where $\bar{r}_{\text{children}}(t)$ is the mean reward of subtrees rooted at the current node $t$, and $\bar{r}_{\text{siblings}}(t)$ is the mean reward of sibling nodes at the same level. The key advantage of this tree structure is that trajectories sharing a common prefix automatically control for confounding variables, so the advantage signal primarily reflects the quality of the decision at the branching point. Experiments show that TreeGRPO achieves \textbf{2.4$\times$ training speedup} while maintaining alignment quality, as more precise credit assignment reduces gradient variance and makes each update step more efficient.

\paragraph{Multi-GRPO: Multi-Group Advantage Estimation.} \textbf{Multi-GRPO}~\cite{lyu2025multi} extends GRPO's group normalization along two orthogonal dimensions. First, \textbf{temporal grouping}: inspired by Monte Carlo Tree Search (MCTS), branching points are introduced at different time segments of the denoising trajectory, so that trajectories within the same group share decisions in some segments and explore independently in others, enabling isolation of contributions from specific time intervals. Second, \textbf{reward grouping}: when multiple reward functions $R_1, R_2, \ldots, R_K$ are used (e.g., aesthetics score, text consistency, safety), group-normalized advantages $\hat{A}_i^{(k)}$ are independently computed for each reward function, then aggregated through weighted combination:
\begin{equation}
    \hat{A}_i = \sum_{k=1}^{K} w_k \hat{A}_i^{(k)}.
\end{equation}
This decomposition prevents mutual interference from different reward scales and variances, enabling the model to more uniformly optimize multiple objectives.

\paragraph{G$^2$RPO: Fine-Grained GRPO.} \textbf{G$^2$RPO}~\cite{zhou2025fine} addresses the weak correlation between noise and rewards in multi-step diffusion denoising, proposing two key techniques. First, \textbf{singular stochastic sampling}: at each training iteration, stochastic perturbation is injected at only a single randomly selected timestep $t^*$, with all other steps using deterministic ODE. This ensures that sampling randomness is entirely concentrated at $t^*$, establishing a strong causal link between that step's noise decision and the final reward. Second, \textbf{multi-granularity advantage integration}: advantages are estimated separately at different diffusion scales (coarse-grained global structure vs.\ fine-grained local details) and then aggregated across scales: $\hat{A}_i^{\text{multi}} = \sum_s \alpha_s \hat{A}_i^{(s)}$, where $\alpha_s$ are scale weights.

\paragraph{BranchGRPO: Structured Branching.} \textbf{BranchGRPO}~\cite{li2025branchgrpo} restructures GRPO's rollout process into a \textbf{branching tree structure}, converting sparse terminal rewards into step-level signals through ``depth-wise advantage estimation.'' Unlike TreeGRPO~\cite{ding2025treegrpo}, BranchGRPO's branching strategy is more flexible: it allows branching on demand at any timestep, with the number of branches adaptively adjusted based on the reward variance of that time interval. Experiments show that BranchGRPO improves alignment by 16\% over DanceGRPO~\cite{xue2025dancegrpo} while reducing training time by 55\%. Its hybrid variant (Hybrid BranchGRPO), which shares most of the trajectory and branches only at critical steps, achieves a further \textbf{4.7$\times$ speedup}.

\paragraph{TempFlow-GRPO: Temporal Awareness.} \textbf{TempFlow-GRPO}~\cite{he2025tempflow} focuses on temporally-aware credit assignment for flow matching models. The method uses \textbf{trajectory branching} to create divergence points at different timesteps, estimating each step's \textbf{process reward} by comparing terminal reward differences between branched trajectories. Additionally, TempFlow-GRPO introduces two flow-model-specific designs: (1) \textbf{noise-aware weighting}---assigning higher exploration weight to early steps with more noise and higher exploitation weight to later steps with less noise; and (2) \textbf{seed group strategy}---grouping initial noise samples $x_0$ by semantic similarity to ensure fairness of within-group comparisons.

\paragraph{Dynamic-TreeRPO: Dynamic Noise Tree Structure.} \textbf{Dynamic-TreeRPO}~\cite{fu2025dynamic} introduces a \textbf{dynamic noise intensity} mechanism into tree-structured search: different levels of the tree (corresponding to different timesteps) use different strengths of stochasticity injection. Early steps (high-noise regions) use larger branching randomness to cover a broader search space, while late steps (low-noise regions) use smaller randomness to refine details. Additionally, Dynamic-TreeRPO combines a \textbf{LayerTuning-RL} strategy---applying RL updates only to parameters in specific model layers while freezing the rest---to reduce parameter update volume and prevent overfitting. On the HPS-v2.1, PickScore, and ImageReward benchmarks, it surpasses SOTA by 4.9\%, 5.91\%, and 8.66\% respectively, with approximately 50\% training efficiency improvement.

\paragraph{Chunk-GRPO: Block-Level Optimization.} \textbf{Chunk-GRPO}~\cite{luo2025sample} observes that the denoising process of flow matching models exhibits natural \textbf{temporal coherence}: decisions at adjacent timesteps are highly correlated, and single-step credit assignment may introduce excessive noise. Based on this observation, Chunk-GRPO groups consecutive denoising steps into several coherent ``chunks,'' each containing $k$ consecutive steps. Policy optimization operates at the chunk level rather than the individual step level, with steps within a chunk sharing the same advantage signal. This design captures the intrinsic temporal dynamics hierarchy of flow matching, reducing advantage estimation variance while maintaining sufficient temporal resolution.

\paragraph{PCPO: Proportionate Credit Policy Optimization.} \textbf{PCPO}~\cite{lee2025pcpo} addresses the credit assignment problem from a theoretical perspective. The method proves that in flow matching RL, the proportional contribution of different timesteps to the final reward should be proportional to the policy gradient norm at that step. Accordingly, PCPO uses a \textbf{principled timestep reweighting} mechanism to ensure that the gradient weight each step receives in the policy update strictly reflects its proportional contribution to the reward. Both theoretical analysis and experiments demonstrate that PCPO significantly outperforms uniformly-credited policy gradient baselines such as DanceGRPO, with the advantage being particularly pronounced in long-trajectory (many-step denoising) scenarios.

% ============================================================================
\begin{table*}[!h]
\centering
\small
\caption{Sampling efficiency and training acceleration strategies for Flow-GRPO. 
These methods reduce rollout cost via entropy-aware pruning, sliding windows, 
noise distribution optimization, forward-process training, 
distribution-level matching, and preference-based learning.}
\setlength{\neuripsOverhang}{\dimexpr(\paperwidth-\textwidth)/2 -6mm\relax}
\begin{adjustwidth*}{-\neuripsOverhang}{-\neuripsOverhang}
\setlength{\tabcolsep}{4pt}
% \begin{tabularx}{\dimexpr\paperwidth-12mm\relax}{
\begin{tabularx}{\dimexpr\paperwidth-12mm\relax}{
@{\hspace{2mm}}
>{\raggedright\arraybackslash}p{0.16\linewidth}
>{\raggedright\arraybackslash}p{0.11\linewidth}
>{\raggedright\arraybackslash}p{0.34\linewidth}
>{\raggedright\arraybackslash}p{0.34\linewidth}
@{\hspace{4mm}}
}
\toprule
\textbf{Acceleration Type} & \textbf{Method} & \textbf{Core Mechanism} & \textbf{Speedup / Key Result} \\
\midrule

\multirow{2}{*}{\shortstack[l]{Entropy / \\Temporal Pruning}}
& E-GRPO~\cite{zhang2026grpo}
& Identifies high-entropy timesteps via $H(t)$; Merges low-entropy steps; 
Group-normalized advantages within merged segments 
& Reduced SDE computation; Improved signal-to-noise ratio \\
& MixGRPO~\cite{li2025mixgrpo} 
& Sliding ODE–SDE window $[t_w, t_w+W]$; SDE only inside window; ODE elsewhere 
& MixGRPO-Flash: 71\% time reduction with maintained alignment \\

\midrule

\multirow{2}{*}{\shortstack[l]{Sampling Distribution \\Optimization}}
& Smart-GRPO~\cite{yu2025smart} 
& Iteratively reweights initial noise distribution based on reward feedback 
& More efficient exploration; Avoids low-reward regions \\
& Pro-GRPO~\cite{ge2025expand} 
& Expand-and-Prune: Increase $G'$, diversity pruning, multi-step OVF 
& Mitigates reward clustering; Higher signal variance utilization \\

\midrule

\multirow{3}{*}{\shortstack[l]{Forward / \\Matching-Based RL}}
& DiffusionNFT~\cite{zheng2025diffusionnft} 
& Online RL on forward noising process; Contrastive positive/negative flow matching 
& 25$\times$ speedup; GenEval 0.24$\rightarrow$0.98 in 1k steps \\
& AWM~\cite{xue2025advantage} 
& Advantage-weighted flow matching:
$\mathcal{L}_{AWM}=\hat{A}_i \|v_\theta - \text{target}\|^2$ 
& 24$\times$ speedup; No SDE rollouts; Simplified training \\
& DGPO~\cite{luo2025reinforcing} 
& Direct group preference optimization (DPO-style); Deterministic ODE sampling 
& $\sim$20$\times$ faster training; Removes policy gradient overhead \\

\midrule

\multirow{1}{*}{Design Space Analysis}
& GRPO-ELBO-ODE~\cite{choi2026rethinking}
& Factorized RL design space: Objective, likelihood estimator, sampler; 
ELBO shown dominant 
& 4.6$\times$ efficiency gain; GenEval 0.24$\rightarrow$0.95 (90 GPU hours) \\

\bottomrule
\label{tab:efficiency}
\end{tabularx}
\end{adjustwidth*}
\end{table*}

\subsection{Sampling Efficiency and Training Acceleration}
\label{sec:efficiency}
The primary computational bottleneck of Flow-GRPO is that each policy update requires generating $G$ complete denoising trajectories (rollouts), each containing $T$ denoising steps, with each step requiring a full model forward pass. For large-scale models (such as SD3.5-Medium with 2B+ parameters), the cost of a single training iteration is extremely high. The methods reviewed in this section significantly reduce training cost from different angles, achieving up to 25$\times$ speedup. The summary is shown in Table~\ref{tab:efficiency}, and the details are as follows.

\paragraph{E-GRPO: High-Entropy Step Driven.} \textbf{E-GRPO}~\cite{zhang2026grpo} observes that not all steps in a denoising trajectory have equal influence on the final result---most steps' decisions are nearly deterministic (low entropy), with only a few steps exhibiting genuine uncertainty (high entropy). E-GRPO computes the policy entropy at each timestep $H(t) = -\mathbb{E}[\log \pi_\theta(a_t|s_t)]$, merges consecutive low-entropy steps into equivalent high-entropy SDE steps for optimization, and uses efficient deterministic ODE sampling for the remaining steps. Additionally, E-GRPO employs \textbf{multi-step group-normalized advantages}---normalizing within merged high-entropy step groups rather than across the full trajectory---to improve the signal-to-noise ratio.

\paragraph{MixGRPO: Mixed ODE-SDE Sliding Window.}
\textbf{MixGRPO}~\cite{li2025mixgrpo} proposes an elegant \textbf{sliding window} strategy: SDE + GRPO policy optimization is applied only within a moving window $[t_w, t_w + W]$, while regions outside the window use deterministic ODE with high-order solvers (such as DPM-Solver++~\cite{lu2025dpm}) for efficient sampling. The window slides along the time axis during training, ensuring all timesteps are eventually covered. The core insight is that at any given training step, the model only needs to learn improvements within the current window, while steps outside the window can ``freely'' use the already-learned policy. Its accelerated variant \textbf{MixGRPO-Flash} further narrows the window width and uses more aggressive ODE step sizes, reducing training time by \textbf{71\%} while maintaining alignment quality comparable to full SDE training.

\paragraph{Smart-GRPO: Intelligent Noise Sampling.}
\textbf{Smart-GRPO}~\cite{yu2025smart} is the first method to directly optimize the \textbf{initial noise distribution} in flow matching RL. Standard approaches uniformly sample initial noise from $x_0 \sim \mathcal{N}(0, I)$, but most initial noise samples may lead to low-reward regions, wasting computation. Smart-GRPO refines the noise sampling distribution through an iterative search process: after each training round, noise sampling weights are updated based on final reward feedback for each initial noise, biasing subsequent sampling toward high-reward regions. This approach does not modify the denoising process itself but optimizes the starting point of exploration.

\paragraph{Pro-GRPO: Expand and Prune.}
\textbf{Pro-GRPO}~\cite{ge2025expand} discovers that Flow-GRPO training suffers from severe \textbf{reward clustering}: the $G$ samples for a given prompt tend to cluster in a narrow region of reward space, resulting in extremely small advantage signal variance after group normalization, rendering policy gradients near zero. To address this, Pro-GRPO proposes an ``Expand-and-Prune'' three-stage strategy: (1) \textbf{Expand}---increase the initial sampling count to $G' \gg G$ to cover a broader reward space; (2) \textbf{Prune}---prune samples based on latent-space feature diversity metrics, retaining the most representative $G$ samples; (3) \textbf{Multi-step OVF (Online Value Function)}---execute multi-step policy optimization on retained samples to maximize information utilization from each sample.

\paragraph{DiffusionNFT: Forward-Process Online RL.}
\textbf{DiffusionNFT} (Noise-Free Training)~\cite{zheng2025diffusionnft} proposes a fundamentally different training paradigm: rather than optimizing on the reverse denoising process, it performs online RL directly on the \textbf{forward noising process} via the flow matching loss. Specifically, DiffusionNFT first generates a batch of complete samples, divides them into positive (high-reward) and negative (low-reward) examples, then uses a contrastive learning objective---increasing the flow matching likelihood of positive examples and decreasing that of negative examples---to update the model. By avoiding complete SDE rollouts and per-step policy gradient computation, DiffusionNFT is \textbf{25$\times$ more efficient} than Flow-GRPO. Within only 1,000 training steps, it improves SD3.5-Medium's GenEval score from 0.24 to 0.98, demonstrating remarkable sample efficiency.

\paragraph{AWM: Advantage Weighted Matching.}
\textbf{AWM}~\cite{xue2025advantage} establishes a theoretical connection between DDPO~\cite{black2023training} and flow matching~\cite{lipman2022flow}. The authors prove that DDPO's policy gradient objective is mathematically equivalent to an implicit score/flow matching loss with noisy targets. Based on this insight, AWM proposes directly using the \textbf{same flow matching loss as pretraining}, but reweighted by sample-level advantage $\hat{A}_i$:
\begin{equation}
    \mathcal{L}_{\text{AWM}} = \mathbb{E}_{i, t} \left[ \hat{A}_i \cdot \|v_\theta(x_t^{(i)}, t) - (x_1^{(i)} - x_0^{(i)})\|^2 \right].
\end{equation}
This design completely avoids SDE rollouts, per-step log-likelihood computation, and importance ratio clipping, simplifying the training process to ``weighted fine-tuning.'' On SD 3.5 Medium and FLUX, AWM achieves a \textbf{24$\times$ speedup} over Flow-GRPO while reaching comparable or superior alignment quality.

\paragraph{DGPO: Direct Group Preference Optimization.}
\textbf{DGPO}~\cite{luo2025reinforcing} takes an even more radical approach: \textbf{completely abandoning the policy gradient framework}. DGPO does not compute importance ratios, does not clip, and does not perform per-step gradient backpropagation. Instead, it learns directly from group-level preference relations---for each group of $G$ samples, DGPO constructs pairwise preference comparisons, then uses a DPO-like objective to directly optimize the flow matching model's parameters. The key advantage of this design is that it unlocks \textbf{deterministic ODE samplers}---no longer requiring SDE stochasticity to define the policy distribution. Since ODE sampling is several times faster than SDE and supports high-order solvers, DGPO's overall training speed is approximately \textbf{20$\times$} that of existing methods.

\paragraph{Systematic Analysis of the RL Design Space.}
\textbf{GRPO-ELBO-ODE}~\cite{choi2026rethinking} conduct the first systematic ablation study of the design space for RL fine-tuning of diffusion/flow matching models. They decompose the design space into three orthogonal factors: (1) \textbf{policy gradient objective} (REINFORCE~\cite{sutton1999policy}, PPO-clip~\cite{schulman2017proximal}, GRPO~\cite{shao2024deepseekmath}, etc.); (2) \textbf{likelihood estimator} (per-step log-likelihood, ELBO, score matching, etc.); (3) \textbf{rollout sampling scheme} (full SDE, mixed ODE-SDE, pure ODE, etc.). Through exhaustive experiments, they reach a surprising conclusion: \textbf{the ELBO likelihood estimator is the overwhelmingly dominant factor}---configurations using ELBO significantly outperform per-step log-likelihood under nearly all policy gradient objective and sampling scheme combinations. Based on this finding, they achieve GenEval improvement from 0.24 to 0.95 on SD 3.5 Medium in only 90 GPU hours, \textbf{4.6$\times$} more efficient than Flow-GRPO.

% ============================================================================

\begin{table*}[!h]
\centering
\small
\caption{Mode collapse mitigation and diversity preservation strategies in GRPO / Flow-GRPO alignment. 
These methods address reward-induced preference collapse via distribution-level regularization, 
orthogonal perturbation, reward reshaping, directional decoupling, and identity-aware diversity optimization.}
\setlength{\neuripsOverhang}{\dimexpr(\paperwidth-\textwidth)/2 -6mm\relax}
\begin{adjustwidth*}{-\neuripsOverhang}{-\neuripsOverhang}
\setlength{\tabcolsep}{4pt}
% \begin{tabularx}{\dimexpr\paperwidth-12mm\relax}{
\begin{tabularx}{\dimexpr\paperwidth-12mm\relax}{
@{\hspace{2mm}}
>{\raggedright\arraybackslash}p{0.16\linewidth}
>{\raggedright\arraybackslash}p{0.11\linewidth}
>{\raggedright\arraybackslash}p{0.34\linewidth}
>{\raggedright\arraybackslash}p{0.34\linewidth}
@{\hspace{4mm}}
}
\toprule
\textbf{Category} & \textbf{Method} & \textbf{Core Idea} & \textbf{Key Results} \\
\midrule

\multirow{2}{*}{\shortstack[l]{Distribution-Level \\Regularization}}
& DiverseGRPO~\cite{liu2025diversegrpo} 
& Spectral clustering in CLIP space; Exploration reward 
$r_i^{\text{div}} = r_i + \lambda / |C_i|$; 
Early-step stronger KL to prevent structural collapse 
& +13\%–18\% Vendi diversity 
without quality drop \\
& DRIFT~\cite{liu2026beyond} 
& Reward-concentrated sampling; Stochastic prompt perturbation; 
Potential-based shaping $\Phi(s_{t+1})-\gamma\Phi(s_t)$ 
& +9–43\% diversity at equal alignment; 
Pareto-dominant trade-off \\

\midrule

\multirow{1}{*}{Orthogonal Perturbation}
& OSCAR~\cite{wu2025oscar} 
& Inject stochastic noise orthogonal to generation flow in latent space; 
Monotonic increase of distribution volume proxy 
& Training-free diversity boost; 
No alignment degradation \\

\midrule

\multirow{1}{*}{\shortstack[l]{Bias Direction \\Decoupling}}
& D$^2$-Align~\cite{chen2025taming} 
& Identify bias direction in reward embedding space; 
Remove preference bias via lightweight correction module; 
Introduces DivGenBench benchmark 
& Significant mitigation of preference mode collapse \\

\midrule

\multirow{1}{*}{Identity Diversity}
& DisCo~\cite{borse2026resolvingidentitycrisistexttoimage} 
& RL with compositional reward; 
Pairwise facial similarity penalty across subjects 
& 98.6\% Unique Face Accuracy 
(vs.\ <50\% baseline) \\

\bottomrule
\label{tab:diversity}

\end{tabularx}
\end{adjustwidth*}
\end{table*}

\subsection{Mode Collapse and Diversity Preservation}
\label{sec:diversity}
GRPO~\cite{shao2024deepseekmath}/Flow-GRPO~\cite{liu2025flow} alignment training faces a fundamental tension: reward models typically exhibit \textbf{inherent preference biases}---favoring certain specific styles, compositions, or color schemes with higher scores. When models are repeatedly optimized to maximize these preference scores, their outputs gradually converge to the narrow modes preferred by the reward model, causing \textbf{mode collapse} or \textbf{preference mode collapse}. The methods in this section address this problem from the perspectives of distributional regularization, orthogonal perturbation, and reward reshaping. The summary is shown in Table~\ref{tab:diversity}, and the details are as follows.

\paragraph{DiverseGRPO: Distributional Creativity Rewards.}
\textbf{DiverseGRPO}~\cite{liu2025diversegrpo} is the first method to design diversity rewards from a \textbf{distributional level} (rather than sample level). Standard GRPO computes rewards and advantages only at the sample level, completely ignoring the overall distributional characteristics of the generated sample set for a given prompt. DiverseGRPO introduces two key techniques:
First, \textbf{semantically-grouped distributional representation}: for the $G$ generated samples of a given prompt, spectral clustering is performed in CLIP embedding space, dividing samples into $M$ semantic clusters $C_1, C_2, \ldots, C_M$. An \textbf{exploration reward} inversely proportional to cluster size is then added to each sample: $r_i^{\text{diverse}} = r_i + \lambda / |C_i|$. This means samples falling into rare clusters (small $|C_i|$) receive higher reward bonuses, incentivizing the model to explore underrepresented semantic regions.
Second, \textbf{structure-aware early regularization}: stronger KL divergence regularization is applied during early denoising steps (the phase determining global structure and layout) to prevent premature policy collapse, while regularization is relaxed during late steps (the detail refinement phase) to allow quality improvement.
Experiments demonstrate that DiverseGRPO improves semantic diversity (Vendi Score) by 13\%--18\% while matching quality.

\paragraph{OSCAR: Orthogonal Stochastic Control.}
\textbf{OSCAR} (Orthogonal Stochastic Control for Alignment Rewards)~\cite{wu2025oscar} proposes an elegant training-free diversity enhancement method from a differential geometry perspective. The core observation is: in the latent space of flow matching, variations along the generation flow direction affect image quality/alignment, while variations \textbf{orthogonal to the generation flow} affect only diversity without changing quality. OSCAR accordingly injects \textbf{stochastic perturbations orthogonal to the generation flow direction} during denoising, operating in feature space rather than pixel space. Theoretical analysis proves that this perturbation monotonically increases a volume proxy metric (i.e., the coverage of the generated distribution) without degrading alignment quality.

\paragraph{DRIFT: Beyond the Dirac Delta.}
\textbf{DRIFT}~\cite{liu2026beyond}---whose name hints at its goal of broadening the Dirac-delta-like narrow distributions of RL-aligned model outputs to cover a wider range of modes---proposes three complementary mechanisms: (1) \textbf{reward-concentrated sampling}---concentrating sampling in high-density regions of reward space rather than extreme-value regions, avoiding domination by a few extremely high-scoring samples; (2) \textbf{stochastic prompt variation}---applying slight random perturbations to text prompts during training to prevent the model from overfitting to specific prompt formats; (3) \textbf{potential-based reward shaping}---overlaying a shaping term based on a potential function $\Phi(s_{t+1}) - \gamma\Phi(s_t)$ on the original reward, where the potential function reflects state space coverage. The three mechanisms work synergistically, achieving \textbf{9--43\% diversity improvement} at equivalent alignment, and 60--66\% alignment improvement at equivalent diversity, exhibiting Pareto dominance.

\paragraph{Taming Preference Mode Collapse.}
\textbf{D$^2$-Align}~\cite{chen2025taming} proposes a novel \textbf{directional decoupling} method to combat preference mode collapse. The core insight is that reward model biases can be represented in its embedding space as a \textbf{bias direction vector}---all high-scoring samples tend to cluster along this direction in embedding space. D$^2$-Align learns a lightweight directional correction module that, without modifying the reward model itself, removes the bias direction component from the embedding space, so that the corrected reward signal reflects only genuine quality rather than stylistic preferences. The method also proposes \textbf{DivGenBench}---the first benchmark specifically designed to evaluate preference mode collapse.

\paragraph{DisCo: Multi-Human Identity Diversity.}
\textbf{DisCo} (Diversity in Social Context)~\cite{borse2026resolvingidentitycrisistexttoimage} addresses a specific but important diversity problem: \textbf{identity diversity} in multi-human T2I generation. When prompts request generating multiple people, existing models tend to produce characters with highly similar facial features (the ``clone face'' problem). DisCo is the first framework to directly optimize identity diversity via RL, designing a \textbf{compositional reward function} that performs pairwise facial feature comparison across all faces in the generated image, penalizing excessive facial similarity. On the DiverseHumans benchmark, DisCo achieves \textbf{98.6\% Unique Face Accuracy}, a qualitative leap from the baseline's below 50\%.

% ============================================================================
\begin{table*}[!h]
\centering
\small
\caption{Reward hacking mitigation strategies in GRPO / Flow-GRPO alignment. 
These methods address reward exploitation via regulated clipping, 
uncertainty-aware penalization, data anchoring, sampler redesign, 
artifact detection, reward model robustness, and offline correction.}
\setlength{\neuripsOverhang}{\dimexpr(\paperwidth-\textwidth)/2 -6mm\relax}
\begin{adjustwidth*}{-\neuripsOverhang}{-\neuripsOverhang}
\setlength{\tabcolsep}{4pt}
% \begin{tabularx}{\dimexpr\paperwidth-12mm\relax}{
\begin{tabularx}{\dimexpr\paperwidth-12mm\relax}{
@{\hspace{2mm}}
>{\raggedright\arraybackslash}p{0.16\linewidth}
>{\raggedright\arraybackslash}p{0.11\linewidth}
>{\raggedright\arraybackslash}p{0.34\linewidth}
>{\raggedright\arraybackslash}p{0.34\linewidth}
@{\hspace{4mm}}
}
\toprule
\textbf{Mitigation Strategy} & \textbf{Method} & \textbf{Core Idea} & \textbf{Key Result / Contribution} \\
\midrule

\multirow{1}{*}{Clipping Stabilization}
& GRPO-Guard~\cite{wang2025grpo} 
& Ratio normalization $\tilde{\rho}_i=\rho_i/\bar{\rho}$; 
Gradient reweighting for asymmetric importance shift 
& Stabilizes PPO clipping; Prevents ratio collapse \\

\midrule

\multirow{2}{*}{\shortstack[l]{Uncertainty \& \\Diversity Regularization}}
& GARDO~\cite{he2025gardo} 
& Penalize high-uncertainty reward outputs; 
EMA reference update; Diversity-amplified reward 
& Reduces blind-spot exploitation; Improves stability \\
& ConsistentRFT~\cite{tan2026consistentrft} 
& Multi-granularity rollout (DGR); 
Consistent policy gradient optimization (CPGO) 
& -49\% perceptual hallucination; 
-38\% semantic hallucination \\

\midrule

\multirow{1}{*}{Data Anchoring}
& DDRL~\cite{ye2025data} 
& Forward KL regularization 
$D_{\text{KL}}(p_{\text{data}} \| \pi_\theta)$ 
& Industrial-scale validation; 
Prevents forgetting data modes \\

\midrule

\multirow{2}{*}{\shortstack[l]{Reward Model Analysis \\ \& Robustification}}
& ArtifactReward~\cite{hong2026understanding}
& Typology of hacking (over-stylization vs.\ deformation); 
Artifact-aware auxiliary reward 
& First systematic categorization of T2I reward hacking \\
& SoliReward~\cite{lian2025solireward} 
& Single-item binary labeling; Cross-prompt pairing; 
Hierarchical progressive query attention 
& Robust video reward modeling under noisy labels \\

\midrule

\multirow{1}{*}{Sampler Redesign}
& CPS~\cite{wang2025coefficients} 
& Coefficients-preserving SDE sampler; 
Noise–signal conservation constraints 
& Reduces stochastic artifact exploitation; 
More stable convergence \\

\midrule

\multirow{1}{*}{Offline Correction}
& GDRO~\cite{wang2026gdro} 
& Fully offline training; Corrected evaluation score via importance sampling 
& Avoids online hacking amplification; 
Stable rectified flow alignment \\

\bottomrule
\label{tab:hacking}

\end{tabularx}
\end{adjustwidth*}
\end{table*}

\subsection{Reward Hacking Mitigation}
\label{sec:hacking}
\textbf{Reward hacking} refers to models learning to exploit reward model vulnerabilities or biases to obtain high scores without genuinely improving generation quality. In RL alignment of diffusion/flow matching models, this manifests as: reward scores continue rising during late training while actual generation quality degrades---artifacts appear, colors become oversaturated, and style becomes monotonous. This section reviews various mitigation strategies. The summary is shown in Table~\ref{tab:hacking}, and the details are as follows.

\paragraph{GRPO-Guard: Regulated Clipping.}
\textbf{GRPO-Guard}~\cite{wang2025grpo} first systematically diagnoses the failure of PPO's clipping mechanism in Flow-GRPO. By analyzing the distributional changes of the importance ratio $\rho_i = \pi_\theta(o_i|c)/\pi_{\text{ref}}(o_i|c)$ throughout training, the authors discover that as training progresses, the distribution of $\rho_i$ exhibits a \textbf{systematic left-shift}---the mean persistently falls below 1. This means the probability of most model outputs decreases relative to the reference policy, while PPO clipping is designed only for the symmetric interval $[1-\epsilon, 1+\epsilon]$ and cannot effectively handle this asymmetric shift. GRPO-Guard proposes two fixes: (1) \textbf{ratio normalization}---normalizing $\rho_i$ to $\tilde{\rho}_i = \rho_i / \bar{\rho}$, re-centering the clipping interval at the distribution mean; (2) \textbf{gradient reweighting}---adjusting gradient weights based on the degree of $\rho_i$'s deviation from 1, preventing extreme ratios from dominating update directions.

\paragraph{GARDO: Gated Adaptive Regularization.}
\textbf{GARDO} (Gated Adaptive Regularized Diffusion Optimization)~\cite{he2025gardo} mitigates reward hacking from an \textbf{uncertainty-aware} perspective. The core idea is that reward hacking often occurs in input regions where the reward model has high uncertainty---the model learns to generate samples in these ``blind spots'' to obtain inflated scores. GARDO introduces three techniques: (1) \textbf{selective penalization}---imposing additional penalties on samples with high reward model output uncertainty, discouraging the model from drifting toward uncertain regions; (2) \textbf{adaptive reference model update}---progressively updating the reference model $\pi_{\text{ref}}$ via exponential moving average rather than using a fixed reference, allowing the KL divergence constraint to dynamically adjust with training; (3) \textbf{diversity-amplified reward}---overlaying a reward term reflecting within-batch diversity onto the original reward, constraining model behavior along the diversity dimension.

\paragraph{DDRL: Data-Regularized RL.}
\textbf{DDRL} (Data-regularized Diffusion RL)~\cite{ye2025data} proposes a reward hacking mitigation approach validated both theoretically and at industrial scale. The core idea is to anchor the policy to the offline data distribution using \textbf{forward KL divergence}:
\begin{equation}
    \mathcal{L}_{\text{DDRL}} = \mathcal{L}_{\text{RL}} + \alpha \cdot D_{\text{KL}}(p_{\text{data}} \| \pi_\theta).
\end{equation}
Unlike the standard reverse KL ($D_{\text{KL}}(\pi_\theta \| \pi_{\text{ref}})$), forward KL penalizes the policy for assigning low probability in high-density regions of the data distribution, preventing the model from ``forgetting'' common patterns in real data. DDRL has been validated in \textbf{over 1 million GPU hours} of industrial-scale experiments and \textbf{10,000 human evaluations}, demonstrating that data regularization is critical for preventing reward hacking in large-scale training.

\paragraph{ConsistentRFT: Reducing Visual Hallucinations.}
\textbf{ConsistentRFT}~\cite{tan2026consistentrft} treats reward hacking as a form of \textbf{visual hallucination}: the model generates content that appears plausible but is semantically inconsistent with the prompt. The method proposes two techniques: (1) \textbf{Dynamic Granularity Rollout (DGR)}---simultaneously using rollouts at different resolutions/granularities within a single training iteration, with coarse granularity capturing global semantics and fine granularity focusing on local details, complementing each other to avoid blind spots of any single granularity; (2) \textbf{Consistent Policy Gradient Optimization (CPGO)}---enforcing consistency of policy update directions across steps in the multi-step denoising process, preventing contradictory gradients from different steps from causing instability. Experiments show that ConsistentRFT reduces low-level perceptual hallucinations (e.g., artifacts, deformations) by 49\% and high-level semantic hallucinations (e.g., missing objects, incorrect attributes) by 38\%.

\paragraph{Understanding Reward Hacking in T2I RL.}\textbf{ArtifactReward}~\cite{hong2026understanding} conduct the first systematic typological study of reward hacking. Through separate analysis of aesthetic/preference rewards (such as HPS, PickScore) and prompt consistency rewards (such as CLIP Score, VQA-based), they find that: aesthetic reward hacking manifests as \textbf{over-stylization} (e.g., oversaturated colors, extreme contrast), while consistency reward hacking manifests as \textbf{object deformation} (e.g., distorting text to improve OCR scores). Based on these findings, the authors propose a \textbf{lightweight adaptive artifact reward model} as a regularizer, specifically designed to detect common artifact types in the RL alignment process.

\paragraph{CPS: Coefficients-Preserving Sampling.}
\textbf{CPS}~\cite{wang2025coefficients} traces the root cause of reward hacking to \textbf{sampler design}. Through theoretical analysis, the authors prove that excessive stochasticity in SDE sampling introduces noise artifacts in generated images, which may be misinterpreted by reward models as beneficial features (e.g., ``texture detail''), leading to reward hacking. Inspired by DDIM~\cite{song2020denoising}, CPS designs a \textbf{coefficients-preserving sampler} that eliminates artifact-producing excessive stochasticity while preserving the randomness needed for SDE exploration. Specifically, CPS ensures that signal and noise coefficients in the sampling process satisfy specific conservation relations, thereby guaranteeing sampling quality. Experiments show CPS achieves more accurate reward modeling and more stable training convergence.

\paragraph{SoliReward: Robust Video Reward Model.}
\textbf{SoliReward}~\cite{lian2025solireward} prevents reward hacking from the perspective of \textbf{reward model training}. Training data for video reward models typically contains substantial labeling noise (different annotators having inconsistent preferences for the same video pair), which causes reward models to learn superficial shortcuts. SoliReward proposes two techniques to combat labeling noise: (1) \textbf{single-item binary annotation}---converting traditional pairwise comparison labels into independent quality assessments of individual videos, reducing the subjective bias of pairwise labeling; (2) \textbf{cross-prompt pairing}---pairing videos from different prompts to construct preference pairs, forcing the reward model to learn universal quality features rather than prompt-specific surface patterns. Additionally, \textbf{Hierarchical Progressive Query Attention} enhances the spatiotemporal consistency of video feature aggregation.

\paragraph{GDRO: Offline Group-Level Direct Reward Optimization.}
\textbf{GDRO}~\cite{wang2026gdro} proposes a \textbf{fully offline} training method, fundamentally circumventing the amplification of reward hacking in online rollouts. Designed for rectified flow diffusion models, GDRO trains on pre-generated offline datasets without requiring online image rollouts. To address the distributional shift problem of reward scores in offline training, GDRO introduces a \textbf{corrected evaluation score}, using importance sampling weighting to correct the offline data's reward distribution to the expected distribution under the current policy.

% ============================================================================
\begin{table*}[!h]
\centering
\small
\caption{ODE–SDE sampling spectrum in Flow-GRPO alignment. 
Methods explore different stochasticity regimes ranging from pure ODE (zero noise) 
to structured SDE and amortized stochastic flow maps, with corresponding theoretical insights.}
\setlength{\neuripsOverhang}{\dimexpr(\paperwidth-\textwidth)/2 -6mm\relax}
\begin{adjustwidth*}{-\neuripsOverhang}{-\neuripsOverhang}
\setlength{\tabcolsep}{4pt}
% \begin{tabularx}{\dimexpr\paperwidth-12mm\relax}{
\begin{tabularx}{\dimexpr\paperwidth-12mm\relax}{
@{\hspace{2mm}}
>{\raggedright\arraybackslash}p{0.16\linewidth}
>{\raggedright\arraybackslash}p{0.11\linewidth}
>{\raggedright\arraybackslash}p{0.34\linewidth}
>{\raggedright\arraybackslash}p{0.34\linewidth}
@{\hspace{4mm}}
}
\toprule
\textbf{Stochasticity Regime} & \textbf{Method} & \textbf{Core Idea} & \textbf{Key Contribution} \\
\midrule

\multirow{1}{*}{Pure ODE}
& Neighbor GRPO~\cite{he2025neighbor} 
& Deterministic ODE sampling; Diversity from perturbed initial noise 
$x_0^{(i)} = x_0 + \delta_i$; Softmax distance proxy for policy gradient 
& Efficient; No noise artifacts; ODE-compatible high-order solvers \\

\midrule

Few-Step Annealed Noise
& TAFS-GRPO~\cite{yue2026know} 
& Adaptive temporal noise injection with annealed temperature 
$\tau(k)$; Step-aware advantage weighting 
& Efficient RL alignment for 1--4 step generation \\

\midrule

\multirow{2}{*}{\shortstack[l]{Structured \\ Stochastic Flows}}
& GLASS Flows~\cite{holderrieth2025glass} 
& Transition kernel $p(x_{t+\Delta t}|x_t)$ modeled via auxiliary flow matching; 
Structured stochasticity instead of Gaussian noise; Feynman-Kac steering 
& Improved inference-time scaling; Coherent exploration \\
& Diamond Maps~\cite{holderrieth2026diamond} 
& Amortized stochastic flow map $(x_0, z)\mapsto x_1$; 
Distilled from GLASS Flows 
& Single-step stochastic sampling; Reward-adaptive inference \\

\midrule

\multirow{1}{*}{Theoretical Analysis}
& SDE–ODE Reward Gap~\cite{sheng2025understanding}
& Reward gap bounds under gDDIM; 
High-stochasticity training improves ODE inference; 
Optimal stochasticity level characterization 
& First formal theory of SDE–ODE reward gap \\

\bottomrule
\label{tab:sampling}

\end{tabularx}
\end{adjustwidth*}
\end{table*}

\subsection{ODE vs.\ SDE: Sampling Strategies}
\label{sec:sampling}
Flow-GRPO's conversion from ODE to SDE introduces a fundamental design choice: how much stochasticity should the sampling process use? SDE (high stochasticity) provides better exploration but introduces noise artifacts and additional computation; ODE (zero stochasticity) is efficient and quality-stable but lacks exploratory diversity. This section reviews different design choices across this spectrum and their theoretical underpinnings. The summary is shown in Table~\ref{tab:sampling}, and the details are as follows.

\paragraph{Neighbor GRPO: Pure ODE Contrastive Optimization.}
\textbf{Neighbor GRPO}~\cite{he2025neighbor} completely \textbf{bypasses SDE}, achieving policy optimization within a pure ODE sampling framework. The core idea is: diversity need not come from SDE's per-step stochasticity, but can instead come from \textbf{variations in initial noise}. Specifically, for a given prompt $c$, Neighbor GRPO generates a set of candidate trajectories by sampling multiple perturbed versions within the neighborhood of the initial noise $x_0$: $x_0^{(i)} = x_0 + \delta_i$ (where $\delta_i$ are small Gaussian perturbations). Since each trajectory uses deterministic ODE sampling, the entire process is efficient and free of noise artifacts. Policy optimization is achieved through a \textbf{softmax distance proxy}---rather than directly computing SDE log-likelihoods, it uses distance relations between terminal samples of candidate trajectories to construct soft labels. The authors prove the theoretical connection between this proxy and the true policy gradient. Neighbor GRPO preserves all advantages of ODE: efficient sampling, high-order solver compatibility, and no noise artifacts.

\paragraph{TAFS-GRPO: Temperature-Annealed Few-Step Sampling.}
\textbf{TAFS-GRPO}~\cite{yue2026know} focuses on \textbf{few-step generation} scenarios (e.g., 1--4 steps), where standard SDE multi-step stochasticity injection is infeasible. TAFS-GRPO's strategy is: \textbf{iteratively inject adaptive temporal noise} on single-step or few-step generated samples. Specifically, given a single-step deterministic sample $\hat{x}_1$, TAFS-GRPO adds noise with an annealing temperature $\tau(k)$ (decreasing with iteration $k$) to $\hat{x}_1$ then re-denoises, generating diversified candidate samples. Combined with \textbf{step-aware advantage}---adjusting advantage weights based on the timestep and strength of injected noise---it achieves efficient alignment for few-step T2I generation.

\paragraph{GLASS Flows: Transition Sampling.}
\textbf{GLASS Flows} (Generalized LAyer-wise Stochastic Sampling for Flows)~\cite{holderrieth2025glass} introduces an entirely new sampling paradigm: \textbf{``flow matching within flow matching.''}  Standard SDE injects independent Gaussian noise at each discrete timestep, but this independent noise may disrupt the coherent structure learned by flow matching. The core idea of GLASS Flows is: the Markov transition $p(x_{t+\Delta t} | x_t)$ between discrete timesteps is itself modeled as a small flow matching problem---using an auxiliary flow matching model to sample the conditional distribution $p(x_{t+\Delta t} | x_t)$ rather than simple Gaussian noise. This maintains ODE's structural efficiency while introducing more structured stochasticity than simple Gaussian noise. Combined with \textbf{Feynman-Kac Steering}~\cite{singhal2025general}---using importance sampling to bias the transition kernel toward high-reward regions---GLASS Flows achieves state-of-the-art inference-time scaling performance.

\paragraph{Diamond Maps: Stochastic Flow Maps.}
\textbf{Diamond Maps}~\cite{holderrieth2026diamond} goes further than GLASS Flows by \textbf{amortizing the multi-step simulation into a single-step stochastic flow map sampler}. Specifically, Diamond Maps learns a mapping $(x_0, z) \mapsto x_1$ (where $z$ is an auxiliary random variable) such that the output distribution of this single-step mapping matches the output distribution of the complete multi-step GLASS Flow. Learning is achieved through GLASS Flows distillation. The key advantage is: at inference time, only a single forward pass is needed to obtain a stochastic generated sample, while preserving the exploration capability needed for alignment. This enables Diamond Maps to \textbf{rapidly adapt to arbitrary reward functions at inference time with minimal cost}---simply by changing the sampling strategy for the random variable $z$, generation can be steered toward different reward directions.

\paragraph{Theoretical Analysis of Sampler Stochasticity.}
\textbf{SDE–ODE Reward Gap}~\cite{sheng2025understanding} provide the first rigorous \textbf{theoretical characterization} of the reward gap between SDE training and ODE inference. Specific contributions include: (1) proving that under the gDDIM (generalized DDIM)~\cite{zhang2022gddim} framework, arbitrary different stochasticity levels can be used at training and inference time, with corresponding reward gap upper bounds; (2) proving a counterintuitive result: \textbf{higher-stochasticity SDE training can actually improve the quality of low/zero-stochasticity ODE inference}---intuitively, high-stochasticity SDE explores a broader policy space during training, enabling the learned policy to select high-quality deterministic paths during ODE inference; (3) providing theoretical guidance on the optimal training stochasticity level---it should not be increased indefinitely but has an optimal point related to model capacity and reward complexity.

% ============================================================================
\begin{table*}[!h]
\centering
\small
\caption{Reward model designs for GRPO-aligned generative models. 
Methods range from lightweight task-specific reward signals to large-scale 
generative reward models with reasoning capability.}
\setlength{\neuripsOverhang}{\dimexpr(\paperwidth-\textwidth)/2 -6mm\relax}
\begin{adjustwidth*}{-\neuripsOverhang}{-\neuripsOverhang}
\setlength{\tabcolsep}{4pt}
% \begin{tabularx}{\dimexpr\paperwidth-12mm\relax}{
\begin{tabularx}{\dimexpr\paperwidth-12mm\relax}{
@{\hspace{2mm}}
>{\raggedright\arraybackslash}p{0.16\linewidth}
>{\raggedright\arraybackslash}p{0.11\linewidth}
>{\raggedright\arraybackslash}p{0.34\linewidth}
>{\raggedright\arraybackslash}p{0.34\linewidth}
@{\hspace{4mm}}
}
\toprule
\textbf{Category} & \textbf{Model} & \textbf{Core Design} & \textbf{Key Contribution} \\
\midrule

\multirow{2}{*}{\shortstack[l]{Lightweight / \\ Task-Specific}}
& HuDA~\cite{ashutosh2026human} 
& Human detector confidence + temporal prompt alignment 
& 73\% human win rate vs.\ Wan 2.1 baseline \\
& RealGen~\cite{ye2025realgen} 
& Detector-guided artifact reward 
(reverse-use of synthetic image detector) 
& Label-free photorealism alignment \\

\midrule

\multirow{2}{*}{\shortstack[l]{Generative \\Reward Paradigm}}
& RewardDance~\cite{wu2025rewarddance} 
& Large-scale generative reward (up to 26B); 
Natural language reasoning before scoring 
& Strong reward hacking resistance \\
& OneReward~\cite{gong2025onereward} 
& Unified VLM generative reward for generation/editing/SR 
& Multi-task winner–loser discrimination \\

\midrule

\multirow{3}{*}{\shortstack[l]{Structured / \\Granular Feedback}}
& VR-Thinker~\cite{wang2025vr} 
& CoT-style structured visual reasoning before scoring 
& 80.5\% accuracy on video preference benchmarks \\
& REACT~\cite{wang2026thinking} 
& Frame-level structural distortion evaluation (SFT+GRPO) 
& Fine-grained video feedback \\
& ImageDoctor~\cite{guo2025imagedoctor} 
& Pixel-level heatmap reward signals 
& Spatially dense defect localization \\

\midrule

\multirow{2}{*}{\shortstack[l]{Personalized / \\Domain-Specific}}
& UnifiedReward-Flex~\cite{wang2026unified} 
& Context-adaptive reasoning; hierarchical evaluation 
& Subjective preference modeling \\
& EditScore~\cite{luo2025editscore} 
& 7B–72B editing reward models 
& Matches/surpasses GPT-5-level editing evaluation \\

\bottomrule
\label{tab:reward_models}

\end{tabularx}
\end{adjustwidth*}
\end{table*}

\subsection{Reward Model Design and Evaluation Benchmarks}
\label{sec:reward_models}
The effectiveness of GRPO alignment critically depends on reward signal quality. An ideal reward model should accurately reflect human preferences, provide fine-grained feedback, and be robust to adversarial manipulation. Meanwhile, standardized evaluation benchmarks are essential for fair comparison of different methods. This section reviews important advances in both reward model design and evaluation benchmarks. The summary is shown in Table~\ref{tab:reward_models}, and the details are as follows.

\paragraph{Reward Models.}
Recent work on reward models for GRPO-aligned generation spans a wide spectrum from lightweight task-specific designs to large-scale generative paradigms. At one extreme, \textbf{HuDA}~\cite{ashutosh2026human} demonstrates that an extremely simple reward model---composed only of human detector confidence scores and temporal prompt alignment scores---can achieve a 73\% human evaluation win rate against the Wan~2.1 baseline in video generation GRPO post-training, suggesting that carefully designed lightweight signals may outperform complex general-purpose reward models for specific tasks. At the other extreme, \textbf{RewardDance}~\cite{wu2025rewarddance} proposes a scalable generative reward paradigm supporting simultaneous scaling along model size (up to 26B) and context (incorporating CoT reasoning chains), with the key finding that generative reward models---which output natural language evaluations before converting to scores---are naturally more resistant to reward hacking than discriminative counterparts. Between these extremes, several models target specific feedback granularities: \textbf{VR-Thinker}~\cite{wang2025vr}, a 7B-parameter ``visual thinking'' video reward model, first generates structured visual analysis reasoning (similar to Chain-of-Thought) before producing final scores, achieving open-source SOTA (80.5\% accuracy) on video preference benchmarks; \textbf{REACT}~\cite{wang2026thinking} specializes in frame-level structural distortion evaluation for video generation through SFT+GRPO two-stage training, providing finer-grained feedback than video-level rewards; and \textbf{ImageDoctor}~\cite{guo2025imagedoctor} provides pixel-level heatmaps as dense reward signals, enabling spatial quality/defect information to directly feed back into the denoising process. On the personalization front, \textbf{UnifiedReward-Flex}~\cite{wang2026unified} combines context-adaptive reasoning with multi-dimensional hierarchical evaluation to accommodate the inherently subjective nature of human preferences. For domain-specific applications, \textbf{EditScore}~\cite{luo2025editscore} releases a 7B--72B series of image editing reward models matching or surpassing GPT-5 level, \textbf{RealGen}~\cite{ye2025realgen} proposes detector-guided rewards that ``reverse-use'' pretrained synthetic image detectors to quantify AI artifacts without human labeling, and \textbf{OneReward}~\cite{gong2025onereward} provides a unified VLM generative reward model capable of accurately distinguishing winners from losers across multiple tasks including generation, editing, and super-resolution.

\paragraph{Evaluation Benchmarks.}
Several unified evaluation benchmarks have been proposed for generative models. SpatialGenEval~\cite{wang2026everything} is the first benchmark focused on spatial intelligence for text-to-image tasks, containing 1,230 spatial reasoning prompts covering relative positions, occlusion relations, perspective and other relations across 25 scenes. UniGenBench++~\cite{wang2025unigenbench++} provides a comprehensive suite of 600 prompts spanning 5 themes, 20 subthemes and 27 evaluation criteria, aiming to assess all key dimensions of T2I evaluation. T2AV‑Compass~\cite{cao2025t2av} is a unified framework for text-to-audio-video generation, containing 500 prompts and simultaneously evaluating both visual quality and audio consistency. MagicBench~\cite{wang2025magicmirror} offers a 340K-item human-annotated dataset of fine-grained artifacts, with detailed annotations for artifact type, location and severity. DivGenBench~\cite{chen2025taming} specifically measures preference mode collapse by evaluating semantic diversity across multiple generations for the same prompt, while RealBench~\cite{jin2025realbench} focuses on automated photorealistic quality assessment using synthetic image detectors as the main measurement tool.

%% file: section/2_application.tex
% ============================================================================
\section{Extensions to Generative Tasks}
\begin{table*}[!h]
\centering
\small
\caption{Representative extensions of Flow-GRPO in text-to-image generation. 
We categorize existing works into reasoning augmentation, multi-objective alignment, 
structural controllability, distribution preservation, adversarial rewards, 
test-time scaling, and advanced preference modeling.}

\setlength{\neuripsOverhang}{\dimexpr(\paperwidth-\textwidth)/2 -6mm\relax}
\begin{adjustwidth*}{-\neuripsOverhang}{-\neuripsOverhang}
\setlength{\tabcolsep}{4pt}
% \begin{tabularx}{\dimexpr\paperwidth-12mm\relax}{
\begin{tabularx}{\dimexpr\paperwidth-12mm\relax}{
@{\hspace{2mm}}
>{\raggedright\arraybackslash}p{0.16\linewidth}
>{\raggedright\arraybackslash}p{0.15\linewidth}
>{\raggedright\arraybackslash}p{0.32\linewidth}
>{\raggedright\arraybackslash}p{0.32\linewidth}
@{\hspace{4mm}}
}

\toprule
\textbf{Category} & \textbf{Method} & \textbf{Core Innovation} & \textbf{Key Contribution / Result} \\
\midrule

\multirow{7}{*}{\shortstack[l]{Reasoning-Augmented \\Alignment}}
& PromptRL~\cite{wang2026promptrl} & Trainable LM as prompt optimizer within RL loop & End-to-end co-optimization; 0.97 GenEval; 50\% fewer rollouts \\
& Think-Then-Generate~\cite{kou2026think} & Two-stage reasoning + generation & Dual-GRPO optimization of reasoner and generator \\
& ThinkGen~\cite{jiao2025thinkgen} & Think-driven structured visual planning & SepGRPO alternating optimization for stability \\
& Unified Thinker~\cite{zhou2026unified} & Fully decoupled reasoning engine & Task-agnostic reasoning across T2I/editing/SR \\
& CoF-T2I~\cite{tong2026cof} & Chain-of-Frame visual reasoning & Progressive scene construction via intermediate frames \\
& AgentComp~\cite{zarei2025agentcomp} & Agent-based compositional data construction & Tool-use-based preference supervision \\
& Alignment of DM \& FM~\cite{ouyang2026alignment} & Lightweight alignment schemes & 60\% compute reduction via guidance/schedule tuning \\

\midrule

\multirow{2}{*}{\shortstack[l]{Multi-Objective \\Alignment}}
& APEX~\cite{chen2026apex} & Dual-stage reward normalization & Mitigates variance hijacking and gradient conflict \\
& MapReduce LoRA~\cite{chen2025mapreduce} & Modular preference factorization & Map--Reduce LoRA experts + controllable test-time merging \\

\midrule

\multirow{3}{*}{\shortstack[l]{Text Rendering \& \\Long Text}}
& DCText~\cite{song2025dctext} & Training-free region disentanglement & Divide-and-conquer attention masking \\
& Ovis-Image~\cite{wang2025ovis} & Text-centric pretraining pipeline & Strong bilingual OCR at moderate scale \\
& Qwen-Image~\cite{wu2025qwen} & Paragraph-level curriculum learning & Improved long-text semantic control \\

\midrule

\multirow{3}{*}{Identity Preservation}
& PSR~\cite{wang2025psr} & Pairwise subject-consistency reward & Relational identity optimization \\
& MultiCrafter~\cite{wu2025multicrafter} & Disentangled identity attention & Reduced multi-subject interference \\
& Customized-GRPO~\cite{huang2025competition} & Synergy-aware reward shaping & Time-aware dynamic weighting strategy \\

\midrule

\multirow{2}{*}{Photorealism}
& RealGen~\cite{ye2025realgen} & Artifact detector as reward & $r = 1 - D(x)$; full pipeline alignment \\
& ProxT2I~\cite{fang2025proxt2i} & Conditional proximal modeling & Structural face-quality constraints \\

\midrule

\multirow{2}{*}{\shortstack[l]{Compositionality \& \\Numerosity}}
& HiCoGen~\cite{yang2025hicogen} & Chain of Synthesis decomposition & Hierarchical object + relation rewards \\
& Demystifying Numerosity~\cite{zhao2025demystifying} & Count-aware noise injection & Counting accuracy improved from 20\% to 85.3\% \\

\midrule

\multirow{2}{*}{Few-Step Generation}
& Flash-DMD~\cite{chen2025flash} & Joint distillation + RL & Stable few-step generation at low training cost \\
& DMDR~\cite{jiang2025distribution} & Distribution matching regularization & $\mathcal{L} = \mathcal{L}_{RL} + \beta \mathcal{L}_{DMD}$ \\

\midrule

\multirow{2}{*}{\shortstack[l]{Adversarial / \\Reward-Free}}
& Adv-GRPO~\cite{mao2025image} & Discriminator as reward signal & 70\%+ human win rate \\
& MoGAN~\cite{xue2025mogan} & Optical-flow discriminator & Motion realism reward + distribution regularization \\

\midrule

\multirow{6}{*}{Test-Time Scaling}
& TTSnap~\cite{yu2025ttsnap} & Noise-aware early pruning & 16\%+ performance improvement \\
& Text Embedding Perturbation~\cite{xu2025highly} & Frequency-domain embedding modification & Sharper high-frequency visual details \\
& HyperAlign~\cite{xie2026hyperalign} & Hypernetwork-generated LoRA & Dynamic reward-conditioned adaptation \\
& RLG~\cite{guidanceinference} & Geometric distribution fusion & Adjustable alignment trade-off \\
& Exploration-Exploitation~\cite{su2025navigating} & Phase-aware adaptive temperature & Funnel schedule for semantic transitions \\
& MILR~\cite{mi2025milr} & Multi-modal latent reasoning & 80\% improvement on WISE benchmark \\

\midrule

\multirow{8}{*}{\shortstack[l]{Preference \& \\Reward Modeling}}
& Pref-GRPO~\cite{wang2025pref} & Pairwise preference fitting & Avoids pointwise reward amplification \\
& RubricRL~\cite{feng2025rubricrl} & Interpretable checklist reward & Transparent modular scoring \\
& EmoFeedback$^2$~\cite{jia2025emofeedback} & Emotion-aware reward feedback & MLLM-based affective supervision \\
& PaCo-RL~\cite{ping2025paco} & Pairwise consistency reward & Resolution-decoupled optimization \\
& BideDPO~\cite{zhou2025bidedpo} & Bidirectionally decoupled DPO & Text/condition conflict mitigation \\
& Emotion-Director~\cite{jia2025emotion} & Affective shortcut mitigation & Cross-modal diffusion alignment \\
& Data-Driven Loss~\cite{yiflach2025data} & Learned spatial reasoning loss & Spatial accuracy improved (0.20 $\rightarrow$ 0.61) \\
& James-Stein Shrinkage~\cite{yu2025designing} & Instance-level schedule learning & 5-step inference matches full-quality generation \\

\bottomrule
\label{tab:t2i}
\end{tabularx}
\end{adjustwidth*}
\end{table*}

\subsection{Text-to-Image Generation}
\label{sec:t2i}

Text-to-image (T2I) generation is the broadest and most method-rich application domain of Flow-GRPO. This section systematically reviews progress across multiple sub-directions including reasoning-enhanced generation, multi-objective alignment, text rendering, identity preservation, photorealism, compositional reasoning, few-step generation, adversarial rewards, and test-time scaling. 
The summary is shown in Table~\ref{tab:t2i}, and the details are as follows.

\paragraph{Core Alignment Methods.}
Recent extensions of Flow-GRPO progressively move beyond pure rollout-based reward optimization and incorporate explicit reasoning modules, prompt refinement mechanisms, agent-based coordination, and lightweight alignment strategies into the flow reinforcement learning loop, reflecting a shift from trajectory-level optimization toward structured multi-module co-optimization. At the input level, \textbf{PromptRL}~\cite{wang2026promptrl} introduces a new paradigm that integrates a \textbf{trainable language model as a prompt optimization agent} into the flow RL loop, based on the insight that generation quality depends not only on the diffusion model but also on the quality of the input prompt; it jointly optimizes a lightweight LM that rewrites user prompts together with the diffusion model, with reward signals backpropagated to both modules simultaneously, forming an end-to-end co-optimization strategy that achieves \textbf{0.97 GenEval, 0.98 OCR accuracy, and 24.05 PickScore}, while halving rollout requirements because improved prompts make each rollout more informative. Moving beyond prompt rewriting, \textbf{Think-Then-Generate (T2G)}~\cite{kou2026think} decouples reasoning and generation into two stages, where an LLM encoder first performs \textbf{reasoning rewriting}—analyzing spatial relationships, quantity requirements, and attribute constraints to produce a structured enhanced prompt—followed by image synthesis via diffusion; both stages are jointly optimized through \textbf{Dual-GRPO}, rewarding the reasoner for producing prompts more helpful to the generator and the generator for better executing the reasoner’s instructions, achieving 0.79 on the WISE benchmark and approaching GPT-4-level reasoning performance. Similarly, \textbf{ThinkGen}~\cite{jiao2025thinkgen} proposes the first \textbf{think-driven visual generation} framework composed of an MLLM and a DiT, introducing \textbf{SepGRPO (Separated GRPO)} to alternately freeze one module while optimizing the other to avoid instability from simultaneously updating two large models; the MLLM generates structured visual plans (e.g., layouts and key object descriptions), which the DiT translates into final images. Extending this modular separation, \textbf{Unified Thinker}~\cite{zhou2026unified} proposes a \textbf{task-agnostic reasoning architecture} that fully decouples the Thinker'' and Generator,'' enabling a general-purpose reasoning engine to provide structured guidance across tasks such as T2I, editing, and super-resolution. Beyond textual reasoning, \textbf{CoF-T2I}~\cite{tong2026cof} introduces the novel paradigm of \textbf{Chain-of-Frame (CoF) reasoning}, leveraging intermediate frames of a video model as explicit visual reasoning steps that progressively construct a scene—first placing the background, then main objects, then refining attribute details—with the final frame corresponding to the complete generation result, thereby translating chain-of-thought into chain-of-visualization for compositional prompts. At the data and preference level, \textbf{AgentComp}~\cite{zarei2025agentcomp} integrates LLM \textbf{reasoning capabilities and tool-use abilities} into compositional dataset construction, where an LLM agent analyzes prompt structure, calls detectors and segmenters to verify outputs, and constructs precise preference pairs, leading to significant improvements in compositional generation quality. Finally, \textbf{Alignment of DM and FM}~\cite{ouyang2026alignment} proposes two alignment schemes: a \textbf{guidance network framework} for diffusion models that learns a lightweight network to modify denoising directions, and a \textbf{fully training-free alignment framework} for flow matching models that modifies the inference-time noise schedule, both reducing computational cost by over 60\% while maintaining alignment effectiveness. Collectively, these works demonstrate that Flow-GRPO is evolving from a pure reinforcement-based optimization framework into a \textbf{reasoning-augmented, agent-driven, and modular optimization paradigm}, where prompts, structured plans, visual reasoning trajectories, preference construction, and inference-time alignment are jointly integrated to improve both performance and efficiency.

\paragraph{Multi-Objective Alignment.}
In parallel with reasoning-augmented extensions, recent works further investigate how to stabilize and modularize multi-objective optimization within the GRPO framework. \textbf{APEX}~\cite{chen2026apex} addresses two fundamental challenges in multi-objective GRPO: \textbf{variance hijacking}, where high-variance rewards (e.g., aesthetics scores) generate disproportionately large advantage signals after group normalization and thus dominate low-variance but equally important objectives such as safety, and \textbf{gradient conflict}, where gradients from different objectives point in contradictory directions and hinder coordinated optimization. To mitigate these issues, APEX proposes \textbf{dual-stage adaptive normalization}, first normalizing rewards within each objective and then across objectives to rebalance advantage magnitudes, together with \textbf{P$^3$ Adaptive Priorities}, which dynamically adjusts objective weights based on each objective’s current degree of under-optimization, thereby preventing premature saturation of easier objectives while encouraging lagging ones; on SD 3.5, APEX achieves balanced four-objective improvements (+1.31 PickScore, +0.35 DeQA, +0.53 Aesthetics), demonstrating stable multi-objective alignment. Complementarily, \textbf{MapReduce LoRA}~\cite{chen2025mapreduce} introduces a modular preference factorization strategy inspired by distributed computing: during the \textbf{Map phase}, a dedicated LoRA expert is independently trained for each preference dimension (e.g., aesthetics LoRA, consistency LoRA, safety LoRA), and during the \textbf{Reduce phase}, these LoRA experts are iteratively merged into a unified backbone model; at inference time, \textbf{RaTE (Rank-Adaptive Test-time Editing)} adaptively controls the contribution weight of each preference dimension, enabling user-controllable personalization without retraining. This Map–Reduce-style decomposition decouples preference optimization across dimensions while preserving unified deployment, achieving SOTA performance across T2I, T2V, and cross-modal language tasks.

\paragraph{Text Rendering and Visual Text.}
Beyond optimization-level and modular alignment improvements, recent works further enhance text rendering capability and long-text conditioning within generative frameworks. \textbf{DCText}~\cite{song2025dctext} proposes a \textbf{training-free divide-and-conquer} strategy for long-text and multi-text rendering, where complex text prompts are segmented into multiple sub-regions and each region is controlled by independent attention masks, explicitly preventing cross-talk between textual areas and enabling precise spatial disentanglement without additional training. Complementarily, \textbf{Ovis-Image}~\cite{wang2025ovis} develops a compact 7B-parameter T2I model with a text-centric training pipeline that extensively incorporates images containing text during pretraining, thereby strengthening text recognition and rendering alignment; its bilingual (Chinese–English) text rendering performance is reported to be comparable to GPT-4o, demonstrating that targeted data-centric design can significantly improve textual fidelity even at moderate model scale. Extending toward long-form conditioning, \textbf{Qwen-Image}~\cite{wu2025qwen} achieves high-quality long-text-conditioned image generation through \textbf{paragraph-level description curriculum learning}, gradually transitioning from short captions to full paragraph descriptions during training, and integrates this curriculum with the Qwen2.5-VL dual-encoder architecture to better capture semantic structure in extended textual contexts. Collectively, these approaches illustrate a complementary direction to reward-based optimization: strengthening structural controllability and textual grounding through inference-time disentanglement, data-centric scaling, and curriculum-based long-text conditioning.

\paragraph{Identity/Subject Preservation.}
Extending reinforcement-based alignment to multi-subject personalization, recent works focus on mitigating identity interference and competition during joint optimization. \textbf{PSR}~\cite{wang2025psr} introduces \textbf{pairwise subject-consistency rewards} for multi-subject personalized generation, where the reward function explicitly measures identity consistency between each pair of subjects in the generated image and their corresponding reference images, thereby transforming identity preservation into a structured relational objective rather than independent per-subject scoring. Building on this direction, \textbf{MultiCrafter}~\cite{wu2025multicrafter} addresses the identity confusion problem through a \textbf{disentangled attention} mechanism that assigns independent attention channels to different identity subjects, reducing feature entanglement during generation, and combines this architectural separation with identity-preserving preference optimization to further stabilize subject fidelity. Moreover, \textbf{Customized-GRPO}~\cite{huang2025competition} identifies the \textbf{competitive degradation} phenomenon in multi-subject personalization—where improving one subject’s identity consistency inadvertently degrades another’s—and proposes two complementary solutions: \textbf{Synergy-Aware Reward Shaping (SARS)}, which formulates the reward as a joint function over all subjects’ consistency to encourage cooperative optimization, and \textbf{Time-Aware Dynamic Weighting (TDW)}, which dynamically adjusts attention weights assigned to different subjects across different phases of the denoising process to prevent early-stage dominance or late-stage suppression. Collectively, these approaches reveal that multi-subject personalization introduces higher-order coupling effects in GRPO optimization, requiring relational reward design, architectural disentanglement, and time-dependent weighting strategies to maintain balanced identity preservation.

\paragraph{Photorealism.}
In pursuit of improved photorealism and domain-specific quality alignment, recent works explore alternative reward formulations and constraint-based modeling. \textbf{RealGen}~\cite{ye2025realgen} ``reverses'' the conventional use of AI artifact detectors by employing a pretrained synthetic image detector $D(x)$ as a reward signal, defining $r = 1 - D(x)$ to explicitly penalize detectable artifacts; notably, it optimizes the \textbf{full pipeline}, jointly fine-tuning both the diffusion model and the LLM prompt optimizer so that the entire generation system is end-to-end aligned toward photorealistic outputs rather than only adjusting the backbone model. Complementarily, \textbf{ProxT2I}~\cite{fang2025proxt2i} formulates face generation from the perspective of \textbf{conditional proximal operators}, learning a mapping that projects noisy latent variables onto a subspace satisfying facial quality constraints, thereby embedding structural priors directly into the generation dynamics; it further releases the \textbf{LAION-Face-T2I-15M} dataset, comprising 15 million high-quality face image–text pairs, to support large-scale, face-specific text-to-image alignment.

\paragraph{Compositionality and Spatial Reasoning.}
Addressing compositional reasoning and numerosity limitations in generative models, recent works investigate structured decomposition and data-bias-aware interventions. \textbf{HiCoGen}~\cite{yang2025hicogen} introduces the \textbf{Chain of Synthesis (CoS)} paradigm for complex compositional prompts (e.g., ``red cup to the left of a blue plate, green ball above the cup''), where an LLM first decomposes the prompt into multiple sub-generation tasks (e.g., separately generating the cup, plate, and ball), followed by iterative object-wise synthesis via the diffusion model; during optimization, \textbf{hierarchical rewards} are applied to simultaneously evaluate individual sub-object quality and the correctness of global spatial relationships, enabling reinforcement learning to supervise both local fidelity and relational consistency. Complementarily, \textbf{Demystifying Numerosity}~\cite{zhao2025demystifying} provides a thought-provoking empirical analysis demonstrating that diffusion models’ counting ability largely \textbf{depends on noise biases in training data}—such as the frequency of images containing specific object counts—rather than genuine semantic grounding of textual numerosity cues; based on this insight, the authors inject \textbf{count-aware layout information} by encoding spatial layout priors for the target count directly into the initial noise, leading to a dramatic improvement in counting accuracy from 20\% to \textbf{85.3\%}. Together, these works suggest that compositional and counting failures stem from structural decomposition gaps and data-induced biases, and can be mitigated through hierarchical synthesis strategies and noise-level layout conditioning.

\paragraph{Few-Step Generation.}
To reconcile efficiency-oriented distillation with reward-driven optimization, recent works explore tighter integration between distribution matching and RL objectives. \textbf{Flash-DMD}~\cite{chen2025flash} proposes \textbf{joint training} of distillation and RL, where the distillation loss constrains the few-step student model to remain close to the teacher model’s distribution—thereby preventing RL-induced policy collapse or mode drift—while the RL loss simultaneously drives the student toward higher-reward regions; this dual-loss design enables stable reward improvement under aggressive step reduction, surpassing DMD2 at only \textbf{2.1\% of the training cost}. In a related but more explicit regularization formulation, \textbf{DMDR}~\cite{jiang2025distribution} incorporates \textbf{DMD (Distribution Matching Distillation) loss as an RL regularization term}, defining the overall objective as $\mathcal{L} = \mathcal{L}{\text{RL}} + \beta \mathcal{L}{\text{DMD}}$, where the DMD component acts as a distributional anchor that prevents the model from drifting away from the original data manifold during RL optimization, while the RL term encourages movement toward high-reward directions. Together, these approaches illustrate a principled strategy for balancing reward maximization and distribution preservation, enabling efficient few-step generation without sacrificing stability.

\paragraph{Adversarial Rewards and Reward-Model-Free.}
Beyond reliance on pretrained reward models, recent works explore reward-model-free adversarial optimization within the GRPO framework. \textbf{Adv-GRPO}~\cite{mao2025image} introduces an \textbf{adversarial training paradigm} in which the generated image itself becomes the source of reward signals: a discriminator is trained to distinguish real images from generated ones, and its discrimination score is directly used as the GRPO reward, while the generator is simultaneously optimized via GRPO to fool the discriminator, forming a co-evolving adversarial game without requiring an external pretrained reward model. This fully coupled adversarial–reinforcement loop enables automatic reward shaping grounded in data realism, and in human evaluations, Adv-GRPO achieves a \textbf{70\%+ win rate} against Flow-GRPO and SD3 baselines. In a related direction for video generation, \textbf{MoGAN}~\cite{xue2025mogan} proposes another reward-model-free scheme by introducing an \textbf{optical flow discriminator} that explicitly evaluates motion naturalness in generated videos, combined with \textbf{distribution matching regularization} to prevent the generated distribution from drifting away from real data statistics. Together, these approaches demonstrate that adversarial discriminators can serve as adaptive, data-driven reward sources, enabling GRPO-style optimization without dependence on fixed pretrained reward models while maintaining distributional stability.

\paragraph{Test-Time Scaling.}
Complementing training-time optimization, a growing line of work explores inference-time alignment and control strategies that improve reward performance with minimal additional cost. \textbf{TTSnap}~\cite{yu2025ttsnap} introduces a \textbf{noise-aware pruning} strategy in which multiple candidate samples are generated at inference time, but low-quality candidates are eliminated during early denoising steps based on intermediate state quality estimates; since early denoising steps incur far lower computational cost than full generation, this approach achieves a \textbf{16\%+ performance improvement} with negligible overhead. \textbf{Text Embedding Perturbation}~\cite{xu2025highly} uncovers, through \textbf{frequency-domain analysis}, that small high-frequency perturbations applied to text embeddings can enhance high-frequency visual details such as textures and edges in generated images, yielding improved fidelity with nearly zero additional computation. Moving toward adaptive parameterization, \textbf{HyperAlign}~\cite{xie2026hyperalign} trains a \textbf{hypernetwork} that dynamically generates low-rank adaptation (LoRA) weights at test time conditioned on reward signals, enabling alignment adjustments without modifying latent variables or requiring multi-step optimization. From a probabilistic perspective, \textbf{RLG}~\cite{guidanceinference} combines baseline and RL-optimized output distributions via \textbf{geometric averaging}, $p_{\text{RLG}} = p_{\text{base}}^{1-\alpha} \cdot p_{\text{RL}}^{\alpha}$, allowing dynamic trade-offs between alignment strength and base model quality through $\alpha \in [0, 1]$. \textbf{Exploration-Exploitation}~\cite{su2025navigating} further observes that diffusion denoising exhibits \textbf{phase transition behavior}, characterized by abrupt semantic shifts in latent variables near specific timesteps, and accordingly designs a \textbf{funnel schedule} with \textbf{adaptive temperature}—increasing exploration (high temperature) near transition points and strengthening exploitation (low temperature) elsewhere. Finally, \textbf{MILR} (Multi-modal Inference-time Latent Reasoning)~\cite{mi2025milr} proposes \textbf{joint reasoning over image and text latent variables} at test time, simultaneously optimizing image latents and text embeddings so that both cooperatively converge toward high-reward regions, achieving an \textbf{80\% improvement} on the WISE benchmark. Collectively, these methods demonstrate that substantial alignment gains can be realized purely at inference time through pruning, perturbation, adaptive parameter generation, probabilistic fusion, temperature scheduling, and multi-modal latent co-optimization.

\paragraph{Other T2I Methods.}
Beyond architectural and inference-time strategies, a significant line of work revisits the design of reward functions and preference modeling within GRPO and related alignment frameworks. \textbf{Pref-GRPO}~\cite{wang2025pref} exposes the \textbf{hidden danger of pointwise rewards} in standard GRPO: when reward scores differ only marginally within a group, group normalization can amplify these small differences into extreme advantage signals, leading to unstable gradient updates; it instead adopts \textbf{pairwise preference fitting}, directly modeling relative preference relations between samples to avoid pointwise score amplification. Complementarily, \textbf{RubricRL}~\cite{feng2025rubricrl} proposes \textbf{rubric-based interpretable reward design}, dynamically constructing a \textbf{visual checklist} for each prompt (e.g., Does it contain a red car?'' Is the background a city?'') where each item is independently scored and then aggregated, making reward signals transparent and debuggable. Targeting affective alignment, \textbf{EmoFeedback$^2$}~\cite{jia2025emofeedback} introduces \textbf{emotion-aware reward feedback}, in which an MLLM evaluates whether generated images convey the intended emotion, together with \textbf{self-promotion textual feedback} that generates improvement suggestions to iteratively refine outputs. For consistency-focused tasks, \textbf{PaCo-RL}~\cite{ping2025paco} designs a \textbf{pairwise consistency reward model} for consistent character generation across scenes and proposes \textbf{resolution-decoupled optimization}, separately optimizing low-resolution global consistency and high-resolution detail fidelity. Addressing text–condition conflicts, \textbf{BideDPO}~\cite{zhou2025bidedpo} proposes \textbf{Bidirectionally Decoupled DPO}, separately modeling text-dominant and condition-dominant preference signals to mitigate failures when textual descriptions contradict reference conditions. \textbf{Emotion-Director}~\cite{jia2025emotion} tackles the \textbf{affective shortcut} problem—models substituting complex emotional expression with simple visual proxies—through cross-modal collaborative diffusion combined with DPO to encourage richer affective representations. Moving beyond handcrafted rewards, \textbf{Data-Driven Loss}~\cite{yiflach2025data} learns a \textbf{spatial reasoning loss function} via contrastive learning on images with correct and incorrect spatial relations, enabling end-to-end neural evaluation of spatial correctness and improving spatial accuracy on FLUX.1-dev from 0.20 to \textbf{0.61}. Finally, \textbf{James-Stein Shrinkage}~\cite{yu2025designing} explores an alternative to weight fine-tuning by \textbf{learning instance-level sampling schedules}: for each prompt, an optimal denoising schedule (timestep sizes and noise strengths) is learned and then shrunk toward a global average schedule inspired by the James–Stein estimator, balancing adaptivity and generalization such that 5-step Flux-Dev matches the quality of Flux-Schnell. Collectively, these approaches highlight a broad shift from naive scalar reward optimization toward structured preference modeling, interpretable and domain-aware reward construction, adaptive scheduling, and learned loss formulations.

% ============================================================================

\begin{table*}[!h]
\centering
\small
\caption{Representative extensions of Flow-GRPO in video generation. 
We categorize existing works into core T2V/I2V alignment, identity preservation, 
motion control, editing, video reasoning, frontier large-scale systems, and reward modeling.}
\setlength{\neuripsOverhang}{\dimexpr(\paperwidth-\textwidth)/2 -6mm\relax}
\begin{adjustwidth*}{-\neuripsOverhang}{-\neuripsOverhang}
\setlength{\tabcolsep}{4pt}
% \begin{tabularx}{\dimexpr\paperwidth-12mm\relax}{
\begin{tabularx}{\dimexpr\paperwidth-12mm\relax}{
@{\hspace{2mm}}
>{\raggedright\arraybackslash}p{0.16\linewidth}
>{\raggedright\arraybackslash}p{0.15\linewidth}
>{\raggedright\arraybackslash}p{0.32\linewidth}
>{\raggedright\arraybackslash}p{0.32\linewidth}
@{\hspace{4mm}}
}
\toprule
\textbf{Category} & \textbf{Method} & \textbf{Core Innovation} & \textbf{Key Contribution / Result} \\
\midrule

\multirow{7}{*}{Core T2V / I2V Methods}
& TAGRPO~\cite{wang2026tagrpo} & Memory-bank contrastive I2V alignment & Progressive lower-bound quality improvement \\
& PhysRVG~\cite{zhang2026physrvg} & Physics-engine-based reward & Verifiable rigid-body trajectory consistency \\
& Diffusion-DRF~\cite{wang2026diffusion} & Frozen VLM critic (e.g., GPT-4V) & Training-free natural-language reward backprop \\
& Self-Paced GRPO~\cite{li2025growing} & Competence-aware curriculum reward & Stable progressive temporal alignment \\
& PG-DPO~\cite{xu2025beyond} & Adaptive rejection scaling & Stabilizes high-dimensional DPO optimization \\
& DenseDPO~\cite{deng2026densegrpo} & Segment-level preference labeling & Motion improvement with 1/3 annotation data \\
& McSc~\cite{yang2025mcsc} & Multi-dimensional self-critic reasoning & Motion-corrective DPO while preserving strengths \\

\midrule

\multirow{4}{*}{Identity \& Consistency}
& Identity-GRPO~\cite{meng2025identity} & Temporally-aware identity reward model & +18.9\% identity consistency \\
& IPRO~\cite{shen2025identity} & Multi-angle identity feature pooling & Pose-robust identity scoring \\
& ID-Crafter~\cite{pan2025id} & Hierarchical identity attention + online RL & Joint identity and semantic alignment \\
& DreamID-V~\cite{guo2026dreamid} & Identity-coherence RL for face swapping & Unified realism and temporal consistency \\

\midrule

\multirow{4}{*}{Motion Control}
& AR-Drag~\cite{zhao2025real} & RL-enhanced trajectory-conditioned diffusion & Real-time few-step motion control \\
& Camera-Controlled Video~\cite{wang2025taming} & Geometry-verifiable 3D alignment reward & Segment-level camera trajectory consistency \\
& MoGAN~\cite{xue2025mogan} & Optical-flow discriminator & +7.3\% motion realism improvement \\
& HY-Motion 1.0~\cite{wen2025hy} & Billion-scale flow matching for 3D motion & 200+ motion categories; improved text-motion alignment \\

\midrule

\multirow{2}{*}{Video Editing}
& VIVA~\cite{cong2025viva} & Edit-GRPO for instruction fidelity + temporal coherence & Scalable synthetic editing dataset \\
& ReViSE~\cite{liu2025revise} & Reason-informed semantic decomposition & +32\% editing score improvement \\

\midrule

\multirow{3}{*}{\shortstack[l]{Video Understanding \& \\Prediction}}
& Video-as-Answer~\cite{cheng2025video} & Joint-GRPO for video QA generation & Video responses as causal answers \\
& What Happens Next~\cite{li2025happens} & Causal consistency reward & Plausible next-scene prediction \\
& RLIR~\cite{ye2025reinforcement} & Inverse-dynamics-based verifiable reward & +5--10\% action-following performance \\

\midrule

\multirow{6}{*}{Frontier Video Synthesis}
& FSVideo~\cite{team2026fsvideo} & High compression-ratio latent space & 10$\times$ faster inference \\
& LongCat-Video~\cite{team2025longcat} & Block sparse attention & Minute-scale 720p generation \\
& Seedance 1.0~\cite{gao2025seedance} & Multi-dimensional RLHF + engineering acceleration & 10$\times$ inference acceleration \\
& Seedance 1.5 pro~\cite{seedance2025seedance} & Native audio-visual joint generation & Multilingual lip-sync capability \\
& Self-Forcing++~\cite{cui2025self} & Autoregressive ultra-long generation & 4m15s video; 50$\times$ longer than teacher \\
& Video Text Preservation~\cite{liu2025video} & Text-rich synthetic fine-tuning & Improved video text rendering fidelity \\

\midrule

\multirow{6}{*}{Video Reward Modeling}
& VR-Thinker~\cite{wang2025vr} & Reasoning-enhanced video reward model & Structured multi-dimensional scoring \\
& SoliReward~\cite{lian2025solireward} & Multi-aspect video preference modeling & Improved evaluation reliability \\
& HuDA~\cite{ashutosh2026human} & Human-aware dense alignment reward & Fine-grained human feedback modeling \\
& REACT~\cite{wang2026thinking} & Thinking-based video critic & Reasoning-augmented reward generation \\
& Reward Forcing~\cite{lu2025reward} & Reward-biased DMD (Re-DMD) & Distribution matching toward high-reward regions \\
& VQ-Insight~\cite{zhang2025vq} & Progressive visual RL critic training & From binary to multi-dimensional evaluation \\

\bottomrule
\label{tab:video}

\end{tabularx}
\end{adjustwidth*}
\end{table*}

\subsection{Video Generation}
\label{sec:video}

Video generation extends Flow-GRPO's alignment challenges from 2D static images to the 3D spatiotemporal domain, introducing new dimensions of problems including temporal consistency, motion naturalness, and identity preservation. This section systematically reviews progress in T2V/I2V core methods, identity consistency, motion control, video editing, video understanding and prediction.
The summary is shown in Table~\ref{tab:video}, and the details are as follows.

\paragraph{Core T2V/I2V Methods.}
Extending GRPO-style alignment to video generation introduces additional challenges such as temporal coherence, reference consistency, and high-dimensional preference instability, prompting a range of specialized solutions. \textbf{TAGRPO}~\cite{wang2026tagrpo} focuses on \textbf{image-to-video (I2V)} alignment, where generated videos must remain visually consistent with the input reference image; it proposes a contrastive post-training framework equipped with a \textbf{memory bank} that stores successfully generated high-quality samples, and trains new samples via contrastive learning against this bank, progressively elevating the lower bound of generation quality. From a physics-grounded perspective, \textbf{PhysRVG}~\cite{zhang2026physrvg} employs \textbf{physical rules} as verifiable reward signals: for interaction-heavy videos (e.g., bouncing balls, object collisions), a physics engine simulates rigid-body mechanics and compares simulated trajectories with those observed in generated videos, using discrepancies as penalty terms, thereby providing rewards that are inherently resistant to reward hacking since physical laws cannot be ``fooled.'' In a training-free alignment direction, \textbf{Diffusion-DRF}~\cite{wang2026diffusion} utilizes a \textbf{frozen VLM (e.g., GPT-4V) as a critic}, which evaluates each video frame in natural language, parses the evaluation into structured scores, and \textbf{directly backpropagates} them to the diffusion model without requiring dedicated reward model training. Addressing curriculum and stability, \textbf{Self-Paced GRPO}~\cite{li2025growing} introduces a \textbf{competence-aware progressive reward} mechanism that begins with simple criteria (e.g., coarse visual fidelity) and gradually transitions to more demanding standards (e.g., temporal coherence, motion naturalness) as model capability improves, mitigating instability from prematurely applying complex rewards; it achieves consistent gains on VBench and generalizes across multiple video backbones. From a preference-learning perspective, \textbf{PG-DPO}~\cite{xu2025beyond} identifies \textbf{likelihood displacement} in high-dimensional video latent spaces, where standard DPO's implicit reward becomes unstable, and proposes \textbf{adaptive rejection scaling} together with \textbf{implicit preference regularization} to stabilize optimization. \textbf{DenseDPO}~\cite{deng2026densegrpo} further refines video-level supervision into \textbf{segment-level preference labeling}, enabling annotators to independently evaluate temporal segments rather than entire videos, dramatically improving motion quality with only \textbf{1/3 of the labeling data}. Finally, \textbf{McSc}~\cite{yang2025mcsc} introduces a \textbf{self-critic dimensional reasoning} mechanism in which the model first performs multi-dimensional self-evaluation (e.g., visual fidelity, motion naturalness, text consistency) and then applies \textbf{motion-corrective DPO} to selectively improve deficient motion aspects while preserving strengths in other dimensions. Collectively, these methods illustrate that video alignment requires memory-augmented contrastive learning, physics-grounded verification, training-free VLM critics, curriculum-style reward scheduling, high-dimensional DPO stabilization, fine-grained preference annotation, and self-corrective reasoning to jointly address temporal, physical, and semantic consistency.

\paragraph{Identity and Consistency.}
Focusing specifically on identity preservation in video generation, recent works tackle the severe challenge of maintaining consistent character appearance across long temporal horizons under variations in facial angle, lighting, and occlusion. \textbf{Identity-GRPO}~\cite{meng2025identity} is the first method explicitly targeting \textbf{multi-person identity preservation} in video generation, training a \textbf{human-feedback-driven identity reward model} that scores identity consistency of the same character across different frames; after GRPO optimization, identity consistency improves by \textbf{18.9\%}, demonstrating the effectiveness of temporally-aware identity rewards. Complementarily, \textbf{IPRO}~\cite{shen2025identity} introduces a dedicated \textbf{face identity scorer} together with a \textbf{multi-angle feature pool}, which averages features extracted from multiple facial orientations to produce orientation-robust identity scores, mitigating identity drift caused by pose variation. \textbf{ID-Crafter}~\cite{pan2025id} integrates three complementary mechanisms: \textbf{hierarchical identity-preserving attention} to maintain identity information across spatial scales, \textbf{VLM semantic understanding} to ensure textual alignment, and \textbf{online RL} to jointly optimize identity preservation and semantic consistency in an end-to-end manner. Finally, \textbf{DreamID-V}~\cite{guo2026dreamid} presents the first \textbf{DiT-based video face-swapping framework} enhanced with \textbf{identity-coherence RL}, where the reward simultaneously evaluates post-swap facial realism and cross-frame identity consistency, unifying realism and temporal coherence within a reinforcement learning objective. Collectively, these methods highlight that multi-person video identity alignment requires temporally consistent reward modeling, pose-robust feature aggregation, hierarchical attention design, and joint optimization of identity and semantic objectives.

\paragraph{Motion Control.}
Targeting controllable and physically grounded motion generation, recent works integrate reinforcement learning with trajectory-level and geometry-level supervision across both video and 3D motion domains. \textbf{AR-Drag}~\cite{zhao2025real} presents the first \textbf{RL-enhanced few-step autoregressive video diffusion model} capable of real-time motion control, where users specify motion trajectories through drag operations and the model generates videos that conform to these motion instructions within only a few inference steps; the RL reward explicitly measures the correspondence between generated motion and user-specified trajectories, enabling precise trajectory adherence under extremely low step budgets. Complementarily, \textbf{Camera-Controlled Video}~\cite{wang2025taming} leverages \textbf{verifiable geometry rewards} by recovering 3D camera trajectories from generated videos using monocular depth estimation and visual odometry, and performing dense \textbf{segment-level 3D alignment} against target trajectories, thereby grounding motion control in explicit geometric consistency. From a motion realism perspective, \textbf{MoGAN}~\cite{xue2025mogan} incorporates an \textbf{optical flow discriminator} into the DiT architecture to evaluate whether extracted optical flow fields conform to natural motion statistics, improving motion quality scores by \textbf{7.3\%}. Reinforcement alignment has also been extended to large-scale 3D human motion synthesis. \textbf{HY-Motion 1.0}~\cite{wen2025hy} is the first \textbf{billion-parameter-scale} DiT flow matching model for text-driven 3D human motion generation, producing realistic motion sequences across \textbf{200+ motion categories} spanning six major classes. Compared to prior million-parameter models, its scaling enables generalization to unseen motion compositions, while Flow-GRPO alignment further enhances motion naturalness and text–motion consistency. Collectively, these works demonstrate that trajectory-conditioned RL, geometry-verifiable supervision, flow-based motion discrimination, and large-scale flow matching together enable controllable, physically plausible motion generation across both video and 3D human motion spaces.

\paragraph{Video Editing.}
Extending Flow-GRPO-style alignment to video editing tasks, recent works incorporate editing accuracy and semantic reasoning into reinforcement-based optimization. \textbf{VIVA}~\cite{cong2025viva} adapts Flow-GRPO to the video editing setting by introducing \textbf{Edit-GRPO}, where the reward function jointly evaluates editing accuracy—whether the specified editing instruction is correctly executed—and temporal consistency—whether edited regions remain coherent across frames—thereby balancing local edit correctness with global temporal stability; moreover, VIVA constructs a synthetic data pipeline that automatically generates large-scale video editing training pairs, enabling scalable post-training for editing alignment. Complementarily, \textbf{ReViSE}~\cite{liu2025revise} proposes \textbf{reason-informed video editing}, in which the model first performs semantic analysis of editing instructions through a \textbf{self-reflective reasoning framework}, explicitly decomposing high-level instructions (e.g., ``make the sky look like sunset'') into underlying visual operations such as color tone adjustment, lighting modification, and cloud augmentation, and then executes the edit step-by-step guided by this reasoning process; this structured reasoning-enhanced editing improves the overall editing score by \textbf{32\%}. Together, these approaches demonstrate that effective video editing alignment requires joint optimization of instruction fidelity and temporal coherence, as well as explicit semantic decomposition of complex editing intents.

\paragraph{Video Understanding and Prediction.}
Extending reinforcement-based alignment toward reasoning and causality in video generation, recent works move beyond pure perceptual quality optimization. \textbf{Video-as-Answer}~\cite{cheng2025video} introduces a new video question-answering paradigm in which the model \textbf{generates video as the answer} rather than textual responses—for instance, generating a ball trajectory video to answer What would happen if the ball were kicked?''—and employs \textbf{Joint-GRPO} to simultaneously align the VLM responsible for question understanding and the video diffusion model responsible for answer generation. \textbf{What Happens Next}~\cite{li2025happens} addresses the \textbf{next-scene prediction} task, designing a \textbf{causal consistency reward} that enforces plausible causal relationships between predicted scenes and preceding video context. Complementarily, \textbf{RLIR}~\cite{ye2025reinforcement} leverages an \textbf{inverse dynamics model} to recover latent action sequences from generated videos, using these recovered actions as \textbf{verifiable rewards}—e.g., if the generated video claims a turning right'' action, the inverse model must infer a consistent directional change—improving action-following performance by \textbf{5--10\%}. Together, these methods highlight a shift from surface-level alignment toward causal reasoning, counterfactual prediction, and action-verifiable reward design in video generation.

\paragraph{Frontier Video Synthesis.}
Alongside alignment and reward-design advances, recent large-scale video systems push the frontier of model capacity, efficiency, and long-form generation. \textbf{FSVideo}~\cite{team2026fsvideo} is a 14B-parameter video generation system that adopts an extremely high compression-ratio latent space, reducing videos to highly compact latent representations and achieving inference speeds an order of magnitude faster than contemporary open-source models. \textbf{LongCat-Video}~\cite{team2025longcat}, a 13.6B-parameter long-video model, enables \textbf{minute-scale 720p generation} by leveraging \textbf{block sparse attention}, reducing computational complexity from $O(n^2)$ to $O(n \log n)$ and making long-context temporal modeling tractable. \textbf{Seedance 1.0}~\cite{gao2025seedance} integrates multi-dimensional RLHF—separately optimizing visual quality, motion naturalness, and text consistency—while achieving \textbf{10$\times$ inference acceleration} through engineering optimization; its successor, \textbf{Seedance 1.5 pro}~\cite{seedance2025seedance}, further introduces \textbf{native audio-visual joint generation}, simultaneously synthesizing visual content and synchronized audio with multilingual lip-sync capability. From an autoregressive perspective, \textbf{Self-Forcing++}~\cite{cui2025self} enables ultra-long video generation of \textbf{4 minutes 15 seconds}—over 50$\times$ longer than its teacher model—via autoregressive frame generation, overcoming traditional diffusion-based length limitations. Finally, \textbf{Video Text Preservation}~\cite{liu2025video} demonstrates that fine-tuning on synthetic text-rich videos substantially enhances video text rendering capability, highlighting the importance of data-centric scaling for multimodal consistency. Collectively, these systems illustrate that scaling latent compression, sparse attention, multi-objective RLHF, autoregressive extension, and data augmentation are key drivers of efficiency and long-horizon capability in modern video generation.

\paragraph{Video Reward Models.}
\textbf{VR-Thinker}~\cite{wang2025vr}, \textbf{SoliReward}~\cite{lian2025solireward}, \textbf{HuDA}~\cite{ashutosh2026human}, and \textbf{REACT}~\cite{wang2026thinking} have been discussed in detail in \S\ref{sec:reward_models}, where we analyze their respective reward modeling strategies and alignment mechanisms. Beyond explicit reward model design, \textbf{Reward Forcing}~\cite{lu2025reward} introduces a novel \textbf{distribution biasing} paradigm by modifying the DMD (Distribution Matching Distillation) objective to ``force'' the student model’s output distribution toward high-reward regions, resulting in Re-DMD (Reward-biased DMD) that integrates reward preference directly into distribution matching. Complementarily, \textbf{VQ-Insight}~\cite{zhang2025vq} enhances VLM-based AIGC video evaluation through \textbf{progressive visual RL}, where training gradually evolves from coarse binary preference judgments (good/bad) to fine-grained multi-dimensional quality assessment, enabling the critic to develop richer and more discriminative evaluation capability. Together, these approaches reflect a broader trend of integrating reward awareness into distribution matching and progressively strengthening visual evaluation models to support more reliable alignment.

% ============================================================================

\begin{table*}[!h]
\centering
\small
\caption{Representative GRPO-based extensions in image editing. 
Recent works emphasize reasoning-driven editing, dense reward shaping, 
motion- and geometry-aware control, scalable data construction, and paradigm-agnostic RL alignment.}
\setlength{\neuripsOverhang}{\dimexpr(\paperwidth-\textwidth)/2 -6mm\relax}
\begin{adjustwidth*}{-\neuripsOverhang}{-\neuripsOverhang}
\setlength{\tabcolsep}{4pt}
% \begin{tabularx}{\dimexpr\paperwidth-12mm\relax}{
\begin{tabularx}{\dimexpr\paperwidth-12mm\relax}{
@{\hspace{2mm}}
>{\raggedright\arraybackslash}p{0.16\linewidth}
>{\raggedright\arraybackslash}p{0.15\linewidth}
>{\raggedright\arraybackslash}p{0.32\linewidth}
>{\raggedright\arraybackslash}p{0.32\linewidth}
@{\hspace{4mm}}
}
\toprule
\textbf{Category} & \textbf{Method} & \textbf{Core Innovation} & \textbf{Key Contribution / Result} \\
\midrule

\multirow{4}{*}{Reasoning-Driven Editing}
& ThinkRL-Edit~\cite{li2026thinkrl} & CoT-based semantic analysis + binary checklist reward & Improves correctness–preservation balance \\
& Re-Align~\cite{he2026re} & In-Context CoT (IC-CoT) alignment & Example-guided reasoning during inference \\
& JarvisEvo~\cite{lin2025jarvisevo} & Editor–Evaluator alternating optimization (SEPO) & +18.95\% preservation; +44.96\% pixel fidelity \\
& Skywork UniPic 2.0~\cite{wei2025skywork} & Progressive Dual-Task Reinforcement (PDTR) & Joint image understanding + editing; Surpasses larger models \\

\midrule

\multirow{3}{*}{\shortstack[l]{Dense Reward \& \\Attention Control}}
& CogniEdit~\cite{li2025cogniedit} & Dense GRPO + Dynamic Token Focus Relocation & Region-aware attention for edit vs. preserve \\
& EditScore~\cite{luo2025editscore} & Structured editing reward evaluation & Interpretable multi-criterion scoring \\
& OmniRefiner~\cite{liu2025omnirefiner} & Detail-aware post-edit refinement & Reference-guided fine-grained enhancement \\

\midrule

\multirow{2}{*}{\shortstack[l]{Motion \& \\Geometry-Aware Editing}}
& MotionEdit~\cite{wan2025motionedit} & Motion alignment rewards & Physically plausible motion modifications \\
& Talk2Move~\cite{tan2026talk2move} & Text-instructed geometric transformation RL & Precise translation/rotation/scaling control \\

\midrule

\multirow{3}{*}{Application-Driven Editing}
& RePainter~\cite{guo2025repainter} & RL-based object removal + spatial matting & Seamless inpainting for e-commerce scenes \\
& Uniworld-V2 / Edit-R1~\cite{li2025uniworld} & DiffusionNFT-style efficient training + MLLM feedback & SOTA 4.49 on ImgEdit benchmark \\
& NPEdit~\cite{kumari2025learning} & No-paired-data editing training & Editing alignment using VLM-only feedback \\

\midrule

\multirow{2}{*}{\shortstack[l]{Scalable Data \& \\Curriculum}}
& Scaling Video Editing~\cite{bai2025scaling} & Ditto-1M dataset + Editto curriculum & Million-scale edit dataset; Progressive training \\
& Re-Align~\cite{he2026re} & Progressive edit complexity scheduling & Improved stability across edit types \\

\midrule

\multirow{2}{*}{\shortstack[l]{Paradigm-Agnostic \\RL Alignment}}
& EARL~\cite{zhao2025real} & RL for autoregressive image editing & Extends alignment beyond diffusion models \\
& Skywork UniPic 2.0~\cite{wei2025skywork} & Lightweight DiT + RL optimization & 2B model surpassing 7B/12B competitors \\

\bottomrule
\label{tab:editing}
\end{tabularx}
\end{adjustwidth*}
\end{table*}

\subsection{Image Editing}
\label{sec:editing}

Image editing is an important application scenario for GRPO alignment, where the core challenge lies in balancing editing accuracy (correctly executing instructions) and preservation (not modifying irrelevant regions). Recent works have introduced reasoning capabilities into the editing workflow, achieving significant quality improvements. The summary is shown in Table~\ref{tab:editing}, and the details are as follows.

Recent advances in RL-enhanced image and video editing increasingly emphasize reasoning, dense reward shaping, and scalable data construction. \textbf{ThinkRL-Edit}~\cite{li2026thinkrl} introduces a \textbf{reasoning-centric RL editing framework}, where the model first performs CoT-based semantic analysis of editing instructions (e.g., understanding that ``change the shirt to red'' requires modifying only color without altering shape), and optimizes using a \textbf{binary checklist} reward composed of independent yes/no questions evaluating correctness and preservation. \textbf{Re-Align}~\cite{he2026re} further proposes \textbf{In-Context CoT (IC-CoT) reasoning-guided alignment}, leveraging example editing cases during inference to guide current edits. \textbf{CogniEdit}~\cite{li2025cogniedit} combines \textbf{Dense GRPO optimization} with \textbf{Dynamic Token Focus Relocation}, reallocating attention during denoising to emphasize editable regions while protecting preserved areas. \textbf{JarvisEvo}~\cite{lin2025jarvisevo} presents \textbf{Synergistic Editor-Evaluator Policy Optimization (SEPO)}, alternating optimization between editor and evaluator modules, and—combined with multimodal CoT reasoning—achieves 18.95\% improvement in preservation editing and 44.96\% gain in pixel fidelity. For motion-aware edits, \textbf{MotionEdit}~\cite{wan2025motionedit} introduces \textbf{motion alignment rewards} to ensure plausibility in motion-modifying edits, while \textbf{Talk2Move}~\cite{tan2026talk2move} applies GRPO to optimize precision in \textbf{text-instructed geometric transformations} such as translation, rotation, and scaling. In application-driven scenarios, \textbf{RePainter}~\cite{guo2025repainter} focuses on e-commerce \textbf{object removal + spatial matting}, using RL to ensure seamless blending of filled regions, and \textbf{OmniRefiner}~\cite{liu2025omnirefiner} performs \textbf{detail-aware refinement} guided by reference images after primary generation. \textbf{Uniworld-V2/Edit-R1}~\cite{li2025uniworld} integrates DiffusionNFT-style efficient training with MLLM implicit feedback, achieving a \textbf{SOTA score of 4.49} on ImgEdit, while \textbf{NPEdit}~\cite{kumari2025learning} enables \textbf{no-paired-data editing training} using only VLM feedback. At scale, \textbf{Scaling Video Editing}~\cite{bai2025scaling} releases the million-scale \textbf{Ditto-1M} dataset and proposes the \textbf{Editto} curriculum from simple to complex edits. \textbf{EditScore}~\cite{luo2025editscore} has been discussed in \S\ref{sec:reward_models}. Broadening model paradigms, \textbf{EARL}~\cite{zhao2025real} applies RL to \textbf{autoregressive image editing}, demonstrating that alignment is not confined to diffusion-based models, and \textbf{Skywork UniPic 2.0}~\cite{wei2025skywork}, a 2B-parameter DiT model, leverages \textbf{Progressive Dual-Task Reinforcement (PDTR)} to jointly optimize image understanding and editing, surpassing larger models such as BAGEL 7B and Flux-Kontext 12B. Collectively, these works indicate a shift toward reasoning-driven editing, dense and interpretable rewards, motion- and geometry-aware control, scalable dataset construction, and paradigm-agnostic RL alignment.

% ============================================================================
\begin{table*}[!h]
\centering
\small
\caption{Representative extensions of GRPO alignment in speech and audio generation. 
These works demonstrate the modality-agnostic nature of GRPO in continuous-time 
signal modeling, multi-metric optimization, and style-preserving generation.}
\setlength{\neuripsOverhang}{\dimexpr(\paperwidth-\textwidth)/2 -6mm\relax}
\begin{adjustwidth*}{-\neuripsOverhang}{-\neuripsOverhang}
\setlength{\tabcolsep}{4pt}
% \begin{tabularx}{\dimexpr\paperwidth-12mm\relax}{
\begin{tabularx}{\dimexpr\paperwidth-12mm\relax}{
@{\hspace{2mm}}
>{\raggedright\arraybackslash}p{0.16\linewidth}
>{\raggedright\arraybackslash}p{0.15\linewidth}
>{\raggedright\arraybackslash}p{0.32\linewidth}
>{\raggedright\arraybackslash}p{0.32\linewidth}
@{\hspace{4mm}}
}
\toprule
\textbf{Category} & \textbf{Method} & \textbf{Core Innovation} & \textbf{Key Contribution / Result} \\
\midrule

\multirow{1}{*}{Speech Enhancement}
& FlowSE-GRPO~\cite{wang2026flowse} 
& Online GRPO for flow-matching speech enhancement with multi-metric reward balancing 
& Joint optimization of PESQ, STOI, and SI-SDR within unified RL framework \\

\midrule

\multirow{2}{*}{\shortstack[l]{Music \& \\Singing Generation}}
& YingMusic-Singer~\cite{zheng2025yingmusic} 
& Multi-objective GRPO for zero-shot singing voice synthesis 
& Style-consistent singing from lyrics + melody without target-speaker training data \\
& YingMusic-SVC~\cite{chen2025yingmusic} 
& GRPO-based zero-shot singing voice conversion with singing-specific inductive biases 
& Timbre conversion preserving lyrics and melody; Reward models formant and vibrato structure \\

\bottomrule
\label{tab:speech}

\end{tabularx}
\end{adjustwidth*}
\end{table*}
\subsection{Speech and Audio}
\label{sec:speech}

GRPO alignment extends beyond the visual modality, and its core principles naturally generalize to continuous-time audio signal generation. As show in Table~\ref{tab:speech}, \textbf{FlowSE-GRPO}~\cite{wang2026flowse} is the first work integrating online GRPO into flow matching speech enhancement (SE), addressing challenges unique to audio generation: speech signals are continuous time series rather than fixed-size images, and quality evaluation requires simultaneously optimizing multiple potentially conflicting metrics, including speech quality (PESQ), speech intelligibility (STOI), and signal-to-distortion ratio (SI-SDR); accordingly, FlowSE-GRPO designs a \textbf{multi-metric reward optimization} scheme that adaptively balances these objectives within a unified reinforcement learning framework. Extending Flow-GRPO to music generation, \textbf{YingMusic-Singer}~\cite{zheng2025yingmusic} applies a multi-objective reward formulation to \textbf{zero-shot singing voice synthesis}, generating singing in a specific singer’s style from lyrics and melody without using training data from the target singer, where the reward evaluates pitch accuracy, rhythm alignment, and timbre similarity. Similarly, \textbf{YingMusic-SVC}~\cite{chen2025yingmusic} adapts Flow-GRPO to \textbf{zero-shot singing voice conversion (SVC)}, preserving original lyrics and melody while converting the singer’s timbre to a target identity, and introduces singing-specific inductive biases—such as formant structure and vibrato patterns—so that GRPO rewards can capture fine-grained perceptual qualities of singing. Collectively, these works demonstrate that GRPO-based alignment provides a modality-agnostic optimization paradigm capable of handling continuous signals, multi-metric trade-offs, and style-preserving generation in the audio domain.

% ============================================================================
\begin{table*}[!h]
\centering
\small
\caption{Extensions of GRPO alignment in 3D generation and scientific foundation models. 
These works demonstrate hierarchical optimization, cross-dimensional reward transfer, 
efficient 3D representations, and physically grounded reinforcement objectives.}
\setlength{\neuripsOverhang}{\dimexpr(\paperwidth-\textwidth)/2 -6mm\relax}
\begin{adjustwidth*}{-\neuripsOverhang}{-\neuripsOverhang}
\setlength{\tabcolsep}{4pt}
% \begin{tabularx}{\dimexpr\paperwidth-12mm\relax}{
\begin{tabularx}{\dimexpr\paperwidth-12mm\relax}{
@{\hspace{2mm}}
>{\raggedright\arraybackslash}p{0.16\linewidth}
>{\raggedright\arraybackslash}p{0.15\linewidth}
>{\raggedright\arraybackslash}p{0.32\linewidth}
>{\raggedright\arraybackslash}p{0.32\linewidth}
@{\hspace{4mm}}
}
\toprule
\textbf{Category} & \textbf{Method} & \textbf{Core Innovation} & \textbf{Key Contribution / Result} \\
\midrule

\multirow{4}{*}{3D Generation}
& Hi-GRPO / AR3D-R1~\cite{tang2025we} 
& Hierarchical coarse-to-fine RL optimization for text-to-3D 
& Progressive global shape $\rightarrow$ local texture refinement; Improved multi-view consistency \\
& Nabla-R2D3~\cite{liu2025nabla} 
& Cross-dimensional reward transfer (2D $\rightarrow$ 3D) 
& Multi-view rendering + 2D reward averaging for 3D alignment \\
& VIST3A~\cite{go2025vist3a} 
& Video-to-3D model stitching 
& Multi-view consistent video frames enable coherent 3D reconstruction \\
& TRIM~\cite{yin2025trim} 
& 3D Gaussian Splatting acceleration 
& Efficient 3D diffusion with preserved visual fidelity \\

\midrule

\multirow{3}{*}{Scientific Applications}
& Open Materials Generation~\cite{hoellmer2026open} 
& RL-guided crystal structure search 
& Thermodynamic stability + symmetry-aware reward via DFT energy \\
& Elign / FED-GRPO~\cite{li2026elign} 
& Force-Energy Disentangled GRPO for molecular modeling 
& Separate rewards for force accuracy and energy smoothness \\
& Flow Marching~\cite{chen2025flow} 
& Flow-matching PDE foundation modeling 
& RL reduces long-horizon rollout drift in dynamical systems \\

\bottomrule
\label{tab:3d_science}
\end{tabularx}
\end{adjustwidth*}
\end{table*}
\subsection{3D Generation and Scientific Applications}
\label{sec:3d_science}

\paragraph{3D Generation.}
GRPO-based alignment has also begun extending into 3D generation, where maintaining multi-view consistency and structural coherence introduces challenges beyond 2D synthesis. As show in Table~\ref{tab:3d_science}, \textbf{Hi-GRPO/AR3D-R1}~\cite{tang2025we} is the first work incorporating RL enhancement into text-to-3D generation, recognizing that 3D objects must remain consistent across all viewpoints; it proposes a \textbf{hierarchical optimization} strategy that first optimizes global shape and structural geometry (low-frequency components) and subsequently refines local details and textures (high-frequency components), enabling progressive coarse-to-fine improvement of 3D quality. \textbf{Nabla-R2D3}~\cite{liu2025nabla} addresses the difficulty of training dedicated 3D reward models by leveraging \textbf{2D reward models to align 3D diffusion models}: 3D objects are rendered from multiple viewpoints into 2D images, each view is independently evaluated by a 2D reward model, and the averaged reward is propagated back to guide 3D optimization, effectively reducing 3D alignment to multi-view 2D supervision. Bridging video and 3D, \textbf{VIST3A}~\cite{go2025vist3a} employs \textbf{model stitching} to connect a video generator with a 3D reconstruction module, where multi-view-consistent video frames produced by the generator are used to recover coherent 3D structures. Finally, \textbf{TRIM}~\cite{yin2025trim} accelerates 3D diffusion generation by adopting 3D Gaussian Splatting representations, significantly improving efficiency while preserving quality. Collectively, these works demonstrate that hierarchical RL optimization, cross-dimensional reward transfer, model stitching, and efficient 3D representations are key enablers for extending GRPO alignment into the 3D domain.

\paragraph{Scientific Applications.}
Beyond perceptual media generation, GRPO-based alignment has begun influencing scientific foundation models operating over structured physical domains. \textbf{Open Materials Generation}~\cite{hoellmer2026open} is the first work applying RL alignment to \textbf{crystal structure prediction}, where the generative model proposes candidate crystal structures and the reward function evaluates thermodynamic stability—via formation energy computed using DFT—as well as symmetry constraints, enabling reinforcement learning to guide structure search toward physically plausible and energetically favorable configurations. \textbf{Elign/FED-GRPO}~\cite{li2026elign} further extends GRPO to \textbf{equivariant diffusion models} for molecular force field computation, introducing \textbf{Force-Energy Disentangled GRPO (FED-GRPO)} that designs independent reward components for force field accuracy and potential energy surface smoothness, thereby mitigating optimization conflicts between force prediction precision and energy consistency. At the level of scientific simulation, \textbf{Flow Marching}~\cite{chen2025flow} adapts the flow matching paradigm to \textbf{PDE (partial differential equation) foundation models}, addressing the long-term error accumulation problem of traditional numerical solvers by learning a unified velocity field that models temporal evolution, with RL optimization reducing rollout drift over extended simulation horizons. Collectively, these works demonstrate that GRPO-style alignment can generalize to physically constrained generative systems, where reward signals derive from thermodynamic principles, symmetry equivariance, or long-horizon dynamical consistency.

% ============================================================================
\begin{table*}[!h]
\centering
\small
\caption{Extensions of Flow-GRPO style in Vision-Language-Action (VLA) and embodied AI. 
These works address continuous high-dimensional action spaces, real-time control, 
simulator-based learning, structured credit assignment, and socially aware policy optimization.}
\setlength{\neuripsOverhang}{\dimexpr(\paperwidth-\textwidth)/2 -6mm\relax}
\begin{adjustwidth*}{-\neuripsOverhang}{-\neuripsOverhang}
\setlength{\tabcolsep}{4pt}
% \begin{tabularx}{\dimexpr\paperwidth-12mm\relax}{
\begin{tabularx}{\dimexpr\paperwidth-12mm\relax}{
@{\hspace{2mm}}
>{\raggedright\arraybackslash}p{0.16\linewidth}
>{\raggedright\arraybackslash}p{0.15\linewidth}
>{\raggedright\arraybackslash}p{0.32\linewidth}
>{\raggedright\arraybackslash}p{0.32\linewidth}
@{\hspace{4mm}}
}
\toprule
\textbf{Category} & \textbf{Method} & \textbf{Core Innovation} & \textbf{Key Contribution / Result} \\
\midrule

\multirow{3}{*}{\shortstack[l]{Spatially-Grounded \\Flow Policies}}
& SA-VLA~\cite{pan2026sa} 
& Spatially-Aware Flow Matching with 3D cues (depth, normals) 
& SCAN exploration strategy for efficient action-space exploration \\
& $\pi$RL~\cite{chen2025pirl} 
& Flow-Noise vs. Flow-SDE RL integration 
& Stability–exploration trade-off analysis in flow-based VLA \\
& PolicyFlow~\cite{yang2026policyflow} 
& Continuous normalizing flow (CNF) policies + Brownian regularizer 
& Structured stochastic exploration under PPO optimization \\

\midrule

\multirow{2}{*}{\shortstack[l]{Structured \\Policy Optimization}}
& FPO~\cite{lyu2025reinforcement} 
& Structure-aware credit assignment 
& Hierarchical reward decomposition + Q-ensemble estimation \\
& Flow Matching Policy Gradients~\cite{mcallister2025flow} 
& Theoretical PPO compatibility proof 
& Validates flow matching as RL policy parameterization \\

\midrule

\multirow{2}{*}{\shortstack[l]{Simulator-Based \& \\World Models}}
& ProphRL~\cite{zhang2025reinforcing} 
& Action-to-video world model for few-shot simulation 
& 5--17\% sim gain; 24--30\% real-robot improvement via FA-GRPO \\
& EmboMatrix / EmboBrain~\cite{lei2025embomatrix} 
& Large-scale embodied decision benchmark + multi-task RL agent 
& Unified navigation, manipulation, and interaction evaluation \\

\midrule

\multirow{2}{*}{Real-Time Control}
& DMPO~\cite{zou2026one} 
& Single-step action generation at $>$120Hz 
& Dispersive regularization prevents policy mode collapse \\
& FPMD~\cite{chen2025one} 
& One-step inference from multi-step flow policies 
& Achieves 1-step execution without distillation \\

\midrule

\multirow{1}{*}{Socially-Aware RL}
& SocialNav~\cite{chen2025socialnav} 
& SAFE-GRPO for social constraint encoding 
& +38\% navigation success; +46\% social compliance \\

\bottomrule
\label{tab:vla}
\end{tabularx}
\end{adjustwidth*}
\end{table*}

\subsection{VLA and Embodied AI}
\label{sec:vla}
Extending Flow-GRPO to Vision-Language-Action (VLA) models and embodied intelligence introduces the additional difficulty that action spaces are continuous, high-dimensional, and must support real-time interaction with physical environments. As show in Table~\ref{tab:vla}, \textbf{SA-VLA}~\cite{pan2026sa} proposes \textbf{Spatially-Aware Flow Matching} for robotic manipulation, incorporating explicit 3D spatial cues such as depth and surface normals into the flow matching condition to produce spatially grounded action trajectories, and introduces the \textbf{SCAN exploration strategy}, which injects structured random perturbations into action space to enhance exploration efficiency. \textbf{$\pi$RL}~\cite{chen2025pirl} investigates two RL integration strategies for flow-based VLA policies: \textbf{Flow-Noise}, where stochasticity is injected only in the initial noise while subsequent evolution remains deterministic (ODE), and \textbf{Flow-SDE}, where randomness is introduced at every step; experiments reveal a trade-off, with Flow-Noise offering greater stability and Flow-SDE providing stronger exploration. \textbf{FPO (Flow Policy Optimization)}~\cite{lyu2025reinforcement} introduces \textbf{structure-aware credit assignment}, leveraging hierarchical task decomposition to distribute task-level rewards to individual action steps, and combines this with \textbf{Q-ensemble value estimation} for more robust policy updates. \textbf{ProphRL}~\cite{zhang2025reinforcing} trains an \textbf{action-to-video world model} that predicts future visual observations from current states and action sequences, serving as a \textbf{few-shot simulator} to drastically reduce real-world interaction cost; coupled with \textbf{FA-GRPO (Flow-Aware GRPO)} for policy optimization on simulated rollouts, it achieves 5--17\% improvement in simulation and 24--30\% gains on real robots. In pursuit of real-time control, \textbf{DMPO}~\cite{zou2026one} enables \textbf{true single-step action generation} at inference frequencies exceeding \textbf{120Hz}, employing \textbf{dispersive regularization} to prevent policy mode collapse and preserve action diversity under environmental uncertainty, while \textbf{FPMD}~\cite{chen2025one} derives 1-step inference capability directly from multi-step flow policies \textbf{without distillation}. From a theoretical standpoint, \textbf{Flow Matching Policy Gradients}~\cite{mcallister2025flow} proves compatibility between flow matching policies and PPO-clip, establishing flow matching as a valid policy parameterization for standard RL algorithms. Beyond single-agent manipulation, \textbf{SocialNav}~\cite{chen2025socialnav} applies GRPO to \textbf{social navigation}, introducing \textbf{SAFE-GRPO} to encode social constraints such as personal space preservation alongside efficiency objectives, improving success rate by \textbf{38\%} and social compliance by \textbf{46\%}. At scale, \textbf{EmboMatrix}~\cite{lei2025embomatrix} constructs a large unified embodied decision-making benchmark spanning navigation, manipulation, and interaction, and releases \textbf{EmboBrain}, a multi-task RL-trained embodied agent. Finally, \textbf{PolicyFlow}~\cite{yang2026policyflow} parameterizes action policies as \textbf{continuous normalizing flows (CNF)}, integrates PPO optimization, and introduces a \textbf{Brownian regularizer} to inject structured stochasticity during policy updates, promoting exploration while maintaining flow-based structure. Collectively, these works demonstrate that flow-based RL alignment extends naturally to embodied agents, enabling spatial grounding, structured credit assignment, simulator-based policy learning, real-time control, theoretical RL compatibility, and socially aware behavior in continuous action domains.

% ============================================================================
\begin{table*}[!h]
\centering
\small
\caption{Representative unified multimodal models aligned with GRPO-style optimization. 
We categorize existing works into unified RL optimization, hybrid/parallel architectures, 
reasoning-augmented generation, efficiency-oriented structural design, 
and auxiliary constraint / system-level scaling.}
\setlength{\neuripsOverhang}{\dimexpr(\paperwidth-\textwidth)/2 -6mm\relax}
\begin{adjustwidth*}{-\neuripsOverhang}{-\neuripsOverhang}
\setlength{\tabcolsep}{4pt}
% \begin{tabularx}{\dimexpr\paperwidth-12mm\relax}{
\begin{tabularx}{\dimexpr\paperwidth-12mm\relax}{
@{\hspace{2mm}}
>{\raggedright\arraybackslash}p{0.16\linewidth}
>{\raggedright\arraybackslash}p{0.15\linewidth}
>{\raggedright\arraybackslash}p{0.32\linewidth}
>{\raggedright\arraybackslash}p{0.32\linewidth}
@{\hspace{4mm}}
}
\toprule
\textbf{Category} & \textbf{Method} & \textbf{Core Innovation} & \textbf{Key Contribution / Result} \\
\midrule

\multirow{3}{*}{Unified RL Optimization}
& UAE~\cite{yan2025unified} 
& Auto-encoder unification (encode=understanding, decode=generation) + Unified-GRPO 
& Joint optimization of VQA and preference rewards; Bidirectional enhancement \\

& UniRL-Zero~\cite{wang2025unirl} 
& Single RL algorithm for language + diffusion experts 
& Eliminates modality-decoupled training; Unified policy optimization \\

& Unified T2I (Weakness-Targeted PT)~\cite{chen2026unified} 
& Weakness-targeted post-training reinforcement 
& Automatic strengthening of underperforming prompt categories \\

\midrule

\multirow{3}{*}{\shortstack[l]{Hybrid / \\Parallel Architectures}}
& BLIP3o-NEXT~\cite{chen2025blip3o} 
& AR (planning) + Diffusion (generation) hybrid framework 
& Open-source multimodal frontier with RL-enhanced generation \\

& ParaUni~\cite{tan2025parauni} 
& Hierarchical parallel VLM interaction + LDAM gradient balancing 
& Cross-layer bidirectional information coupling \\

& MMaDA-Parallel~\cite{tian2025mmada} 
& Parallel multimodal diffusion with cross-attention; ParaRL training 
& Simultaneous denoising of text and image streams \\

\midrule

\multirow{3}{*}{\shortstack[l]{Reasoning-Augmented \\Generation}}
& Visual-Aware CoT~\cite{ye2025visual} 
& Adaptive visual planning + iterative correction 
& Intermediate visual reasoning refinement during generation \\

& SEER~\cite{tang2026endogenous} 
& Endogenous reprompting for self-evolving cognitive alignment 
& Bridges reasoning gap between understanding and generation \\

& Visual Generation for Reasoning~\cite{wu2026visual} 
& Interleaved visual–verbal CoT (Visual Superiority Hypothesis) 
& Generation-assisted reasoning performance improvement \\

\midrule

\multirow{3}{*}{\shortstack[l]{Efficiency / \\Structural Design}}
& MoS~\cite{liu2025mixture} 
& Learnable token-level router for sparse task allocation 
& Efficient computation balancing between modalities \\

& MANZANO~\cite{modelmanzano} 
& Hybrid vision tokenizer design 
& Reduced understanding-generation task interference \\

& FUDOKI~\cite{wang2025fudoki} 
& Pure discrete flow matching with unified token space 
& Homogeneous multimodal generation framework \\

\midrule

\multirow{3}{*}{\shortstack[l]{Auxiliary Constraints \& \\ System Scaling}}
& COOPER~\cite{zhang2025cooper} 
& Auxiliary depth + segmentation supervision 
& Enhanced geometric and spatial reasoning consistency \\

& ShowTable~\cite{liu2025showtable} 
& MLLM–diffusion collaborative generation 
& Information-preserving multimodal table visualization \\

& Seedream 4.0~\cite{seedream2025seedream} 
& Native 1K--4K high-resolution unified generation 
& Production-scale multimodal system integration \\

\bottomrule
\label{tab:unified}
\end{tabularx}
\end{adjustwidth*}
\end{table*}

\subsection{Unified Multimodal Models}
\label{sec:unified}
Unified multimodal models seek to jointly support visual understanding and visual generation within a single architecture, with GRPO serving as a key alignment mechanism to reconcile bidirectional objectives. As show in Table~\ref{tab:unified}, \textbf{UAE}~\cite{yan2025unified} formulates unification from an \textbf{auto-encoder perspective}, treating visual understanding as encoding (image $\to$ semantics) and generation as decoding (semantics $\to$ image), and introduces \textbf{Unified-GRPO} to simultaneously optimize understanding rewards (e.g., VQA accuracy) and generation rewards (e.g., human preference), enabling bidirectional enhancement. \textbf{BLIP3o-NEXT}~\cite{chen2025blip3o} adopts a fully open-source \textbf{AR + Diffusion} hybrid architecture, using the autoregressive module for understanding and planning and diffusion for high-fidelity generation, with RL optimization pushing generation performance to the frontier. \textbf{UniRL-Zero}~\cite{wang2025unirl} proposes \textbf{unified RL}, jointly optimizing language and diffusion experts under a single RL algorithm rather than separate training. Architecturally, \textbf{ParaUni}~\cite{tan2025parauni} introduces \textbf{hierarchical parallel information interaction} between VLM layers, enabling intermediate representations of understanding and generation to influence each other, combined with \textbf{Layer-wise Dynamic Adjustment Mechanism (LDAM)} to balance gradient flows during RL training; similarly, \textbf{MMaDA-Parallel}~\cite{tian2025mmada} proposes \textbf{parallel multimodal diffusion}, where text and images are denoised simultaneously and interact through cross-attention, and \textbf{ParaRL} provides a dedicated RL training strategy. From a reasoning perspective, \textbf{Visual-Aware CoT}~\cite{ye2025visual} introduces \textbf{adaptive visual planning} and \textbf{iterative visual correction}, allowing models to generate and refine intermediate visual reasoning steps, while \textbf{SEER}~\cite{tang2026endogenous} proposes \textbf{self-evolving cognitive alignment} via endogenous reprompting to bridge the cognitive gap between understanding and generation. \textbf{Visual Generation for Reasoning}~\cite{wu2026visual} advances the \textbf{Visual Superiority Hypothesis}, advocating \textbf{interleaved visual-verbal CoT} that alternates textual reasoning and visual generation for improved reasoning performance. Efficiency and structural design innovations include \textbf{MoS}~\cite{liu2025mixture}, which employs a learnable token-level router to sparsely allocate computation between understanding and generation; \textbf{MANZANO}~\cite{modelmanzano}, which introduces a hybrid vision tokenizer to reduce task conflicts; and \textbf{FUDOKI}~\cite{wang2025fudoki}, which unifies text and images as discrete tokens under pure discrete flow matching. At the system level, \textbf{Seedream 4.0}~\cite{seedream2025seedream} supports seamless \textbf{1K--4K native high-resolution} generation, and \textbf{Unified T2I with Weakness-Targeted PT}~\cite{chen2026unified} proposes \textbf{weakness-targeted post-training} to automatically reinforce underperforming prompt categories. \textbf{COOPER}~\cite{zhang2025cooper} enhances spatial reasoning by introducing auxiliary depth and segmentation outputs to constrain geometric consistency, and \textbf{ShowTable}~\cite{liu2025showtable} demonstrates multimodal collaboration between MLLM and diffusion models for creative, information-preserving table visualization. Collectively, these works illustrate that GRPO-enabled unified models rely on bidirectional reward optimization, parallel architectural interaction, visual reasoning augmentation, adaptive routing, auxiliary modality constraints, and targeted post-training to bridge understanding and generation within a single multimodal framework.

% ============================================================================
\begin{table*}[!h]
\centering
\small
\caption{Extensions of GRPO alignment to autoregressive (AR), masked diffusion, 
and discrete flow matching models. These works explore token-level credit assignment, 
adaptive regularization, masking policy optimization, search-based decoding, 
and distribution-level reinforcement learning.}
\setlength{\neuripsOverhang}{\dimexpr(\paperwidth-\textwidth)/2 -6mm\relax}
\begin{adjustwidth*}{-\neuripsOverhang}{-\neuripsOverhang}
\setlength{\tabcolsep}{4pt}
% \begin{tabularx}{\dimexpr\paperwidth-12mm\relax}{
\begin{tabularx}{\dimexpr\paperwidth-12mm\relax}{
@{\hspace{2mm}}
>{\raggedright\arraybackslash}p{0.16\linewidth}
>{\raggedright\arraybackslash}p{0.15\linewidth}
>{\raggedright\arraybackslash}p{0.32\linewidth}
>{\raggedright\arraybackslash}p{0.32\linewidth}
@{\hspace{4mm}}
}
\toprule
\textbf{Category} & \textbf{Method} & \textbf{Core Innovation} & \textbf{Key Contribution / Result} \\
\midrule

\multirow{6}{*}{\shortstack[l]{Autoregressive \\ Image Generation}}
& AR-GRPO~\cite{yuan2025ar} 
& Token-level GRPO for AR image generation 
& Structured reward optimization over discrete visual tokens \\
& STAGE~\cite{ma2025stage} 
& Similarity-aware KL/advantage reweighting 
& Adaptive KL scaling + entropy reward to prevent collapse \\
& VAR RL Done Right~\cite{sun2026var} 
& VMR + PANW + Mask Propagation 
& Token-level return estimation + position-sensitive normalization \\
& GCPO~\cite{zhang2025group} 
& Critical-token selective optimization 
& RL applied to top 30\% important tokens only \\
& Layout-Conditioned AR T2I~\cite{zheng2025layout} 
& Layout-aware masking + spatial rewards 
& Improved spatial controllability in AR generation \\
& AR-Drag~\cite{zhao2025real} 
& RL-enhanced autoregressive trajectory control 
& Extends AR alignment to video/motion generation \\

\midrule

\multirow{3}{*}{Masked Diffusion Models}
& Co-GRPO~\cite{zhou2025co} 
& Joint optimization of model + inference schedule 
& Treats mask schedule as MDP action variable \\
& Mask-GRPO~\cite{luo2025reinforcement} 
& RL over unmasking trajectory 
& Sequential mask selection aligned with reward \\
& MaskGRPO (Multimodal)~\cite{ma2025consolidating} 
& Multimodal discrete masking + RL alignment 
& Unified text-image masking optimization \\

\midrule

\multirow{5}{*}{\shortstack[l]{Discrete Diffusion / \\Flow Matching}}
& Discrete Guidance Matching~\cite{wan2025discrete} 
& Exact reward-consistent discrete guidance 
& Enables guidance in non-differentiable token space \\
& TR2-D2~\cite{tang2025tr2} 
& MCTS-based unmasking search 
& Trajectory-aware fine-tuning via discovered paths \\
& DMPO (Discrete)~\cite{zou2026one} 
& Distribution-level policy optimization 
& 42.9\% accuracy improvement via divergence minimization \\
& LaDiR~\cite{kang2025ladir} 
& Latent diffusion reasoning for LLMs 
& Continuous latent reasoning before discrete decoding \\
& LaDi-RL~\cite{kang2026beyond} 
& Reward-driven latent diffusion steering 
& RL-guided drift in latent space with diversity preservation \\

\bottomrule
\label{tab:ar}
\end{tabularx}
\end{adjustwidth*}
\end{table*}
\subsection{Autoregressive and Masked Diffusion Models}
\label{sec:ar}

GRPO was originally designed for autoregressive LLMs, and bringing it back to autoregressive visual models and masked diffusion models represents a natural extension.
The summary is shown in Table~\ref{tab:ar}, and the details are as follows.

\paragraph{Autoregressive Image Generation.}
Reinforcement alignment has also been extended to autoregressive (AR) image generation models, where images are generated token-by-token (e.g., via VQVAE discrete tokens) and policy optimization operates at the token level. \textbf{AR-GRPO}~\cite{yuan2025ar} is the first work adapting GRPO to \textbf{autoregressive image generation}, optimizing perceptual quality, realism, and semantic alignment through structured reward categories. \textbf{STAGE}~\cite{ma2025stage} observes that standard KL divergence regularization is overly coarse for AR image models and introduces \textbf{similarity-aware advantage/KL reweighting}, dynamically scaling KL penalties according to the semantic similarity between generated samples and the reference distribution; it further incorporates \textbf{entropy-based rewards} to maintain sufficient generation entropy and prevent mode collapse. For VAR (Visual Autoregressive) models, \textbf{VAR RL Done Right}~\cite{sun2026var} proposes three key techniques: \textbf{Value as Middle Return (VMR)} to provide token-level return estimation, \textbf{Per-Action Normalization Weighting (PANW)} to apply position-sensitive normalization schemes, and \textbf{Mask Propagation (MP)} to block gradients from masked regions. \textbf{GCPO}~\cite{zhang2025group} reveals that tokens in AR image models are not equally important and identifies \textbf{critical tokens} based on \textbf{causal dependency}, \textbf{entropy}, and \textbf{diversity} criteria, achieving superior performance by applying RL optimization to only 30\% of tokens. \textbf{Layout-Conditioned AR T2I}~\cite{zheng2025layout} enhances spatial controllability by injecting layout conditions and \textbf{structured masks} into AR generation, combined with \textbf{GRPO layout rewards} to enforce spatial relationship correctness. Finally, \textbf{AR-Drag}~\cite{zhao2025real}, discussed in \S\ref{sec:video}, demonstrates that RL-enhanced autoregressive policies can also extend to video generation and motion control. Collectively, these works show that token-level credit assignment, adaptive regularization, selective optimization, structured masking, and layout-aware rewards are central to stabilizing and enhancing RL alignment in autoregressive generative models.

\paragraph{Masked Diffusion Models.}
Reinforcement-based alignment has also been extended to masked generative paradigms, where inference schedules and masking policies play a central role in generation quality. \textbf{Co-GRPO}~\cite{zhou2025co} formulates a unified MDP framework for \textbf{masked diffusion models (MDM)}, recognizing that inference schedules—such as mask ratio annealing strategies—critically affect performance; it therefore \textbf{jointly models model parameters and inference schedules as components of the MDP action space}, enabling GRPO to simultaneously optimize both network weights and decoding dynamics. Complementarily, \textbf{Mask-GRPO}~\cite{luo2025reinforcement} is the first work to directly introduce GRPO into masked generative models by casting the mask prediction process as sequential decision-making, where each action corresponds to selecting positions to unmask at each step, thereby aligning the unmasking trajectory itself with reward objectives. Extending this paradigm to broader settings, \textbf{MaskGRPO (Multimodal)}~\cite{ma2025consolidating} adapts the framework to \textbf{multimodal discrete diffusion}, jointly handling text and image discrete tokens within a unified masking and RL optimization process, enabling scalable multimodal reinforcement alignment. Collectively, these works demonstrate that masking strategies and inference schedules can be elevated to first-class optimization variables under GRPO, broadening reinforcement alignment beyond parameter updates to include generative decoding policies.

\paragraph{Discrete Diffusion / Flow Matching.}
Recent advances further extend flow and diffusion alignment into discrete and language-centric settings, addressing challenges unique to non-differentiable token spaces. \textbf{Discrete Guidance Matching}~\cite{wan2025discrete} tackles the \textbf{exact guidance} problem in discrete flow matching, where traditional continuous guidance techniques (e.g., Classifier Guidance) are inapplicable due to non-differentiability; it introduces a discretized guidance matching objective that enables exact reward-consistent guidance directly in discrete space. \textbf{TR2-D2}~\cite{tang2025tr2} incorporates \textbf{MCTS (Monte Carlo Tree Search)} into discrete diffusion by treating the unmasking process as a search tree, using MCTS to discover optimal unmasking trajectories and subsequently applying \textbf{trajectory-aware fine-tuning} based on these discovered paths. For discrete diffusion LLMs, \textbf{DMPO}~\cite{zou2026one} proposes \textbf{Distribution Matching Policy Optimization}, replacing per-token policy gradients with distribution-level divergence minimization between generated and target distributions, yielding a substantial \textbf{42.9\%} accuracy improvement. Bridging continuous and discrete reasoning, \textbf{LaDiR}~\cite{kang2025ladir} introduces \textbf{latent diffusion} into LLM text reasoning by performing diffusion-style reasoning in continuous latent space before mapping back to discrete tokens, leveraging the smoothness of continuous dynamics for improved reasoning trajectories. \textbf{LaDi-RL}~\cite{kang2026beyond} further integrates RL into this latent diffusion framework, applying reward-driven drift in latent space to steer discrete outputs toward higher quality while maintaining diversity. Collectively, these works demonstrate that discrete diffusion and language generation can benefit from exact discrete guidance, search-based trajectory optimization, distribution-level policy learning, and latent-space RL steering, effectively bridging continuous flow principles with discrete token generation.

% ============================================================================
\begin{table*}[!h]
\centering
\small
\caption{Extensions of GRPO-based alignment in image restoration and super-resolution. 
These works address blind degradation, hierarchical restoration, spatially adaptive weighting, 
language-guided quality assessment, and hybrid efficiency-oriented enhancement architectures.}
\setlength{\neuripsOverhang}{\dimexpr(\paperwidth-\textwidth)/2 -6mm\relax}
\begin{adjustwidth*}{-\neuripsOverhang}{-\neuripsOverhang}
\setlength{\tabcolsep}{4pt}
% \begin{tabularx}{\dimexpr\paperwidth-12mm\relax}{
\begin{tabularx}{\dimexpr\paperwidth-12mm\relax}{
@{\hspace{2mm}}
>{\raggedright\arraybackslash}p{0.16\linewidth}
>{\raggedright\arraybackslash}p{0.15\linewidth}
>{\raggedright\arraybackslash}p{0.32\linewidth}
>{\raggedright\arraybackslash}p{0.32\linewidth}
@{\hspace{4mm}}
}
\toprule
\textbf{Category} & \textbf{Method} & \textbf{Core Innovation} & \textbf{Key Contribution / Result} \\
\midrule

\multirow{2}{*}{Blind Restoration}
& LRPO~\cite{bin2025lrpo} 
& Online RL for blind face restoration with composite reward 
& Identity + perceptual + degradation removal reward; likelihood regularization \\
& Diffusion Restoration via RL~\cite{xu2025enhancing} 
& MLLM-based IQA reward + RL/SFT hybrid training 
& Difficulty-adaptive switching between SFT and RL \\

\midrule

\multirow{2}{*}{Hierarchical Restoration}
& Pref-Restore~\cite{yao2026bridging} 
& Two-stage coarse-to-fine restoration framework 
& Stage-specific RL preference constraints within diffusion loop \\
& FinPercep-RM~\cite{liu2025finpercep} 
& Spatially-aware degradation maps + co-evolutionary curriculum 
& Region-weighted rewards for heavily corrupted areas \\

\midrule

\multirow{2}{*}{\shortstack[l]{Efficiency \& \\Hybrid Enhancement}}
& FlowLUT~\cite{hu2025flowlut} 
& LUT-based global adjustment + flow matching refinement 
& Fast global color/contrast correction with local detail enhancement \\
& ReasonX~\cite{dirik2025reasonx} 
& MLLM perceptual reward for intrinsic image decomposition 
& 9--25\% WHDR reduction via language-guided supervision \\

\bottomrule
\label{tab:restoration}
\end{tabularx}
\end{adjustwidth*}
\end{table*}
\subsection{Image Restoration and Super-Resolution}
\label{sec:restoration}
Reinforcement-based alignment has also been extended to image restoration and enhancement tasks, where models must recover high-quality content from degraded inputs under uncertain corruption conditions.
As shown in Table~\ref{tab:restoration}, 
\textbf{LRPO}~\cite{bin2025lrpo} is the first work introducing online RL to \textbf{blind face restoration}, addressing unknown degradation types and severity through a \textbf{composite reward function} that jointly evaluates facial identity preservation, perceptual quality, and degradation removal effectiveness, while incorporating \textbf{likelihood regularization} to prevent restored outputs from deviating from natural image distributions. \textbf{Pref-Restore}~\cite{yao2026bridging} adopts a \textbf{hierarchical restoration framework}, separating coarse structural recovery and fine texture refinement, and applies stage-specific rewards with \textbf{RL preference constraints} embedded in the diffusion loop to ensure human-aligned restoration trajectories. \textbf{FinPercep-RM}~\cite{liu2025finpercep} introduces \textbf{spatially-aware degradation maps} for super-resolution, allocating region-dependent degradation weights so RL can prioritize heavily corrupted areas, combined with \textbf{co-evolutionary curriculum learning} where reward and generation models progressively adapt to increasingly challenging degradations. \textbf{Diffusion Restoration via RL}~\cite{xu2025enhancing} leverages \textbf{MLLM-based IQA (Image Quality Assessment)} as reward signals, converting natural language quality evaluations into scalar rewards, and employs a \textbf{difficulty-adaptive RL/SFT hybrid}—using SFT for simpler degradations and RL for harder cases. For enhancement efficiency, \textbf{FlowLUT}~\cite{hu2025flowlut} integrates \textbf{look-up tables (LUT)} for fast global color and contrast adjustment with flow matching for complex local refinement. Finally, \textbf{ReasonX}~\cite{dirik2025reasonx} applies MLLM perceptual judgment as GRPO rewards to \textbf{intrinsic image decomposition}, achieving a \textbf{9--25\%} reduction in WHDR (Weighted Human Disagreement Rate). Collectively, these works demonstrate that composite rewards, hierarchical optimization, spatially adaptive weighting, curriculum-based co-evolution, language-guided quality assessment, and hybrid enhancement architectures are key to extending GRPO-style alignment to restoration and enhancement domains.

%% file: section/4_future.tex
\section{Future Outlook}
Looking forward, Flow-GRPO research is poised to evolve along several interconnected directions. A central challenge lies in developing a \textbf{unified theoretical framework} that coherently integrates dense reward design, efficient sampling, diversity preservation, and stable training dynamics, while providing convergence guarantees in continuous generative models and clarifying the SDE–ODE training–inference gap. As current empirical validation remains limited at extreme scales, systematic studies on \textbf{>$10$B-parameter models} are necessary to assess stability, scalability, and sample efficiency under large-scale training regimes. In video and long-horizon generation, improved \textbf{temporal reward modeling and credit assignment} mechanisms are required to ensure consistent behavior over extended sequences. Multi-objective optimization, as explored by APEX and MapReduce LoRA, raises open questions regarding \textbf{Pareto-optimal alignment} and principled trade-offs across competing reward dimensions. Another promising direction is \textbf{inference-time alignment}, where adaptive mechanisms can flexibly accommodate diverse user preferences without retraining. Beyond vision, Flow-GRPO’s expansion into scientific modeling—including PDE solvers, molecular force fields, and crystal structure generation—suggests a broader applicability to physically grounded domains. Meanwhile, \textbf{reasoning-augmented generation} and unified understanding–generation architectures point toward deeper cognitive integration, where intermediate reasoning and bidirectional reward signals shape generation quality. Finally, extending Flow-GRPO to \textbf{embodied intelligence} highlights the potential for continuous control, simulation-to-real transfer, and policy alignment in high-dimensional action spaces. Collectively, these directions indicate that Flow-GRPO is evolving from a task-specific alignment technique into a general reinforcement-driven framework for scalable, controllable, and multimodal generative intelligence.

%% file: reference.bib
@article{he2025tempflow,
  title={Tempflow-grpo: When timing matters for grpo in flow models},
  author={He, Xiaoxuan and Fu, Siming and Zhao, Yuke and Li, Wanli and Yang, Jian and Yin, Dacheng and Rao, Fengyun and Zhang, Bo},
  journal={arXiv preprint arXiv:2508.04324},
  year={2025}
}

@article{liu2025flow,
  title={Flow-grpo: Training flow matching models via online rl},
  author={Liu, Jie and Liu, Gongye and Liang, Jiajun and Li, Yangguang and Liu, Jiaheng and Wang, Xintao and Wan, Pengfei and Zhang, Di and Ouyang, Wanli},
  journal={arXiv preprint arXiv:2505.05470},
  year={2025}
}

@article{deng2026densegrpo,
  title={DenseGRPO: From Sparse to Dense Reward for Flow Matching Model Alignment},
  author={Deng, Haoyou and Yan, Keyu and Mao, Chaojie and Wang, Xiang and Liu, Yu and Gao, Changxin and Sang, Nong},
  journal={arXiv preprint arXiv:2601.20218},
  year={2026}
}

@article{fu2025dynamic,
  title={Dynamic-treerpo: Breaking the independent trajectory bottleneck with structured sampling},
  author={Fu, Xiaolong and Ma, Lichen and Guo, Zipeng and Zhou, Gaojing and Wang, Chongxiao and Dong, ShiPing and Zhou, Shizhe and Liu, Ximan and Fu, Jingling and Sin, Tan Lit and others},
  journal={arXiv preprint arXiv:2509.23352},
  year={2025}
}

@article{li2025mixgrpo,
  title={Mixgrpo: Unlocking flow-based grpo efficiency with mixed ode-sde},
  author={Li, Junzhe and Cui, Yutao and Huang, Tao and Ma, Yinping and Fan, Chun and Yang, Miles and Zhong, Zhao},
  journal={arXiv preprint arXiv:2507.21802},
  year={2025}
}

@article{lyu2025multi,
  title={Multi-GRPO: Multi-Group Advantage Estimation for Text-to-Image Generation with Tree-Based Trajectories and Multiple Rewards},
  author={Lyu, Qiang and Chen, Zicong and Wang, Chongxiao and Shi, Haolin and Gao, Shibo and Piao, Ran and Zeng, Youwei and Si, Jianlou and Ding, Fei and Li, Jing and others},
  journal={arXiv preprint arXiv:2512.00743},
  year={2025}
}

@article{ge2025expand,
  title={Expand and Prune: Maximizing Trajectory Diversity for Effective GRPO in Generative Models},
  author={Ge, Shiran and Huang, Chenyi and Ai, Yuang and Fan, Qihang and Huang, Huaibo and He, Ran},
  journal={arXiv preprint arXiv:2512.15347},
  year={2025}
}

@article{wang2025pref,
  title={Pref-grpo: Pairwise preference reward-based grpo for stable text-to-image reinforcement learning},
  author={Wang, Yibin and Li, Zhimin and Zang, Yuhang and Zhou, Yujie and Bu, Jiazi and Wang, Chunyu and Lu, Qinglin and Jin, Cheng and Wang, Jiaqi},
  journal={arXiv preprint arXiv:2508.20751},
  year={2025}
}

@article{wang2025grpo,
  title={Grpo-guard: Mitigating implicit over-optimization in flow matching via regulated clipping},
  author={Wang, Jing and Liang, Jiajun and Liu, Jie and Liu, Henglin and Liu, Gongye and Zheng, Jun and Pang, Wanyuan and Ma, Ao and Xie, Zhenyu and Wang, Xintao and others},
  journal={arXiv preprint arXiv:2510.22319},
  year={2025}
}

@article{zheng2025diffusionnft,
  title={Diffusionnft: Online diffusion reinforcement with forward process},
  author={Zheng, Kaiwen and Chen, Huayu and Ye, Haotian and Wang, Haoxiang and Zhang, Qinsheng and Jiang, Kai and Su, Hang and Ermon, Stefano and Zhu, Jun and Liu, Ming-Yu},
  journal={arXiv preprint arXiv:2509.16117},
  year={2025}
}

@article{wang2026gdro,
  title={GDRO: Group-level Reward Post-training Suitable for Diffusion Models},
  author={Wang, Yiyang and Chen, Xi and Xu, Xiaogang and Liu, Yu and Zhao, Hengshuang},
  journal={arXiv preprint arXiv:2601.02036},
  year={2026}
}

@article{li2025branchgrpo,
  title={Branchgrpo: Stable and efficient grpo with structured branching in diffusion models},
  author={Li, Yuming and Wang, Yikai and Zhu, Yuying and Zhao, Zhongyu and Lu, Ming and She, Qi and Zhang, Shanghang},
  journal={arXiv preprint arXiv:2509.06040},
  year={2025}
}

@article{ding2025treegrpo,
  title={TreeGRPO: Tree-Advantage GRPO for Online RL Post-Training of Diffusion Models},
  author={Ding, Zheng and Ye, Weirui},
  journal={arXiv preprint arXiv:2512.08153},
  year={2025}
}

@article{zhou2025co,
  title={Co-GRPO: Co-Optimized Group Relative Policy Optimization for Masked Diffusion Model},
  author={Zhou, Renping and Ni, Zanlin and Chen, Tianyi and Liu, Zeyu and Yue, Yang and Wang, Yulin and Wang, Yuxuan and Liu, Jingshu and Huang, Gao},
  journal={arXiv preprint arXiv:2512.22288},
  year={2025}
}

@article{yue2026know,
  title={Know Your Step: Faster and Better Alignment for Flow Matching Models via Step-aware Advantages},
  author={Yue, Zhixiong and Ni, Zixuan and Ye, Feiyang and Zhang, Jinshan and Shen, Sheng and Mi, Zhenpeng},
  journal={arXiv preprint arXiv:2602.01591},
  year={2026}
}

@article{meng2025identity,
  title={Identity-grpo: Optimizing multi-human identity-preserving video generation via reinforcement learning},
  author={Meng, Xiangyu and Zhang, Zixian and Zhang, Zhenghao and Liao, Junchao and Qin, Long and Wang, Weizhi},
  journal={arXiv preprint arXiv:2510.14256},
  year={2025}
}

@article{li2025growing,
  title={Growing with the Generator: Self-paced GRPO for Video Generation},
  author={Li, Rui and Liang, Yuanzhi and Ni, Ziqi and Huang, Haibing and Zhang, Chi and Li, Xuelong},
  journal={arXiv preprint arXiv:2511.19356},
  year={2025}
}

@article{chen2025superflow,
  title={SuperFlow: Training Flow Matching Models with RL on the Fly},
  author={Chen, Kaijie and Xu, Zhiyang and Shen, Ying and Lin, Zihao and Yao, Yuguang and Huang, Lifu},
  journal={arXiv preprint arXiv:2512.17951},
  year={2025}
}

@article{shaovalue,
  title={Value-Anchored Group Policy Optimization for Flow Models},
  author={Shao, Yawen and Xiao, Jie and Zhu, Kai and Liu, Yu and Zhai, Wei and Cao, Yang and Zha, Zheng-Jun}
}

@article{zhong2026euphonium,
  title={Euphonium: Steering Video Flow Matching via Process Reward Gradient Guided Stochastic Dynamics},
  author={Zhong, Ruizhe and Lian, Jiesong and Mi, Xiaoyue and Zhou, Zixiang and Zhou, Yuan and Lu, Qinglin and Yan, Junchi},
  journal={arXiv preprint arXiv:2602.04928},
  year={2026}
}

@article{kirstain2023pick,
  title={Pick-a-pic: An open dataset of user preferences for text-to-image generation},
  author={Kirstain, Yuval and Polyak, Adam and Singer, Uriel and Matiana, Shahbuland and Penna, Joe and Levy, Omer},
  journal={Advances in neural information processing systems},
  volume={36},
  pages={36652--36663},
  year={2023}
}

@article{wu2023human,
  title={Human preference score v2: A solid benchmark for evaluating human preferences of text-to-image synthesis},
  author={Wu, Xiaoshi and Hao, Yiming and Sun, Keqiang and Chen, Yixiong and Zhu, Feng and Zhao, Rui and Li, Hongsheng},
  journal={arXiv preprint arXiv:2306.09341},
  year={2023}
}

@article{ghosh2023geneval,
  title={Geneval: An object-focused framework for evaluating text-to-image alignment},
  author={Ghosh, Dhruba and Hajishirzi, Hannaneh and Schmidt, Ludwig},
  journal={Advances in Neural Information Processing Systems},
  volume={36},
  pages={52132--52152},
  year={2023}
}

@article{xue2025dancegrpo,
  title={DanceGRPO: Unleashing GRPO on Visual Generation},
  author={Xue, Zeyue and Wu, Jie and Gao, Yu and Kong, Fangyuan and Zhu, Lingting and Chen, Mengzhao and Liu, Zhiheng and Liu, Wei and Guo, Qiushan and Huang, Weilin and others},
  journal={arXiv preprint arXiv:2505.07818},
  year={2025}
}

@article{zhou2025fine,
  title={Fine-Grained GRPO for Precise Preference Alignment in Flow Models},
  author={Zhou, Yujie and Ling, Pengyang and Bu, Jiazi and Wang, Yibin and Zang, Yuhang and Wang, Jiaqi and Niu, Li and Zhai, Guangtao},
  journal={arXiv preprint arXiv:2510.01982},
  year={2025}
}

@article{he2025neighbor,
  title={Neighbor GRPO: Contrastive ODE Policy Optimization Aligns Flow Models},
  author={He, Dailan and Feng, Guanlin and Ge, Xingtong and Niu, Yazhe and Zhang, Yi and Ma, Bingqi and Song, Guanglu and Liu, Yu and Li, Hongsheng},
  journal={arXiv preprint arXiv:2511.16955},
  year={2025}
}

@article{zhang2026grpo,
  title={E-GRPO: High Entropy Steps Drive Effective Reinforcement Learning for Flow Models},
  author={Zhang, Shengjun and Zhang, Zhang and Dai, Chensheng and Duan, Yueqi},
  journal={arXiv preprint arXiv:2601.00423},
  year={2026}
}

@article{yu2025smart,
  title={Smart-grpo: Smartly sampling noise for efficient rl of flow-matching models},
  author={Yu, Benjamin and Liu, Jackie and Cui, Justin},
  journal={arXiv preprint arXiv:2510.02654},
  year={2025}
}

@article{liu2025diversegrpo,
  title={DiverseGRPO: Mitigating Mode Collapse in Image Generation via Diversity-Aware GRPO},
  author={Liu, Henglin and Huang, Huijuan and Wang, Jing and Liu, Chang and Li, Xiu and Ji, Xiangyang},
  journal={arXiv preprint arXiv:2512.21514},
  year={2025}
}

@article{wang2026flowse,
  title={FlowSE-GRPO: Training Flow Matching Speech Enhancement via Online Reinforcement Learning},
  author={Wang, Haoxu and Tian, Biao and Jiang, Yiheng and Pan, Zexu and Zhao, Shengkui and Ma, Bin and Chen, Daren and Li, Xiangang},
  journal={arXiv preprint arXiv:2601.16483},
  year={2026}
}

@article{luo2025sample,
  title={Sample By Step, Optimize By Chunk: Chunk-Level GRPO For Text-to-Image Generation},
  author={Luo, Yifu and Du, Penghui and Li, Bo and Du, Sinan and Zhang, Tiantian and Chang, Yongzhe and Wu, Kai and Gai, Kun and Wang, Xueqian},
  journal={arXiv preprint arXiv:2510.21583},
  year={2025}
}

@article{lee2025pcpo,
  title={PCPO: Proportionate Credit Policy Optimization for Aligning Image Generation Models},
  author={Lee, Jeongjae and Ye, Jong Chul},
  journal={arXiv preprint arXiv:2509.25774},
  year={2025}
}

@article{lu2025dpm,
  title={Dpm-solver++: Fast solver for guided sampling of diffusion probabilistic models},
  author={Lu, Cheng and Zhou, Yuhao and Bao, Fan and Chen, Jianfei and Li, Chongxuan and Zhu, Jun},
  journal={Machine Intelligence Research},
  pages={1--22},
  year={2025},
  publisher={Springer}
}

@article{xue2025advantage,
  title={Advantage weighted matching: Aligning rl with pretraining in diffusion models},
  author={Xue, Shuchen and Ge, Chongjian and Zhang, Shilong and Li, Yichen and Ma, Zhi-Ming},
  journal={arXiv preprint arXiv:2509.25050},
  year={2025}
}

@article{black2023training,
  title={Training diffusion models with reinforcement learning},
  author={Black, Kevin and Janner, Michael and Du, Yilun and Kostrikov, Ilya and Levine, Sergey},
  journal={arXiv preprint arXiv:2305.13301},
  year={2023}
}

@article{lipman2022flow,
  title={Flow matching for generative modeling},
  author={Lipman, Yaron and Chen, Ricky TQ and Ben-Hamu, Heli and Nickel, Maximilian and Le, Matt},
  journal={arXiv preprint arXiv:2210.02747},
  year={2022}
}

@article{luo2025reinforcing,
  title={Reinforcing Diffusion Models by Direct Group Preference Optimization},
  author={Luo, Yihong and Hu, Tianyang and Tang, Jing},
  journal={arXiv preprint arXiv:2510.08425},
  year={2025}
}

@article{choi2026rethinking,
  title={Rethinking the Design Space of Reinforcement Learning for Diffusion Models: On the Importance of Likelihood Estimation Beyond Loss Design},
  author={Choi, Jaemoo and Zhu, Yuchen and Guo, Wei and Molodyk, Petr and Yuan, Bo and Bai, Jinbin and Xin, Yi and Tao, Molei and Chen, Yongxin},
  journal={arXiv preprint arXiv:2602.04663},
  year={2026}
}

@article{sutton1999policy,
  title={Policy gradient methods for reinforcement learning with function approximation},
  author={Sutton, Richard S and McAllester, David and Singh, Satinder and Mansour, Yishay},
  journal={Advances in neural information processing systems},
  volume={12},
  year={1999}
}

@article{shao2024deepseekmath,
  title={Deepseekmath: Pushing the limits of mathematical reasoning in open language models},
  author={Shao, Zhihong and Wang, Peiyi and Zhu, Qihao and Xu, Runxin and Song, Junxiao and Bi, Xiao and Zhang, Haowei and Zhang, Mingchuan and Li, YK and Wu, Yang and others},
  journal={arXiv preprint arXiv:2402.03300},
  year={2024}
}

@article{schulman2017proximal,
  title={Proximal policy optimization algorithms},
  author={Schulman, John and Wolski, Filip and Dhariwal, Prafulla and Radford, Alec and Klimov, Oleg},
  journal={arXiv preprint arXiv:1707.06347},
  year={2017}
}

@article{wu2025oscar,
  title={OSCAR: Orthogonal Stochastic Control for Alignment-Respecting Diversity in Flow Matching},
  author={Wu, Jingxuan and Wan, Zhenglin and Yu, Xingrui and Yang, Yuzhe and An, Bo and Tsang, Ivor},
  journal={arXiv preprint arXiv:2510.09060},
  year={2025}
}

@article{chen2025taming,
  title={Taming Preference Mode Collapse via Directional Decoupling Alignment in Diffusion Reinforcement Learning},
  author={Chen, Chubin and Hu, Sujie and Zhu, Jiashu and Wu, Meiqi and Chen, Jintao and Li, Yanxun and Huang, Nisha and Fang, Chengyu and Wu, Jiahong and Chu, Xiangxiang and others},
  journal={arXiv preprint arXiv:2512.24146},
  year={2025}
}

@article{liu2026beyond,
  title={Beyond the Dirac Delta: Mitigating Diversity Collapse in Reinforcement Fine-Tuning for Versatile Image Generation},
  author={Liu, Jinmei and Li, Haoru and Sun, Zhenhong and Chen, Chaofeng and Bian, Yatao and Wang, Bo and Dong, Daoyi and Chen, Chunlin and Wang, Zhi},
  journal={arXiv preprint arXiv:2601.12401},
  year={2026}
}

@misc{borse2026resolvingidentitycrisistexttoimage,
      title={Resolving the Identity Crisis in Text-to-Image Generation}, 
      author={Shubhankar Borse and Farzad Farhadzadeh and Munawar Hayat and Fatih Porikli},
      year={2026},
      eprint={2510.01399},
      archivePrefix={arXiv},
      primaryClass={cs.CV},
      url={https://arxiv.org/abs/2510.01399}, 
}

@article{he2025gardo,
  title={GARDO: Reinforcing Diffusion Models without Reward Hacking},
  author={He, Haoran and Ye, Yuxiao and Liu, Jie and Liang, Jiajun and Wang, Zhiyong and Yuan, Ziyang and Wang, Xintao and Mao, Hangyu and Wan, Pengfei and Pan, Ling},
  journal={arXiv preprint arXiv:2512.24138},
  year={2025}
}

@article{ye2025data,
  title={Data-regularized Reinforcement Learning for Diffusion Models at Scale},
  author={Ye, Haotian and Zheng, Kaiwen and Xu, Jiashu and Li, Puheng and Chen, Huayu and Han, Jiaqi and Liu, Sheng and Zhang, Qinsheng and Mao, Hanzi and Hao, Zekun and others},
  journal={arXiv preprint arXiv:2512.04332},
  year={2025}
}

@article{tan2026consistentrft,
  title={ConsistentRFT: Reducing Visual Hallucinations in Flow-based Reinforcement Fine-Tuning},
  author={Tan, Xiaofeng and Liu, Jun and Fan, Yuanting and Gao, Bin-Bin and Jiang, Xi and Chen, Xiaochen and Peng, Jinlong and Wang, Chengjie and Wang, Hongsong and Zheng, Feng},
  journal={arXiv preprint arXiv:2602.03425},
  year={2026}
}

@article{hong2026understanding,
  title={Understanding Reward Hacking in Text-to-Image Reinforcement Learning},
  author={Hong, Yunqi and Kao, Kuei-Chun and Zhou, Hengguang and Hsieh, Cho-Jui},
  journal={arXiv preprint arXiv:2601.03468},
  year={2026}
}

@article{wang2025coefficients,
  title={Coefficients-Preserving Sampling for Reinforcement Learning with Flow Matching},
  author={Wang, Feng and Yu, Zihao},
  journal={arXiv preprint arXiv:2509.05952},
  year={2025}
}

@article{song2020denoising,
  title={Denoising diffusion implicit models},
  author={Song, Jiaming and Meng, Chenlin and Ermon, Stefano},
  journal={arXiv preprint arXiv:2010.02502},
  year={2020}
}

@article{lian2025solireward,
  title={SoliReward: Mitigating Susceptibility to Reward Hacking and Annotation Noise in Video Generation Reward Models},
  author={Lian, Jiesong and Zhong, Ruizhe and Zhou, Zixiang and Mi, Xiaoyue and Hao, Yixue and Zhou, Yuan and Lu, Qinglin and Hu, Long and Yan, Junchi},
  journal={arXiv preprint arXiv:2512.22170},
  year={2025}
}

@article{holderrieth2025glass,
  title={GLASS Flows: Transition Sampling for Alignment of Flow and Diffusion Models},
  author={Holderrieth, Peter and Singer, Uriel and Jaakkola, Tommi and Chen, Ricky TQ and Lipman, Yaron and Karrer, Brian},
  journal={arXiv preprint arXiv:2509.25170},
  year={2025}
}

@article{singhal2025general,
  title={A general framework for inference-time scaling and steering of diffusion models},
  author={Singhal, Raghav and Horvitz, Zachary and Teehan, Ryan and Ren, Mengye and Yu, Zhou and McKeown, Kathleen and Ranganath, Rajesh},
  journal={arXiv preprint arXiv:2501.06848},
  year={2025}
}

@article{sheng2025understanding,
  title={Understanding Sampler Stochasticity in Training Diffusion Models for RLHF},
  author={Sheng, Jiayuan and Zhao, Hanyang and Chen, Haoxian and Yao, David D and Tang, Wenpin},
  journal={arXiv preprint arXiv:2510.10767},
  year={2025}
}

@article{holderrieth2026diamond,
  title={Diamond Maps: Efficient Reward Alignment via Stochastic Flow Maps},
  author={Holderrieth, Peter and Chen, Douglas and Eyring, Luca and Shah, Ishin and Anantharaman, Giri and He, Yutong and Akata, Zeynep and Jaakkola, Tommi and Boffi, Nicholas Matthew and Simchowitz, Max},
  journal={arXiv preprint arXiv:2602.05993},
  year={2026}
}

@article{zhang2022gddim,
  title={gddim: Generalized denoising diffusion implicit models},
  author={Zhang, Qinsheng and Tao, Molei and Chen, Yongxin},
  journal={arXiv preprint arXiv:2206.05564},
  year={2022}
}

@article{wang2026unified,
  title={Unified Personalized Reward Model for Vision Generation},
  author={Wang, Yibin and Zang, Yuhang and Han, Feng and Bu, Jiazi and Zhou, Yujie and Jin, Cheng and Wang, Jiaqi},
  journal={arXiv preprint arXiv:2602.02380},
  year={2026}
}

@article{ashutosh2026human,
  title={Human detectors are surprisingly powerful reward models},
  author={Ashutosh, Kumar and Wang, XuDong and Yin, Xi and Grauman, Kristen and Polyak, Adam and Misra, Ishan and Girdhar, Rohit},
  journal={arXiv preprint arXiv:2601.14037},
  year={2026}
}

@article{wang2025vr,
  title={Vr-thinker: Boosting video reward models through thinking-with-image reasoning},
  author={Wang, Qunzhong and Liu, Jie and Liang, Jiajun and Jiang, Yilei and Zhang, Yuanxing and Chen, Jinyuan and Zheng, Yaozhi and Wang, Xintao and Wan, Pengfei and Yue, Xiangyu and others},
  journal={arXiv preprint arXiv:2510.10518},
  year={2025}
}

@article{wu2025rewarddance,
  title={Rewarddance: Reward scaling in visual generation},
  author={Wu, Jie and Gao, Yu and Ye, Zilyu and Li, Ming and Li, Liang and Guo, Hanzhong and Liu, Jie and Xue, Zeyue and Hou, Xiaoxia and Liu, Wei and others},
  journal={arXiv preprint arXiv:2509.08826},
  year={2025}
}

@article{wang2026thinking,
  title={Thinking with Frames: Generative Video Distortion Evaluation via Frame Reward Model},
  author={Wang, Yuan and Liao, Borui and Huang, Huijuan and Lu, Jinda and Li, Ouxiang and Liu, Kuien and Wang, Meng and Wang, Xiang},
  journal={arXiv preprint arXiv:2601.04033},
  year={2026}
}

@article{guo2025imagedoctor,
  title={ImageDoctor: Diagnosing Text-to-Image Generation via Grounded Image Reasoning},
  author={Guo, Yuxiang and Liu, Jiang and Wang, Ze and Chen, Hao and Sun, Ximeng and Zhao, Yang and Wu, Jialian and Yu, Xiaodong and Liu, Zicheng and Barsoum, Emad},
  journal={arXiv preprint arXiv:2510.01010},
  year={2025}
}

@article{luo2025editscore,
  title={Editscore: Unlocking online rl for image editing via high-fidelity reward modeling},
  author={Luo, Xin and Wang, Jiahao and Wu, Chenyuan and Xiao, Shitao and Jiang, Xiyan and Lian, Defu and Zhang, Jiajun and Liu, Dong and others},
  journal={arXiv preprint arXiv:2509.23909},
  year={2025}
}

@article{ye2025realgen,
  title={Realgen: Photorealistic text-to-image generation via detector-guided rewards},
  author={Ye, Junyan and Zhu, Leiqi and Guo, Yuncheng and Jiang, Dongzhi and Huang, Zilong and Zhang, Yifan and Yan, Zhiyuan and Fu, Haohuan and He, Conghui and Li, Weijia},
  journal={arXiv preprint arXiv:2512.00473},
  year={2025}
}

@article{gong2025onereward,
  title={Onereward: Unified mask-guided image generation via multi-task human preference learning},
  author={Gong, Yuan and Wang, Xionghui and Wu, Jie and Wang, Shiyin and Wang, Yitong and Wu, Xinglong},
  journal={arXiv preprint arXiv:2508.21066},
  year={2025}
}

@article{wang2026everything,
  title={Everything in Its Place: Benchmarking Spatial Intelligence of Text-to-Image Models},
  author={Wang, Zengbin and Hu, Xuecai and Wang, Yong and Xiong, Feng and Zhang, Man and Chu, Xiangxiang},
  journal={arXiv preprint arXiv:2601.20354},
  year={2026}
}

@article{wang2025unigenbench++,
  title={Unigenbench++: A unified semantic evaluation benchmark for text-to-image generation},
  author={Wang, Yibin and Li, Zhimin and Zang, Yuhang and Bu, Jiazi and Zhou, Yujie and Xin, Yi and He, Junjun and Wang, Chunyu and Lu, Qinglin and Jin, Cheng and others},
  journal={arXiv preprint arXiv:2510.18701},
  year={2025}
}

@article{cao2025t2av,
  title={T2AV-Compass: Towards Unified Evaluation for Text-to-Audio-Video Generation},
  author={Cao, Zhe and Wang, Tao and Wang, Jiaming and Wang, Yanghai and Zhang, Yuanxing and Chen, Jialu and Deng, Miao and Wang, Jiahao and Guo, Yubin and Liao, Chenxi and others},
  journal={arXiv preprint arXiv:2512.21094},
  year={2025}
}

@article{wang2025magicmirror,
  title={MagicMirror: A Large-Scale Dataset and Benchmark for Fine-Grained Artifacts Assessment in Text-to-Image Generation},
  author={Wang, Jia and Hu, Jie and Ma, Xiaoqi and Ma, Hanghang and Zeng, Yanbing and Wei, Xiaoming},
  journal={arXiv preprint arXiv:2509.10260},
  year={2025}
}

@article{jin2025realbench,
  title={Realbench: Benchmarking verilog generation models with real-world ip designs},
  author={Jin, Pengwei and Huang, Di and Li, Chongxiao and Cheng, Shuyao and Zhao, Yang and Zheng, Xinyao and Zhu, Jiaguo and Xing, Shuyi and Dou, Bohan and Zhang, Rui and others},
  journal={arXiv preprint arXiv:2507.16200},
  year={2025}
}

@article{wang2026promptrl,
  title={PromptRL: Prompt Matters in RL for Flow-Based Image Generation},
  author={Wang, Fu-Yun and Zhang, Han and Gharbi, Michael and Li, Hongsheng and Park, Taesung},
  journal={arXiv preprint arXiv:2602.01382},
  year={2026}
}

@article{kou2026think,
  title={Think-Then-Generate: Reasoning-Aware Text-to-Image Diffusion with LLM Encoders},
  author={Kou, Siqi and Jin, Jiachun and Zhou, Zetong and Ma, Ye and Wang, Yugang and Chen, Quan and Jiang, Peng and Yang, Xiao and Zhu, Jun and Yu, Kai and others},
  journal={arXiv preprint arXiv:2601.10332},
  year={2026}
}

@article{jiao2025thinkgen,
  title={ThinkGen: Generalized Thinking for Visual Generation},
  author={Jiao, Siyu and Lin, Yiheng and Zhong, Yujie and She, Qi and Zhou, Wei and Lan, Xiaohan and Huang, Zilong and Yu, Fei and Yu, Yingchen and Zhao, Yunqing and others},
  journal={arXiv preprint arXiv:2512.23568},
  year={2025}
}

@article{zhou2026unified,
  title={Unified Thinker: A General Reasoning Modular Core for Image Generation},
  author={Zhou, Sashuai and Zhou, Qiang and Hu, Jijin and Yang, Hanqing and Cao, Yue and Ma, Junpeng and Ma, Yinchao and Song, Jun and Ge, Tiezheng and Yu, Cheng and others},
  journal={arXiv preprint arXiv:2601.03127},
  year={2026}
}

@article{tong2026cof,
  title={CoF-T2I: Video Models as Pure Visual Reasoners for Text-to-Image Generation},
  author={Tong, Chengzhuo and Chang, Mingkun and Zhang, Shenglong and Wang, Yuran and Liang, Cheng and Zhao, Zhizheng and An, Ruichuan and Zeng, Bohan and Shi, Yang and Dai, Yifan and others},
  journal={arXiv preprint arXiv:2601.10061},
  year={2026}
}

@article{zarei2025agentcomp,
  title={AgentComp: From Agentic Reasoning to Compositional Mastery in Text-to-Image Models},
  author={Zarei, Arman and Pan, Jiacheng and Gwilliam, Matthew and Feizi, Soheil and Yang, Zhenheng},
  journal={arXiv preprint arXiv:2512.09081},
  year={2025}
}

@article{ouyang2026alignment,
  title={Alignment of Diffusion Model and Flow Matching for Text-to-Image Generation},
  author={Ouyang, Yidong and Xie, Liyan and Zha, Hongyuan and Cheng, Guang},
  journal={arXiv preprint arXiv:2602.00413},
  year={2026}
}

@article{chen2026apex,
  title={APEX: Learning Adaptive Priorities for Multi-Objective Alignment in Vision-Language Generation},
  author={Chen, Dongliang and Zhuang, Xinlin and Xu, Junjie and Xie, Luojian and Wang, Zehui and Zhuang, Jiaxi and Yang, Haolin and Dou, Liang and He, Xiao and Wu, Xingjiao and others},
  journal={arXiv preprint arXiv:2601.06574},
  year={2026}
}

@article{chen2025mapreduce,
  title={MapReduce LoRA: Advancing the Pareto Front in Multi-Preference Optimization for Generative Models},
  author={Chen, Chieh-Yun and Wang, Zhonghao and Chen, Qi and Ye, Zhifan and Shi, Min and Zhao, Yue and Zhao, Yinan and Qu, Hui and Lin, Wei-An and Shen, Yiru and others},
  journal={arXiv preprint arXiv:2511.20629},
  year={2025}
}

@article{song2025dctext,
  title={DCText: Scheduled Attention Masking for Visual Text Generation via Divide-and-Conquer Strategy},
  author={Song, Jaewoo and Choi, Jooyoung and Baek, Kanghyun and Lee, Sangyub and Park, Daemin and Yoon, Sungroh},
  journal={arXiv preprint arXiv:2512.01302},
  year={2025}
}

@article{wang2025ovis,
  title={Ovis-Image Technical Report},
  author={Wang, Guo-Hua and Cao, Liangfu and Cui, Tianyu and Fu, Minghao and Chen, Xiaohao and Zhan, Pengxin and Zhao, Jianshan and Li, Lan and Fu, Bowen and Liu, Jiaqi and others},
  journal={arXiv preprint arXiv:2511.22982},
  year={2025}
}

@article{wu2025qwen,
  title={Qwen-image technical report},
  author={Wu, Chenfei and Li, Jiahao and Zhou, Jingren and Lin, Junyang and Gao, Kaiyuan and Yan, Kun and Yin, Sheng-ming and Bai, Shuai and Xu, Xiao and Chen, Yilei and others},
  journal={arXiv preprint arXiv:2508.02324},
  year={2025}
}

@article{wang2025psr,
  title={PSR: Scaling Multi-Subject Personalized Image Generation with Pairwise Subject-Consistency Rewards},
  author={Wang, Shulei and Wei, Longhui and He, Xin and Ouyang, Jianbo and Lu, Hui and Zhao, Zhou and Tian, Qi},
  journal={arXiv preprint arXiv:2512.01236},
  year={2025}
}

@article{wu2025multicrafter,
  title={MultiCrafter: High-Fidelity Multi-Subject Generation via Spatially Disentangled Attention and Identity-Aware Reinforcement Learning},
  author={Wu, Tao and Jiang, Yibo and Lu, Yehao and Wang, Zhizhong and Huang, Zeyi and Qin, Zequn and Li, Xi},
  year={2025}
}

@article{huang2025competition,
  title={From Competition to Synergy: Unlocking Reinforcement Learning for Subject-Driven Image Generation},
  author={Huang, Ziwei and Shu, Ying and Fang, Hao and Long, Quanyu and Wang, Wenya and Guo, Qiushi and Ge, Tiezheng and Gan, Leilei},
  journal={arXiv preprint arXiv:2510.18263},
  year={2025}
}

@article{fang2025proxt2i,
  title={ProxT2I: Efficient Reward-Guided Text-to-Image Generation via Proximal Diffusion},
  author={Fang, Zhenghan and Zheng, Jian and Gao, Qiaozi and Gao, Xiaofeng and Sulam, Jeremias},
  journal={arXiv preprint arXiv:2511.18742},
  year={2025}
}

@article{yang2025hicogen,
  title={HiCoGen: Hierarchical Compositional Text-to-Image Generation in Diffusion Models via Reinforcement Learning},
  author={Yang, Hongji and Zhou, Yucheng and Han, Wencheng and Tao, Runzhou and Qiu, Zhongying and Yang, Jianfei and Shen, Jianbing},
  journal={arXiv preprint arXiv:2511.19965},
  year={2025}
}

@article{zhao2025demystifying,
  title={Demystifying Numerosity in Diffusion Models--Limitations and Remedies},
  author={Zhao, Yaqi and Wang, Xiaochen and Dong, Li and Zhang, Wentao and Yuan, Yuhui},
  journal={arXiv preprint arXiv:2510.11117},
  year={2025}
}

@article{chen2025flash,
  title={Flash-DMD: Towards High-Fidelity Few-Step Image Generation with Efficient Distillation and Joint Reinforcement Learning},
  author={Chen, Guanjie and Huang, Shirui and Liu, Kai and Zhu, Jianchen and Qu, Xiaoye and Chen, Peng and Cheng, Yu and Sun, Yifu},
  journal={arXiv preprint arXiv:2511.20549},
  year={2025}
}

@article{jiang2025distribution,
  title={Distribution Matching Distillation Meets Reinforcement Learning},
  author={Jiang, Dengyang and Liu, Dongyang and Wang, Zanyi and Wu, Qilong and Li, Liuzhuozheng and Li, Hengzhuang and Jin, Xin and Liu, David and Li, Zhen and Zhang, Bo and others},
  journal={arXiv preprint arXiv:2511.13649},
  year={2025}
}

@article{mao2025image,
  title={The Image as Its Own Reward: Reinforcement Learning with Adversarial Reward for Image Generation},
  author={Mao, Weijia and Chen, Hao and Yang, Zhenheng and Shou, Mike Zheng},
  journal={arXiv preprint arXiv:2511.20256},
  year={2025}
}

@article{xue2025mogan,
  title={MoGAN: Improving Motion Quality in Video Diffusion via Few-Step Motion Adversarial Post-Training},
  author={Xue, Haotian and Chen, Qi and Wang, Zhonghao and Huang, Xun and Shechtman, Eli and Xie, Jinrong and Chen, Yongxin},
  journal={arXiv preprint arXiv:2511.21592},
  year={2025}
}

@article{yu2025ttsnap,
  title={TTSnap: Test-Time Scaling of Diffusion Models via Noise-Aware Pruning},
  author={Yu, Qingtao and Song, Changlin and Sun, Minghao and Yu, Zhengyang and Verma, Vinay Kumar and Roy, Soumya and Negi, Sumit and Li, Hongdong and Campbell, Dylan},
  journal={arXiv preprint arXiv:2511.22242},
  year={2025}
}

@article{xu2025highly,
  title={Highly Efficient Test-Time Scaling for T2I Diffusion Models with Text Embedding Perturbation},
  author={Xu, Hang and Huang, Linjiang and Zhao, Feng},
  journal={arXiv preprint arXiv:2512.03996},
  year={2025}
}

@article{xie2026hyperalign,
  title={HyperAlign: Hypernetwork for Efficient Test-Time Alignment of Diffusion Models},
  author={Xie, Xin and Guo, Jiaxian and Gong, Dong},
  journal={arXiv preprint arXiv:2601.15968},
  year={2026}
}

@article{guidanceinference,
  title={INFERENCE-TIME ALIGNMENT CONTROL FOR DIFFU},
  author={GUIDANCE, REINFORCEMENT LEARNING}
}

@article{su2025navigating,
  title={Navigating the Exploration-Exploitation Tradeoff in Inference-Time Scaling of Diffusion Models},
  author={Su, Xun and Huang, Jianming and Yusen, Yang and Fang, Zhongxi and Kasai, Hiroyuki},
  journal={arXiv preprint arXiv:2508.12361},
  year={2025}
}

@article{mi2025milr,
  title={MILR: Improving Multimodal Image Generation via Test-Time Latent Reasoning},
  author={Mi, Yapeng and Li, Hengli and Zhao, Yanpeng and Li, Chenxi and Wu, Huimin and Ma, Xiaojian and Zhu, Song-Chun and Wu, Ying Nian and Li, Qing},
  journal={arXiv preprint arXiv:2509.22761},
  year={2025}
}

@article{feng2025rubricrl,
  title={RubricRL: Simple Generalizable Rewards for Text-to-Image Generation},
  author={Feng, Xuelu and Li, Yunsheng and Wan, Ziyu and Gao, Zixuan and Yuan, Junsong and Chen, Dongdong and Qiao, Chunming},
  journal={arXiv preprint arXiv:2511.20651},
  year={2025}
}

@article{jia2025emofeedback,
  title={EmoFeedback$^2$: Reinforcement of Continuous Emotional Image Generation via LVLM-based Reward and Textual Feedback},
  author={Jia, Jingyang and Shu, Kai and Yang, Gang and Xing, Long and Chen, Xun and Liu, Aiping},
  journal={arXiv preprint arXiv:2511.19982},
  year={2025}
}

@article{ping2025paco,
  title={PaCo-RL: Advancing Reinforcement Learning for Consistent Image Generation with Pairwise Reward Modeling},
  author={Ping, Bowen and Jia, Chengyou and Luo, Minnan and Xia, Changliang and Shen, Xin and Dang, Zhuohang and Qian, Hangwei},
  journal={arXiv preprint arXiv:2512.04784},
  year={2025}
}

@article{zhou2025bidedpo,
  title={BideDPO: Conditional Image Generation with Simultaneous Text and Condition Alignment},
  author={Zhou, Dewei and Li, Mingwei and Yang, Zongxin and Lu, Yu and Xu, Yunqiu and Wang, Zhizhong and Huang, Zeyi and Yang, Yi},
  journal={arXiv preprint arXiv:2511.19268},
  year={2025}
}

@article{jia2025emotion,
  title={Emotion-Director: Bridging Affective Shortcut in Emotion-Oriented Image Generation},
  author={Jia, Guoli and Hu, Junyao and Long, Xinwei and Tian, Kai and Zhang, Kaiyan and Zhao, KaiKai and Ding, Ning and Zhou, Bowen},
  journal={arXiv preprint arXiv:2512.19479},
  year={2025}
}

@article{yiflach2025data,
  title={Data-Driven Loss Functions for Inference-Time Optimization in Text-to-Image Generation},
  author={Yiflach, Sapir Esther and Atzmon, Yuval and Chechik, Gal},
  journal={arXiv preprint arXiv:2509.02295},
  year={2025}
}

@article{yu2025designing,
  title={Designing Instance-Level Sampling Schedules via REINFORCE with James-Stein Shrinkage},
  author={Yu, Peiyu and Kothawade, Suraj and Xie, Sirui and Wu, Ying Nian and Fei, Hongliang},
  journal={arXiv preprint arXiv:2511.22177},
  year={2025}
}

@article{wang2026tagrpo,
  title={TAGRPO: Boosting GRPO on Image-to-Video Generation with Direct Trajectory Alignment},
  author={Wang, Jin and Lu, Jianxiang and Xu, Guangzheng and Chen, Comi and Yang, Haoyu and Wang, Linqing and Chen, Peng and Chen, Mingtao and Hu, Zhichao and Wu, Longhuang and others},
  journal={arXiv preprint arXiv:2601.05729},
  year={2026}
}

@article{zhang2026physrvg,
  title={PhysRVG: Physics-Aware Unified Reinforcement Learning for Video Generative Models},
  author={Zhang, Qiyuan and Gong, Biao and Tan, Shuai and Zhang, Zheng and Shen, Yujun and Zhu, Xing and Li, Yuyuan and Yao, Kelu and Shen, Chunhua and Zou, Changqing},
  journal={arXiv preprint arXiv:2601.11087},
  year={2026}
}

@article{wang2026diffusion,
  title={Diffusion-DRF: Differentiable Reward Flow for Video Diffusion Fine-Tuning},
  author={Wang, Yifan and Li, Yanyu and Tulyakov, Sergey and Fu, Yun and Kag, Anil},
  journal={arXiv preprint arXiv:2601.04153},
  year={2026}
}

@article{xu2025beyond,
  title={Beyond Reward Margin: Rethinking and Resolving Likelihood Displacement in Diffusion Models via Video Generation},
  author={Xu, Ruojun and Kai, Yu and Ren, Xuhua and Cheng, Jiaxiang and Ma, Bing and Zheng, Tianxiang and Lu, Qinhlin},
  journal={arXiv preprint arXiv:2511.19049},
  year={2025}
}

@article{yang2025mcsc,
  title={McSc: Motion-Corrective Preference Alignment for Video Generation with Self-Critic Hierarchical Reasoning},
  author={Yang, Qiushi and Chen, Yingjie and Yao, Yuan and Men, Yifang and Liu, Huaizhuo and Cui, Miaomiao},
  journal={arXiv preprint arXiv:2511.22974},
  year={2025}
}

@article{shen2025identity,
  title={Identity-preserving image-to-video generation via reward-guided optimization},
  author={Shen, Liao and Jiang, Wentao and Zhu, Yiran and Li, Jiahe and Ge, Tiezheng and Cao, Zhiguo and Zheng, Bo},
  journal={arXiv preprint arXiv:2510.14255},
  year={2025}
}

@article{pan2025id,
  title={ID-Crafter: VLM-Grounded Online RL for Compositional Multi-Subject Video Generation},
  author={Pan, Panwang and Zhao, Jingjing and Lin, Yuchen and Lin, Chenguo and Li, Chenxin and Liu, Hengyu and Shen, Tingting and Mu, Yadong},
  journal={arXiv preprint arXiv:2511.00511},
  year={2025}
}

@article{guo2026dreamid,
  title={DreamID-V: Bridging the Image-to-Video Gap for High-Fidelity Face Swapping via Diffusion Transformer},
  author={Guo, Xu and Ye, Fulong and Li, Xinghui and Tu, Pengqi and Zhang, Pengze and Sun, Qichao and Zhao, Songtao and Hou, Xiangwang and He, Qian},
  journal={arXiv preprint arXiv:2601.01425},
  year={2026}
}

@article{zhao2025real,
  title={Real-Time Motion-Controllable Autoregressive Video Diffusion},
  author={Zhao, Kesen and Shi, Jiaxin and Zhu, Beier and Zhou, Junbao and Shen, Xiaolong and Zhou, Yuan and Sun, Qianru and Zhang, Hanwang},
  journal={arXiv preprint arXiv:2510.08131},
  year={2025}
}

@article{wang2025taming,
  title={Taming Camera-Controlled Video Generation with Verifiable Geometry Reward},
  author={Wang, Zhaoqing and Xia, Xiaobo and Bie, Zhuolin and Liu, Jinlin and Yu, Dongdong and Bian, Jia-Wang and Wang, Changhu},
  journal={arXiv preprint arXiv:2512.02870},
  year={2025}
}

@article{cong2025viva,
  title={VIVA: VLM-Guided Instruction-Based Video Editing with Reward Optimization},
  author={Cong, Xiaoyan and Yang, Haotian and Wang, Angtian and Wang, Yizhi and Yang, Yiding and Zhang, Canyu and Ma, Chongyang},
  journal={arXiv preprint arXiv:2512.16906},
  year={2025}
}

@article{liu2025revise,
  title={ReViSE: Towards Reason-Informed Video Editing in Unified Models with Self-Reflective Learning},
  author={Liu, Xinyu and Yuan, Hangjie and Wei, Yujie and Xing, Jiazheng and Han, Yujin and Pan, Jiahao and Ma, Yanbiao and Chan, Chi-Min and Zhao, Kang and Zhang, Shiwei and others},
  journal={arXiv preprint arXiv:2512.09924},
  year={2025}
}

@article{cheng2025video,
  title={Video-as-Answer: Predict and Generate Next Video Event with Joint-GRPO},
  author={Cheng, Junhao and Hou, Liang and Tao, Xin and Liao, Jing},
  journal={arXiv preprint arXiv:2511.16669},
  year={2025}
}

@article{li2025happens,
  title={What Happens Next? Next Scene Prediction with a Unified Video Model},
  author={Li, Xinjie and Chen, Zhimin and Zhao, Rui and Schiffers, Florian and Liao, Zhenyu and Bhat, Vimal},
  journal={arXiv preprint arXiv:2512.13015},
  year={2025}
}

@article{ye2025reinforcement,
  title={Reinforcement Learning with Inverse Rewards for World Model Post-training},
  author={Ye, Yang and He, Tianyu and Yang, Shuo and Bian, Jiang},
  journal={arXiv preprint arXiv:2509.23958},
  year={2025}
}

@article{team2026fsvideo,
  title={FSVideo: Fast Speed Video Diffusion Model in a Highly-Compressed Latent Space},
  author={Team, FSVideo and Chen, Qingyu and Fang, Zhiyuan and Huang, Haibin and Huang, Xinwei and Jin, Tong and Lin, Minxuan and Liu, Bo and Liu, Celong and Ma, Chongyang and others},
  journal={arXiv preprint arXiv:2602.02092},
  year={2026}
}

@article{team2025longcat,
  title={Longcat-video technical report},
  author={Team, Meituan LongCat and Cai, Xunliang and Huang, Qilong and Kang, Zhuoliang and Li, Hongyu and Liang, Shijun and Ma, Liya and Ren, Siyu and Wei, Xiaoming and Xie, Rixu and others},
  journal={arXiv preprint arXiv:2510.22200},
  year={2025}
}

@article{gao2025seedance,
  title={Seedance 1.0: Exploring the boundaries of video generation models},
  author={Gao, Yu and Guo, Haoyuan and Hoang, Tuyen and Huang, Weilin and Jiang, Lu and Kong, Fangyuan and Li, Huixia and Li, Jiashi and Li, Liang and Li, Xiaojie and others},
  journal={arXiv preprint arXiv:2506.09113},
  year={2025}
}

@article{seedance2025seedance,
  title={Seedance 1.5 pro: A Native Audio-Visual Joint Generation Foundation Model},
  author={Seedance, Team and Chen, Heyi and Chen, Siyan and Chen, Xin and Chen, Yanfei and Chen, Ying and Chen, Zhuo and Cheng, Feng and Cheng, Tianheng and Cheng, Xinqi and others},
  journal={arXiv preprint arXiv:2512.13507},
  year={2025}
}

@article{cui2025self,
  title={Self-forcing++: Towards minute-scale high-quality video generation},
  author={Cui, Justin and Wu, Jie and Li, Ming and Yang, Tao and Li, Xiaojie and Wang, Rui and Bai, Andrew and Ban, Yuanhao and Hsieh, Cho-Jui},
  journal={arXiv preprint arXiv:2510.02283},
  year={2025}
}

@article{liu2025video,
  title={Video Text Preservation with Synthetic Text-Rich Videos},
  author={Liu, Ziyang and Valencia, Kevin and Cui, Justin},
  journal={arXiv preprint arXiv:2511.05573},
  year={2025}
}

@article{lu2025reward,
  title={Reward forcing: Efficient streaming video generation with rewarded distribution matching distillation},
  author={Lu, Yunhong and Zeng, Yanhong and Li, Haobo and Ouyang, Hao and Wang, Qiuyu and Cheng, Ka Leong and Zhu, Jiapeng and Cao, Hengyuan and Zhang, Zhipeng and Zhu, Xing and others},
  journal={arXiv preprint arXiv:2512.04678},
  year={2025}
}

@article{zhang2025vq,
  title={Vq-insight: Teaching vlms for ai-generated video quality understanding via progressive visual reinforcement learning},
  author={Zhang, Xuanyu and Li, Weiqi and Zhao, Shijie and Li, Junlin and Zhang, Li and Zhang, Jian},
  journal={arXiv preprint arXiv:2506.18564},
  year={2025}
}

@article{li2026thinkrl,
  title={ThinkRL-Edit: Thinking in Reinforcement Learning for Reasoning-Centric Image Editing},
  author={Li, Hengjia and Jiang, Liming and Yan, Qing and Song, Yizhi and Kang, Hao and Liu, Zichuan and Lu, Xin and Wu, Boxi and Cai, Deng},
  journal={arXiv preprint arXiv:2601.03467},
  year={2026}
}

@article{he2026re,
  title={Re-Align: Structured Reasoning-guided Alignment for In-Context Image Generation and Editing},
  author={He, Runze and Cheng, Yiji and Hang, Tiankai and Li, Zhimin and Xu, Yu and Yin, Zijin and Zhang, Shiyi and Dai, Wenxun and Du, Penghui and Ma, Ao and others},
  journal={arXiv preprint arXiv:2601.05124},
  year={2026}
}

@article{li2025cogniedit,
  title={CogniEdit: Dense Gradient Flow Optimization for Fine-Grained Image Editing},
  author={Li, Yan and Liu, Lin and Zhang, Xiaopeng and Xue, Wei and Luo, Wenhan and Guo, Yike and Tian, Qi},
  journal={arXiv preprint arXiv:2512.13276},
  year={2025}
}

@article{lin2025jarvisevo,
  title={JarvisEvo: Towards a Self-Evolving Photo Editing Agent with Synergistic Editor-Evaluator Optimization},
  author={Lin, Yunlong and Wang, Linqing and Lin, Kunjie and Lin, Zixu and Gong, Kaixiong and Li, Wenbo and Lin, Bin and Li, Zhenxi and Zhang, Shiyi and Peng, Yuyang and others},
  journal={arXiv preprint arXiv:2511.23002},
  year={2025}
}

@article{wan2025motionedit,
  title={MotionEdit: Benchmarking and Learning Motion-Centric Image Editing},
  author={Wan, Yixin and Ke, Lei and Yu, Wenhao and Chang, Kai-Wei and Yu, Dong},
  journal={arXiv preprint arXiv:2512.10284},
  year={2025}
}

@article{guo2025repainter,
  title={RePainter: Empowering E-commerce Object Removal via Spatial-matting Reinforcement Learning},
  author={Guo, Zipeng and Ma, Lichen and Fu, Xiaolong and Zhou, Gaojing and Yang, Lan and Zhou, Yuchen and Liu, Linkai and He, Yu and Liu, Ximan and Dong, Shiping and others},
  journal={arXiv preprint arXiv:2510.07721},
  year={2025}
}

@article{liu2025omnirefiner,
  title={OmniRefiner: Reinforcement-Guided Local Diffusion Refinement},
  author={Liu, Yaoli and Ouyang, Ziheng and Lou, Shengtao and Song, Yiren},
  journal={arXiv preprint arXiv:2511.19990},
  year={2025}
}

@article{li2025uniworld,
  title={Uniworld-v2: Reinforce image editing with diffusion negative-aware finetuning and mllm implicit feedback},
  author={Li, Zongjian and Liu, Zheyuan and Zhang, Qihui and Lin, Bin and Wu, Feize and Yuan, Shenghai and Yan, Zhiyuan and Ye, Yang and Yu, Wangbo and Niu, Yuwei and others},
  journal={arXiv preprint arXiv:2510.16888},
  year={2025}
}

@article{kumari2025learning,
  title={Learning an image editing model without image editing pairs},
  author={Kumari, Nupur and Wang, Sheng-Yu and Zhao, Nanxuan and Nitzan, Yotam and Li, Yuheng and Singh, Krishna Kumar and Zhang, Richard and Shechtman, Eli and Zhu, Jun-Yan and Huang, Xun},
  journal={arXiv preprint arXiv:2510.14978},
  year={2025}
}

@article{bai2025scaling,
  title={Scaling instruction-based video editing with a high-quality synthetic dataset},
  author={Bai, Qingyan and Wang, Qiuyu and Ouyang, Hao and Yu, Yue and Wang, Hanlin and Wang, Wen and Cheng, Ka Leong and Ma, Shuailei and Zeng, Yanhong and Liu, Zichen and others},
  journal={arXiv preprint arXiv:2510.15742},
  year={2025}
}

@article{wei2025skywork,
  title={Skywork unipic 2.0: Building kontext model with online rl for unified multimodal model},
  author={Wei, Hongyang and Xu, Baixin and Liu, Hongbo and Wu, Size and Liu, Jie and Peng, Yi and Wang, Peiyu and Liu, Zexiang and He, Jingwen and Xietian, Yidan and others},
  journal={arXiv preprint arXiv:2509.04548},
  year={2025}
}

@article{tan2026talk2move,
  title={Talk2Move: Reinforcement Learning for Text-Instructed Object-Level Geometric Transformation in Scenes},
  author={Tan, Jing and Zhang, Zhaoyang and Shen, Yantao and Cai, Jiarui and Yang, Shuo and Wu, Jiajun and Xia, Wei and Tu, Zhuowen and Soatto, Stefano},
  journal={arXiv preprint arXiv:2601.02356},
  year={2026}
}

@article{zheng2025yingmusic,
  title={YingMusic-Singer: Zero-shot Singing Voice Synthesis and Editing with Annotation-free Melody Guidance},
  author={Zheng, Junjie and Hao, Chunbo and Ma, Guobin and Zhang, Xiaoyu and Chen, Gongyu and Ding, Chaofan and Chen, Zihao and Xie, Lei},
  journal={arXiv preprint arXiv:2512.04779},
  year={2025}
}

@article{chen2025yingmusic,
  title={YingMusic-SVC: Real-World Robust Zero-Shot Singing Voice Conversion with Flow-GRPO and Singing-Specific Inductive Biases},
  author={Chen, Gongyu and Zhang, Xiaoyu and Weng, Zhenqiang and Zheng, Junjie and Shen, Da and Ding, Chaofan and Zhang, Wei-Qiang and Chen, Zihao},
  journal={arXiv preprint arXiv:2512.04793},
  year={2025}
}

@article{tang2025we,
  title={Are We Ready for RL in Text-to-3D Generation? A Progressive Investigation},
  author={Tang, Yiwen and Guo, Zoey and Zhu, Kaixin and Zhang, Ray and Chen, Qizhi and Jiang, Dongzhi and Liu, Junli and Zeng, Bohan and Song, Haoming and Qu, Delin and others},
  journal={arXiv preprint arXiv:2512.10949},
  year={2025}
}

@article{liu2025nabla,
  title={Nabla-r2d3: Effective and efficient 3d diffusion alignment with 2d rewards},
  author={Liu, Qingming and Liu, Zhen and Zhang, Dinghuai and Jia, Kui},
  journal={arXiv preprint arXiv:2506.15684},
  year={2025}
}

@article{go2025vist3a,
  title={VIST3A: Text-to-3D by Stitching a Multi-view Reconstruction Network to a Video Generator},
  author={Go, Hyojun and Narnhofer, Dominik and Bhat, Goutam and Truong, Prune and Tombari, Federico and Schindler, Konrad},
  journal={arXiv preprint arXiv:2510.13454},
  year={2025}
}

@article{yin2025trim,
  title={TRIM: Scalable 3D Gaussian Diffusion Inference with Temporal and Spatial Trimming},
  author={Yin, Zeyuan and Liu, Xiaoming},
  journal={arXiv preprint arXiv:2511.16642},
  year={2025}
}

@article{hoellmer2026open,
  title={Open Materials Generation with Inference-Time Reinforcement Learning},
  author={Hoellmer, Philipp and Martiniani, Stefano},
  journal={arXiv preprint arXiv:2602.00424},
  year={2026}
}

@article{li2026elign,
  title={Elign: Equivariant Diffusion Model Alignment from Foundational Machine Learning Force Fields},
  author={Li, Yunyang and Huang, Lin and Xia, Luojia and Zhang, Wenhe and Gerstein, Mark},
  journal={arXiv preprint arXiv:2601.21985},
  year={2026}
}

@article{pan2026sa,
  title={SA-VLA: Spatially-Aware Flow-Matching for Vision-Language-Action Reinforcement Learning},
  author={Pan, Xu and Wan, Zhenglin and Yu, Xingrui and Zheng, Xianwei and Ke, Youkai and Sun, Ming and Wang, Rui and Wang, Ziwei and Tsang, Ivor},
  journal={arXiv preprint arXiv:2602.00743},
  year={2026}
}

@article{chen2025pirl,
  title={$\pi$RL: Online rl fine-tuning for flow-based vision-language-action models},
  author={Chen, Kang and Liu, Zhihao and Zhang, Tonghe and Guo, Zhen and Xu, Si and Lin, Hao and Zang, Hongzhi and Zhang, Quanlu and Yu, Zhaofei and Fan, Guoliang and others},
  journal={arXiv preprint arXiv:2510.25889},
  year={2025}
}

@article{lyu2025reinforcement,
  title={Reinforcement Fine-Tuning of Flow-Matching Policies for Vision-Language-Action Models},
  author={Lyu, Mingyang and Sun, Yinqian and Lin, Erliang and Li, Huangrui and Chen, Ruolin and Zhao, Feifei and Zeng, Yi},
  journal={arXiv preprint arXiv:2510.09976},
  year={2025}
}

@article{zhang2025reinforcing,
  title={Reinforcing action policies by prophesying},
  author={Zhang, Jiahui and Huang, Ze and Gu, Chun and Ma, Zipei and Zhang, Li},
  journal={arXiv preprint arXiv:2511.20633},
  year={2025}
}

@article{chen2025one,
  title={One-step flow policy mirror descent},
  author={Chen, Tianyi and Ma, Haitong and Li, Na and Wang, Kai and Dai, Bo},
  journal={arXiv preprint arXiv:2507.23675},
  year={2025}
}

@article{mcallister2025flow,
  title={Flow matching policy gradients},
  author={McAllister, David and Ge, Songwei and Yi, Brent and Kim, Chung Min and Weber, Ethan and Choi, Hongsuk and Feng, Haiwen and Kanazawa, Angjoo},
  journal={arXiv preprint arXiv:2507.21053},
  year={2025}
}

@article{chen2025socialnav,
  title={Socialnav: Training human-inspired foundation model for socially-aware embodied navigation},
  author={Chen, Ziyi and Guo, Yingnan and Chu, Zedong and Luo, Minghua and Shen, Yanfen and Sun, Mingchao and Hu, Junjun and Xie, Shichao and Yang, Kuan and Shi, Pei and others},
  journal={arXiv preprint arXiv:2511.21135},
  year={2025}
}

@article{lei2025embomatrix,
  title={EmboMatrix: A Scalable Training-Ground for Embodied Decision-Making},
  author={Lei, Zixing and Yin, Sheng and Xiong, Yichen and Ding, Yuanzhuo and Huang, Wenhao and Wei, Yuxi and Xu, Qingyao and Li, Yiming and Li, Weixin and Wang, Yunhong and others},
  journal={arXiv preprint arXiv:2510.12072},
  year={2025}
}

@article{yan2025unified,
  title={Unified Multimodal Model as Auto-Encoder},
  author={Yan, Zhiyuan and Lin, Kaiqing and Li, Zongjian and Ye, Junyan and Han, Hui and Wang, Zhendong and Liu, Hao and Lin, Bin and Li, Hao and Xu, Xue and others},
  journal={arXiv preprint arXiv:2509.09666},
  year={2025}
}

@article{chen2025blip3o,
  title={Blip3o-next: Next frontier of native image generation},
  author={Chen, Jiuhai and Xue, Le and Xu, Zhiyang and Pan, Xichen and Yang, Shusheng and Qin, Can and Yan, An and Zhou, Honglu and Chen, Zeyuan and Huang, Lifu and others},
  journal={arXiv preprint arXiv:2510.15857},
  year={2025}
}

@article{wang2025unirl,
  title={UniRL-Zero: Reinforcement Learning on Unified Models with Joint Language Model and Diffusion Model Experts},
  author={Wang, Fu-Yun and Zhang, Han and Gharbi, Michael and Li, Hongsheng and Park, Taesung},
  journal={arXiv preprint arXiv:2510.17937},
  year={2025}
}

@article{tan2025parauni,
  title={ParaUni: Enhance Generation in Unified Multimodal Model with Reinforcement-driven Hierarchical Parallel Information Interaction},
  author={Tan, Jiangtong and Liu, Lin and Huanng, Jie and Zhang, Xiaopeng and Tian, Qi and Zhao, Feng},
  journal={arXiv preprint arXiv:2512.05422},
  year={2025}
}

@article{tian2025mmada,
  title={MMaDA-Parallel: Multimodal Large Diffusion Language Models for Thinking-Aware Editing and Generation},
  author={Tian, Ye and Yang, Ling and Yang, Jiongfan and Wang, Anran and Tian, Yu and Zheng, Jiani and Wang, Haochen and Teng, Zhiyang and Wang, Zhuochen and Wang, Yinjie and others},
  journal={arXiv preprint arXiv:2511.09611},
  year={2025}
}

@article{ye2025visual,
  title={Visual-Aware CoT: Achieving High-Fidelity Visual Consistency in Unified Models},
  author={Ye, Zixuan and Liu, Quande and Wei, Cong and Zhang, Yuanxing and Wang, Xintao and Wan, Pengfei and Gai, Kun and Luo, Wenhan},
  journal={arXiv preprint arXiv:2512.19686},
  year={2025}
}

@article{tang2026endogenous,
  title={Endogenous Reprompting: Self-Evolving Cognitive Alignment for Unified Multimodal Models},
  author={Tang, Zhenchen and Yang, Songlin and Wang, Zichuan and Peng, Bo and Li, Yang and Dong, Beibei and Dong, Jing},
  journal={arXiv preprint arXiv:2601.20305},
  year={2026}
}

@article{wu2026visual,
  title={Visual Generation Unlocks Human-Like Reasoning through Multimodal World Models},
  author={Wu, Jialong and Zhang, Xiaoying and Yuan, Hongyi and Zhang, Xiangcheng and Huang, Tianhao and He, Changjing and Deng, Chaoyi and Zhang, Renrui and Wu, Youbin and Long, Mingsheng},
  journal={arXiv preprint arXiv:2601.19834},
  year={2026}
}

@article{liu2025mixture,
  title={Mixture of States: Routing Token-Level Dynamics for Multimodal Generation},
  author={Liu, Haozhe and Liu, Ding and Zhuge, Mingchen and Zhou, Zijian and Xie, Tian and He, Sen and Yang, Yukang and Liu, Shuming and Cong, Yuren and Guo, Jiadong and others},
  journal={arXiv preprint arXiv:2511.12207},
  year={2025}
}

@article{modelmanzano,
  title={MANZANO: ASIMPLE AND SCALABLE UNIFIED MUL-TIMODAL MODEL WITH A HYBRID VISION TOKENIZER},
  author={MODEL, TIMODAL}
}

@article{wang2025fudoki,
  title={Fudoki: Discrete flow-based unified understanding and generation via kinetic-optimal velocities},
  author={Wang, Jin and Lai, Yao and Li, Aoxue and Zhang, Shifeng and Sun, Jiacheng and Kang, Ning and Wu, Chengyue and Li, Zhenguo and Luo, Ping},
  journal={arXiv preprint arXiv:2505.20147},
  year={2025}
}

@article{seedream2025seedream,
  title={Seedream 4.0: Toward next-generation multimodal image generation},
  author={Seedream, Team and Chen, Yunpeng and Gao, Yu and Gong, Lixue and Guo, Meng and Guo, Qiushan and Guo, Zhiyao and Hou, Xiaoxia and Huang, Weilin and Huang, Yixuan and others},
  journal={arXiv preprint arXiv:2509.20427},
  year={2025}
}

@article{chen2026unified,
  title={Unified Text-Image Generation with Weakness-Targeted Post-Training},
  author={Chen, Jiahui and Hansen-Estruch, Philippe and Han, Xiaochuang and Hu, Yushi and Dinan, Emily and Kamath, Amita and Drozdzal, Michal and Askari-Hemmat, Reyhane and Zettlemoyer, Luke and Ghazvininejad, Marjan},
  journal={arXiv preprint arXiv:2601.04339},
  year={2026}
}

@article{zhang2025cooper,
  title={COOPER: A Unified Model for Cooperative Perception and Reasoning in Spatial Intelligence},
  author={Zhang, Zefeng and Hao, Xiangzhao and Tang, Hengzhu and Zhang, Zhenyu and Sheng, Jiawei and Li, Xiaodong and Li, Zhenyang and Gao, Li and Shi, Daiting and Yin, Dawei and others},
  journal={arXiv preprint arXiv:2512.04563},
  year={2025}
}

@article{liu2025showtable,
  title={ShowTable: Unlocking Creative Table Visualization with Collaborative Reflection and Refinement},
  author={Liu, Zhihang and Bao, Xiaoyi and Li, Pandeng and Zhou, Junjie and Liao, Zhaohe and He, Yefei and Jiang, Kaixun and Xie, Chen-Wei and Zheng, Yun and Xie, Hongtao},
  journal={arXiv preprint arXiv:2512.13303},
  year={2025}
}

@article{yang2026policyflow,
  title={PolicyFlow: Policy Optimization with Continuous Normalizing Flow in Reinforcement Learning},
  author={Yang, Shunpeng and Liu, Ben and Chen, Hua},
  journal={arXiv preprint arXiv:2602.01156},
  year={2026}
}

@article{zou2026one,
  title={One Step Is Enough: Dispersive MeanFlow Policy Optimization},
  author={Zou, Guowei and Wang, Haitao and Wu, Hejun and Qian, Yukun and Wang, Yuhang and Li, Weibing},
  journal={arXiv preprint arXiv:2601.20701},
  year={2026}
}

@article{ma2025stage,
  title={Stage: Stable and generalizable grpo for autoregressive image generation},
  author={Ma, Xiaoxiao and Qiu, Haibo and Zhang, Guohui and Zeng, Zhixiong and Yang, Siqi and Ma, Lin and Zhao, Feng},
  journal={arXiv preprint arXiv:2509.25027},
  year={2025}
}

@article{sun2026var,
  title={VAR RL Done Right: Tackling Asynchronous Policy Conflicts in Visual Autoregressive Generation},
  author={Sun, Shikun and Qu, Liao and Zhang, Huichao and Liu, Yiheng and Song, Yangyang and Li, Xian and Wang, Xu and Jiang, Yi and Du, Daniel K and Wu, Xinglong and others},
  journal={arXiv preprint arXiv:2601.02256},
  year={2026}
}

@article{zhang2025group,
  title={Group Critical-token Policy Optimization for Autoregressive Image Generation},
  author={Zhang, Guohui and Yu, Hu and Ma, Xiaoxiao and Zhang, JingHao and Pan, Yaning and Yao, Mingde and Xiao, Jie and Huang, Linjiang and Zhao, Feng},
  journal={arXiv preprint arXiv:2509.22485},
  year={2025}
}

@article{zheng2025layout,
  title={Layout-Conditioned Autoregressive Text-to-Image Generation via Structured Masking},
  author={Zheng, Zirui and Isobe, Takashi and Shen, Tong and Jia, Xu and Zhao, Jianbin and Li, Xiaomin and Ge, Mengmeng and Li, Baolu and Wang, Qinghe and Li, Dong and others},
  journal={arXiv preprint arXiv:2509.12046},
  year={2025}
}

@article{luo2025reinforcement,
  title={Reinforcement learning meets masked generative models: Mask-grpo for text-to-image generation},
  author={Luo, Yifu and Hu, Xinhao and Fan, Keyu and Sun, Haoyuan and Chen, Zeyu and Xia, Bo and Zhang, Tiantian and Chang, Yongzhe and Wang, Xueqian},
  journal={arXiv preprint arXiv:2510.13418},
  year={2025}
}

@article{ma2025consolidating,
  title={Consolidating Reinforcement Learning for Multimodal Discrete Diffusion Models},
  author={Ma, Tianren and Zhang, Mu and Wang, Yibing and Ye, Qixiang},
  journal={arXiv preprint arXiv:2510.02880},
  year={2025}
}

@article{wan2025discrete,
  title={Discrete guidance matching: Exact guidance for discrete flow matching},
  author={Wan, Zhengyan and Ouyang, Yidong and Xie, Liyan and Fang, Fang and Zha, Hongyuan and Cheng, Guang},
  journal={arXiv preprint arXiv:2509.21912},
  year={2025}
}

@article{tang2025tr2,
  title={Tr2-d2: Tree search guided trajectory-aware fine-tuning for discrete diffusion},
  author={Tang, Sophia and Zhu, Yuchen and Tao, Molei and Chatterjee, Pranam},
  journal={arXiv preprint arXiv:2509.25171},
  year={2025}
}

@article{kang2025ladir,
  title={Ladir: Latent diffusion enhances llms for text reasoning},
  author={Kang, Haoqiang and Zhang, Yizhe and Kuang, Nikki Lijing and Majamaki, Nicklas and Jaitly, Navdeep and Ma, Yi-An and Qin, Lianhui},
  journal={arXiv preprint arXiv:2510.04573},
  year={2025}
}

@article{kang2026beyond,
  title={Beyond Mode Elicitation: Diversity-Preserving Reinforcement Learning via Latent Diffusion Reasoner},
  author={Kang, Haoqiang and Zhang, Yizhe and Kuang, Nikki Lijing and Ma, Yi-An and Qin, Lianhui},
  journal={arXiv preprint arXiv:2602.01705},
  year={2026}
}

@article{bin2025lrpo,
  title={LRPO: Enhancing Blind Face Restoration through Online Reinforcement Learning},
  author={Bin, WU and Liu, Yahui and Zhang, Chi and Zhao, Yao and Wang, Wei},
  year={2025}
}

@article{yao2026bridging,
  title={Bridging Information Asymmetry: A Hierarchical Framework for Deterministic Blind Face Restoration},
  author={Yao, Zhengjian and Hu, Jiakui and Li, Kaiwen and He, Hangzhou and Zhang, Xinliang and Zeng, Shuang and Zhu, Lei and Lu, Yanye},
  journal={arXiv preprint arXiv:2601.19506},
  year={2026}
}

@article{liu2025finpercep,
  title={FinPercep-RM: A Fine-grained Reward Model and Co-evolutionary Curriculum for RL-based Real-world Super-Resolution},
  author={Liu, Yidi and Fan, Zihao and Huang, Jie and Xiao, Jie and Li, Dong and Zhang, Wenlong and Bai, Lei and Fu, Xueyang and Zha, Zheng-Jun},
  journal={arXiv preprint arXiv:2512.22647},
  year={2025}
}

@article{xu2025enhancing,
  title={Enhancing Diffusion-based Restoration Models via Difficulty-Adaptive Reinforcement Learning with IQA Reward},
  author={Xu, Xiaogang and Chu, Ruihang and Wang, Jian and Zhou, Kun and Shu, Wenjie and Yang, Harry and Lim, Ser-Nam and Chen, Hao and Lin, Liang},
  journal={arXiv preprint arXiv:2511.01645},
  year={2025}
}

@article{hu2025flowlut,
  title={FlowLUT: Efficient Image Enhancement via Differentiable LUTs and Iterative Flow Matching},
  author={Hu, Liubing and Wu, Chen and Wang, Anrui and Lu, Dianjie and Zhang, Guijuan and Zheng, Zhuoran},
  journal={arXiv preprint arXiv:2509.23608},
  year={2025}
}

@article{dirik2025reasonx,
  title={ReasonX: MLLM-Guided Intrinsic Image Decomposition},
  author={Dirik, Alara and Wang, Tuanfeng and Ceylan, Duygu and Zafeiriou, Stefanos and Fr{\"u}hst{\"u}ck, Anna},
  journal={arXiv preprint arXiv:2512.04222},
  year={2025}
}

@article{wen2025hy,
  title={HY-Motion 1.0: Scaling Flow Matching Models for Text-To-Motion Generation},
  author={Wen, Yuxin and Shuai, Qing and Kang, Di and Li, Jing and Wen, Cheng and Qian, Yue and Jiao, Ningxin and Chen, Changhai and Chen, Weijie and Wang, Yiran and others},
  journal={arXiv preprint arXiv:2512.23464},
  year={2025}
}

@article{zhuo2024lumina,
  title={Lumina-next: Making lumina-t2x stronger and faster with next-dit},
  author={Zhuo, Le and Du, Ruoyi and Xiao, Han and Li, Yangguang and Liu, Dongyang and Huang, Rongjie and Liu, Wenze and Zhu, Xiangyang and Wang, Fu-Yun and Ma, Zhanyu and others},
  journal={Advances in Neural Information Processing Systems},
  volume={37},
  pages={131278--131315},
  year={2024}
}

@article{wan2025wan,
  title={Wan: Open and advanced large-scale video generative models},
  author={Wan, Team and Wang, Ang and Ai, Baole and Wen, Bin and Mao, Chaojie and Xie, Chen-Wei and Chen, Di and Yu, Feiwu and Zhao, Haiming and Yang, Jianxiao and others},
  journal={arXiv preprint arXiv:2503.20314},
  year={2025}
}

@article{li2025triposg,
  title={Triposg: High-fidelity 3d shape synthesis using large-scale rectified flow models},
  author={Li, Yangguang and Zou, Zi-Xin and Liu, Zexiang and Wang, Dehu and Liang, Yuan and Yu, Zhipeng and Liu, Xingchao and Guo, Yuan-Chen and Liang, Ding and Ouyang, Wanli and others},
  journal={IEEE Transactions on Pattern Analysis and Machine Intelligence},
  year={2025},
  publisher={IEEE}
}

@inproceedings{li2025shapegen,
  title={ShapeGen: Towards High-Quality 3D Shape Synthesis},
  author={Li, Yangguang and He, Xianglong and Zou, Zi-Xin and Liu, Zexiang and Ouyang, Wanli and Liang, Ding and Cao, Yan-Pei},
  booktitle={Proceedings of the SIGGRAPH Asia 2025 Conference Papers},
  pages={1--12},
  year={2025}
}

@article{du2024cosyvoice,
  title={Cosyvoice 2: Scalable streaming speech synthesis with large language models},
  author={Du, Zhihao and Wang, Yuxuan and Chen, Qian and Shi, Xian and Lv, Xiang and Zhao, Tianyu and Gao, Zhifu and Yang, Yexin and Gao, Changfeng and Wang, Hui and others},
  journal={arXiv preprint arXiv:2412.10117},
  year={2024}
}

@article{chen2025flow,
  title={Flow marching for a generative PDE foundation model},
  author={Chen, Zituo and Deng, Sili},
  journal={arXiv preprint arXiv:2509.18611},
  year={2025}
}

@article{yuan2025ar,
  title={Ar-grpo: Training autoregressive image generation models via reinforcement learning},
  author={Yuan, Shihao and Liu, Yahui and Yue, Yang and Zhang, Jingyuan and Zuo, Wangmeng and Wang, Qi and Zhang, Fuzheng and Zhou, Guorui},
  journal={arXiv preprint arXiv:2508.06924},
  year={2025}
}
